\documentclass[journal]{IEEEtran}

\usepackage{bm}
\usepackage{color}
\usepackage{xcolor}
\usepackage{algorithmic,algorithm}
\usepackage{booktabs}
\usepackage{threeparttable}
\usepackage{multirow}
\usepackage{url}
\usepackage{amsmath}

\ifCLASSINFOpdf
  \usepackage[pdftex]{graphicx}
  \DeclareGraphicsExtensions{.pdf,.jpeg,.png}
\else
   \usepackage[dvips]{graphicx}
  \DeclareGraphicsExtensions{.eps}
\fi

\ifCLASSOPTIONcompsoc
  \usepackage[caption=false,font=normalsize,labelfont=sf,textfont=sf]{subfig}
\else
  \usepackage[caption=false,font=footnotesize]{subfig}
\fi

\begin{document}

\title{Background Subtraction with Real-time Semantic Segmentation}

\author{Dongdong~Zeng,~ %\IEEEmembership{Member,~IEEE,}
        Xiang~Chen,~
        Ming~Zhu,~
        Michael~Goesele
        and~Arjan~Kuijper% <-this % stops a space

%\thanks{This work was supported by  the National Nature Science Foundation of China under Grant No.61401425.}% <-this % stops a space

\thanks{Dongdong Zeng is with the Changchun Institute of Optics, Fine Mechanics and Physics, Chinese Academy of Sciences, Changchun 130033, China. He is also with the University of Chinese Academy of Sciences, Beijing 100049, China, and Fraunhofer IGD, 64283 Darmstadt, Germany(e-mail: zengdongdong13@mails.ucas.edu.cn).}% <-this % stops a space

\thanks{Ming Zhu  is with the Changchun Institute of Optics, Fine Mechanics and Physics, Chinese Academy of Sciences, Changchun 130033, China (e-mail: zhu\_mingca@163.com).}

\thanks{Xiang Chen and Michael Goesele are with the Graphics, Capture and Massively Parallel Computing, TU Darmstadt, 64283 Darmstadt, Germany (e-mail: xiang.chen@gcc.tu-darmstadt.de, goesele@gcc.tu-darmstadt.de).}% <-this % stops a space

\thanks{Arjan Kuijper is with the Mathematical and Applied Visual Computing, TU Darmstadt, 64283 Darmstadt, Germany, and Fraunhofer IGD, 64283 Darmstadt, Germany(e-mail: arjan.kuijper@igd.fraunhofer.de).}

%\thanks{Manuscript received April 19, 2005; revised August 26, 2015.}

}

%\markboth{IEEE TRANSACTIONS ON IMAGE PROCESSING}%
%{Shell \MakeLowercase{\textit{et al.}}: Bare Demo of IEEEtran.cls for IEEE Journals}

% make the title area
\maketitle

%%%%% remember replace the Figure to Fig
\begin{abstract}
% !TEX root = paper_main.tex
Accurate and fast foreground object extraction is very important for object tracking and recognition in video surveillance.
Although many background subtraction (BGS)  methods have been proposed in the recent past, it is still regarded as a tough problem due to  the variety of challenging situations that occur in real-world scenarios.
In this paper, we explore this problem from a new perspective and propose a novel background subtraction framework with real-time semantic segmentation (RTSS).
Our proposed framework consists of two components, a traditional BGS segmenter $\mathcal{B}$ and a real-time semantic segmenter $\mathcal{S}$.
The BGS segmenter $\mathcal{B}$ aims to construct background models and segments foreground objects. The real-time semantic segmenter $\mathcal{S}$ is used to refine the foreground segmentation outputs as feedbacks for improving the model updating accuracy.
$\mathcal{B}$ and $\mathcal{S}$ work in parallel on two threads. For each input frame $I_t$, the BGS segmenter  $\mathcal{B}$ computes a preliminary foreground/background (FG/BG) mask $B_t$. At the same time, the real-time semantic segmenter $\mathcal{S}$ extracts the object-level semantics  ${S}_t$. Then, some specific rules are applied on ${B}_t$ and ${S}_t$ to generate  the final detection ${D}_t$. Finally, the refined FG/BG mask ${D}_t$ is fed back to update the background model.
Comprehensive experiments evaluated on the CDnet 2014 dataset  demonstrate that our proposed method  achieves state-of-the-art performance among all unsupervised background subtraction methods while  operating at real-time, and even performs better than some deep learning based supervised algorithms. In addition, our proposed framework is very flexible and has the potential for generalization.
\end{abstract}


% For peer review papers, you can put extra information on the cover
% page as needed:
% \ifCLASSOPTIONpeerreview
% \begin{center} \bfseries EDICS Category: 3-BBND \end{center}
% \fi
%
% For peerreview papers, this IEEEtran command inserts a page break and
% creates the second title. It will be ignored for other modes.
\IEEEpeerreviewmaketitle

% body
% !TEX root = paper_main.tex

\section{Introduction}

\IEEEPARstart{B}{ackground} subtraction based on change detection is a widely studied topic in computer vision. As a basic preprocessing step in video processing, it is used in many high-level computer vision applications such as video surveillance, traffic monitoring, gesture recognition or remote sensing.
Generally, a complete background subtraction technology contains at least four components: a background model initialization process, a background model representation strategy, a background model maintenance mechanism, and a foreground detection operation. % (see Figure \ref{foursteps}).
%\begin{figure}
%\centering
%\includegraphics[width=1.0\linewidth]{foursteps.pdf}
%\caption{Block diagram of the background subtraction process.}
%\label{foursteps}
%\end{figure}
The output of a background subtraction process is a binary mask which divides the input frame pixels into sets of background and foreground pixels.
Needless to say, the quality of segmentation affects any subsequent higher-level tasks.

\begin{figure}[t]
\centering
\includegraphics[width=1.0\linewidth]{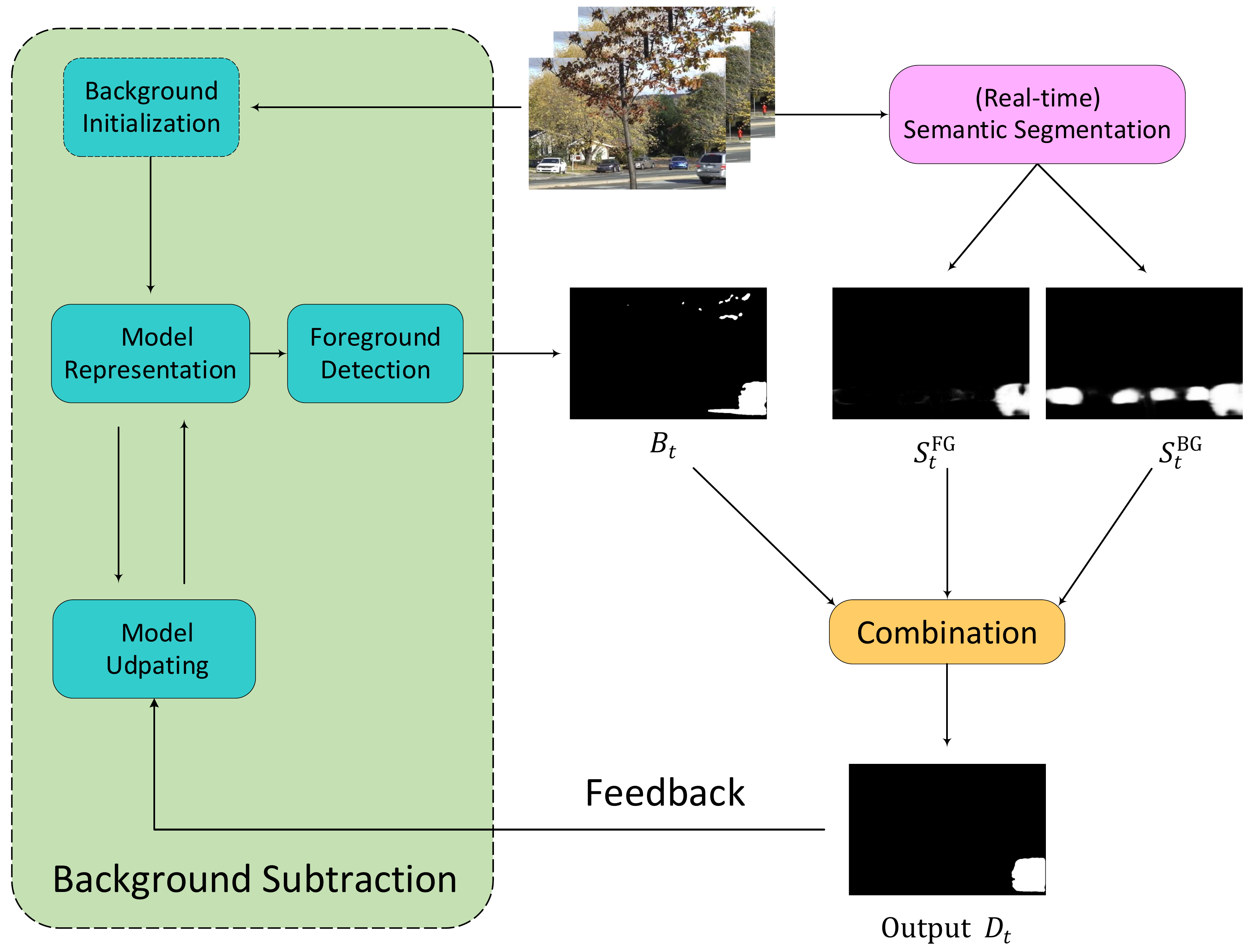}
\vspace{-0.04\linewidth}
\caption{Algorithm flowchart of the proposed method.}  %(a) Algorithm flowchart of the method presented in~\cite{braham2017semantic}. (b)
\label{algorithmContrast}
\end{figure}

The simplest examples of background subtraction are based on the idea that the current frame is compared with a ``static'' background image. Pixels with a high deviation from a threshold are determined as foreground, otherwise as background. This strategy might work in certain specialized scenes. Unfortunately, a clean background image is often not available in real scenarios due to several challenges such as dynamic background, illumination changes or hard shadows. Thus, a multitude of more sophisticated background subtraction methods has been developed in the recent past.
Most researchers  are  mainly focused on two aspects~\cite{bouwmans2014traditional, sobral2014comprehensive}.  The first one is to develop more advanced background models such as Gaussian Mixture Models~\cite{stauffer1999adaptive, zivkovic2006efficient, allili2007robust}, Kernel Density Estimation~\cite{elgammal2002background}, CodeBook~\cite{kim2005real, guo2011hierarchical}, and non-parametric approaches such as ViBe~\cite{barnich2011vibe},
SuBSENSE~\cite{st2015subsense}, PAWCS~\cite{st2016universal}. The second one is to employ  more robust feature representations including color features, edge features~\cite{javed2013foreground}, stereo features~\cite{maddalena2017exploiting}, motion features~\cite{zhou2005modified}, and texture features~\cite{heikkila2006texture, st2014improving}.
Although existing state-of-the-art algorithms have made a great progress, it remains a challenging problem to address the complexity of real scenarios and to produce satisfactory results.

In this paper, we try to solve this problem from a new perspective and propose a novel background subtraction framework by combining conventional BGS algorithms and semantic segmentation techniques.
%Semantic segmentation is a fundamental task in computer vision, which is regarded as an  important tool for providing a complete understanding of the scene.
The goal of semantic segmentation is to predict predefined category labels for each pixel in the image, it gives the locations and shapes of semantic entities found in street scenarios, such as roads, cars, pedestrians, sidewalks, and so on. In recent years, the development of deep convolutional neural networks (CNNs) yielded remarkable progress on semantic segmentation, providing a possibility for leveraging accurate object-level semantic in the scenes.
If we look back at foreground detection results, we can find that in most cases the foreground objects that we are interested in are actually included in the semantic segmentation categories.
However, traditional BGS algorithms are very sensitive to illumination changes, dynamic backgrounds, shadows, and ghosts. These factors have great influence on the reliability of the constructed background model, producing many false positive areas (objects).
Fortunately, semantic segmentation shows great robustness to these factors.
Therefore, we consider combining traditional BGS algorithms and semantic segmentation to create more reliable background models and to improve foreground detection accuracy.

Our idea is inspired by Braham \textit{et al.}~\cite{braham2017semantic}, who present a method to improve  the  BGS segmentation with a post-processing operation by fusing  the  segmentation output of a BGS algorithm and a semantic segmentation algorithm.
%The pipeline of their method is shown in Fig.~\ref{algorithmContrast}(a).
%The \textit{Background Subtraction} block can be any BGS algorithm. For the \textit{Semantic Segmentation} block, the authors used PSPNet~\cite{zhao2017pyramid}, which achieved state-of-the-art segmentation accuracy in the PASCAL VOC 2012 object segmentation dataset.
This method has however several drawbacks. Firstly, it is inefficient in practical applications.  As the authors tested in their paper, the semantic segmentation algorithm PSPNet~\cite{zhao2017pyramid} can only process less than 7 frames per second for a $473\times473$ resolution image with an NVIDIA GTX Titan X GPU so that it is difficult to reach real-time performance.  Secondly,  the semantic segmentation  only acts as a post-processing step to refine the BGS result, without any interaction with the background subtraction process. If the fusion result can be used to feedback and guide the background model updating, more accurate results may be obtained.

Through the above observations, we propose a novel background subtraction framework called \textit{background subtraction with real-time semantic segmentation} (RTSS).
The framework of our RTSS is illustrated in Fig.~\ref{algorithmContrast}. The key idea is to decompose the original background subtraction into two parallel but collaborative components~\cite{fan2017parallel}: a traditional BGS segmenter $\mathcal{B}$ and a real-time semantic segmenter $\mathcal{S}$.
The BGS segmenter $\mathcal{B}$ aims to construct  background models and segments foreground objects. The semantic segmenter $\mathcal{S}$  cooperates to refine the foreground segmentation result and feedback to improve the model updating accuracy.
The BGS segmenter can be any real-time BGS algorithms.
For semantic segmenter, in order to obtain real-time semantic information, we use two strategies. The first is to use real-time semantic segmentation algorithms, and the second is to perform semantic segmentation once every N frames.
%  we use a real-time semantic segmentation algorithm ERFNet~\cite{romera2018erfnet}, which achieved remarkable accuracy on various scene parsing datasets such as Cityscapes~\cite{cordts2016cityscapes} and ADE20K~\cite{zhou2017scene} dataset while operating at real-time.
$\mathcal{B}$ and $\mathcal{S}$ work in parallel on two threads. For each input frame $I_t$, the BGS segmenter  $\mathcal{B}$ performs the background subtraction process to produce a preliminary foreground/background (FG/BG) mask ${B}_t$. At the same time, the real-time semantic segmenter $\mathcal{S}$ segments object-level semantics  ${S}_t$. Then, some specific rules are applied on ${B}_t$ and ${S}_t$ and producing the final detection result ${D}_t$. Finally, the refined FG/BG mask ${D}_t$ is fed back to update the background model.
In contrast to~\cite{braham2017semantic}, the proposed method can not only run in real time but also achieves higher  detection precision due to the  feedback mechanism.

We thus make the following contributions:
\begin{enumerate}
\item We propose a novel background subtraction framework which combines traditional unsupervised BGS algorithms and real-time semantic segmentation algorithms. Experimental results show that RTSS  achieved state-of-the-art performance among all unsupervised BGS methods while operating at real-time and even performs better than some deep learning based methods
%\item Our proposed framework improves processing speed and enables real-time processing.
\item The proposed framework has a component based structure, allowing for flexibly replacing components and adaption to new applications.
\end{enumerate}

The rest of this paper is organized as follows. Relevant work is discussed in Section~\ref{sec:relatedworks}.
Then in Section~\ref{RTSS}, we elaborate the framework of RTSS. Section~\ref{Experimental results} presents the experiments and comparisons with other state-of-the-art algorithms on the CDnet 2014 dataset~\cite{wang2014cdnet}. Final conclusions are given in Section~\ref{CONCLUSION}.

% !TEX root = paper_main.tex

\section{Related Works}
\label{sec:relatedworks}

Background subtraction and semantic segmentation have been extensively studied. A full survey is out of the scope of this paper. In the following we discuss some related works and refer readers to~\cite{bouwmans2014traditional, sobral2014comprehensive, bouwmans2018deep, garcia2017review} for a thorough review.

\textbf{Background subtraction:}
Existing background subtraction algorithms can be categorized as traditional unsupervised  algorithms and recent supervised algirithms which based on deep learning.
The most popular and classic technology   is the  Gaussian Mixture Model (GMM)~\cite{stauffer1999adaptive} which models the per-pixel distribution of gray values observed over time with a mixture of Gaussians. Since its introduction, GMM has been widely used in different scenarios and various improved versions have been proposed~\cite{zivkovic2006efficient, lee2005effective}.
However, GMM model presents some disadvantages. For example, when the background changes quickly, it is difficult to be modeled  with just a few Gaussian models, and it may also fail to provide correct detections.
To overcome these problems, a non-parametric  approach using Kernel Density Estimation (KDE) technique was proposed in~\cite{elgammal2002background}, which estimates the probability density function at each pixel from many samples without any prior assumptions.
This model is robust and can adapt quickly to the background changes with small motions. A more advanced method using adaptive kernel density estimation was also proposed in~\cite{mittal2004motion}.
The Codebook is another classic technology that has been introduced by Kim \textit{et al.} in~\cite{kim2005real}. The values for each pixel are modeled into codebooks which represent a compressed form of background model in a long image sequence. Each codebook is composed of codewords constructed from intensity and temporal features. Incoming pixels
matched against one of the codewords are classified as background; otherwise, foreground.
The more recent ViBe algorithm~\cite{barnich2011vibe}  is built on similar principles as GMMs, but instead of building the probability distribution of the background  pixels using a Parzen window, the authors try to store the distributions as a collection of samples at random.
If a pixel in the new input frame matches a portion of its background samples, it is considered  to be background and may be selected for model update.
Local Binary Pattern (LBP) feature~\cite{heikkila2006texture} is one of the  earliest and popular texture operator  proposed for background subtraction and has shown favourable performance due to its computational simplicity and tolerance against illumination variations. However, LBP is sensitive to subtle local texture changes  caused by image noise.
To deal with this problem, Tan \textit{et al.}~\cite{tan2010enhanced} proposed the Local Ternary Pattern (LTP) operator by adding a small tolerance offset. LTP is however not invariant to illumination variations with scale transform  as the tolerance is constant for every pixel position. Recently, the Scale Invariant Local Ternary Pattern (SILTP) operator was proposed in~\cite{liao2010modeling} and has shown its  robustness against illumination variations and local image noise within a range because it employs  a scalable tolerance offset associated with the center pixel intensity.
More recently, Local Binary Similarity Pattern (LBSP)  was proposed by St-Charles \textit{et al.}~\cite{st2014improving}, this operator is based on absolute difference and is calculated  both inside a region of an image and across regions between two images to take more spatiotemporal information into consideration.
The authors also improved the method by combining LBSP features and pixel intensities with a pixel-level feedback loop model maintainance mechanism. The new method called SuBSENSE~\cite{st2015subsense} achieves state-of-the-art performance among all traditional BGS algorithms.

In the past few years, many CNN-based background subtraction algorithms~{\cite{braham2016deep,babaee2018deep,wang2017interactive,lim2018foreground, zeng2018multiscale}} have been proposed.  The first scene specific CNN-based background subtraction method is proposed in~\cite{braham2016deep}. First, a fixed background image is generated by a temporal median operation over the first 150 frames. Then, for each frame in a sequence, image patches centered on each pixel are extracted from the current frame and the background image. After that,  the combined patches are fed to the network to get the probability of current pixel. After comparing with a score threshold, the pixel is either classified background or foreground.
Babaee \textit{et al.}~\cite{babaee2018deep} proposed a novel background image generation with a motion detector, which produces a more precise background image. Instead of training one network per scene, the authors collected 5\% of frames from each sequence as the training data then trained only one model for all sequences. However, the precision of this algorithm is much worse than that of~\cite{braham2016deep}.
Wang \textit{et al.}~\cite{wang2017interactive} proposed a multiscale cascade-like convolutional neural network architecture for background subtraction without constructing  background images. %They trained the network with 200 manually selected frames and achieved an F-Measure of 0.92 in the CDnet 2014 dataset.
More recently,  Lim \textit{et al.}~\cite{lim2018foreground} proposed an encoder-decoder type neural network with triplet CNN and transposed CNN configuration.
Zeng \textit{et al.}~\cite{zeng2018multiscale} proposed a multiscale fully convolutional network  architecture which takes advantage of different layer features for background subtraction.
Although the previous  mentioned deep learning based algorithms perform better than the traditional algorithms,  they all are supervised. During the training stage, ground truth constructed by a human expert or other unsupervised traditional BGS methods are needed to train their model.
However, strictly speaking, background subtraction methods for video surveillance should be unsupervised.   Therefore, it could be argued whether existing CNN-based background subtraction methods are useful in practical applications.

\textbf{Semantic segmentation:}
Over the last years, development of  convolutional neural networks has enabled remarkable progress in semantic segmentation.  Early approaches performed pixel-wise classification by  explicitly passing image patches through CNNs~\cite{farabet2013learning}.  In~\cite{long2015fully},  the fully convolutional network (FCN) proposed by Long \textit{et al.} opened a new idea for semantic segmentation with end-to-end training. Transposed convolution operations are utilized to upsample low resolution features.
This network architecture is far more efficient than pixel-wise prediction  since it avoids redundant computation on  overlapping patches.
However, the typical FCN architectures involve a number of pooling layers to significantly increase the receptive field size and to make the network more robust against small translations. As a result, the network outputs are a low-resolution. To address this issue, various strategies have been proposed.
Some approaches use dilated convolutions to augment the receptive field size without losing resolution~\cite{yu2015multi, chen2018deeplab},  extract features from different layers with skip-connections operation~\cite{ronneberger2015u, liu2015parsenet}, or multi-scale feature ensembles~\cite{chen2016attention, xia2016zoom}.
Noh \textit{et al.}~\cite{noh2015learning} proposed an encoder-decoder structure. The encoder learns low-dimensional feature representations via  pooling and convolution operations. While the decoder tries to upscale these low-dimensional features via a sequence of unpooling and deconvolution operations.
In~\cite{chen2018deeplab, liu2015semantic}, conditional random fields (CRFs) are applied on the network output in order to obtain more consistent results.
Zhao \textit{et al.}~\cite{zhao2017pyramid} proposed the pyramid scene network (PSPNet) for scene parsing by  embedding difficult scenery context features, and achieved state-of-the-art performance on various semantic segmentation datasets.

All of the above mentioned segmentation approaches mainly focused on accuracy and robustness. Most of them are computationally expensive and cannot be applied in real-time applications. Recent interest in self-driving cars has created a strong need for real-time semantic segmentation algorithms.
%Some works tried to address the segmentation network efficiency problem. %such as~\cite{badrinarayanan2017segnet,paszke2016enet,chaurasia2017linknet, romera2018erfnet, zhao2018icnet}
SegNet~\cite{badrinarayanan2017segnet} tries to abandon layers to reduce network parameters to improve the network reference speed. LinkNet~\cite{chaurasia2017linknet} presents an  architecture that is based on residual connections, and Paszke \textit{et al.}~\cite{paszke2016enet} proposed a lightweight residual network ENet by sacrificing layers to gain efficiency for real-time semantic segmentation.
%Although these networks can achieve real-time running speed, their accuracy drops notably.
More recently, ERFNet~\cite{romera2018erfnet} proposed a novel architecture with residual connections and factorized convolutions, which made a good trade-off between high accuracy and real-time speed. This system is able to run at 83 FPS   for  $640 \times 360$  resolution image with  an NVIDIA GTX Titan X GPU  and  achieves remarkable accuracy on various semantic segmentation datasets.
ICNet~\cite{zhao2018icnet} proposed a compressed-PSPNet-based image cascade network for pixel-wise label inference. Multi-resolution branches under proper label guidance are used to reduce  network computational complexity.  This real-time system can process $1024 \times 2048$ resolution video stream at a speed of 30 fps while accomplishing acceptable segmentation precision.

% !TEX root = paper_main.tex

\section{Background subtraction with real-time semantic segmentation}\label{RTSS}

In this section, we will give a detailed description of the novel background subtraction framework RTSS (\textit{background subtraction with real-time semantic segmentation}).
The RTSS framework consists of two components: a BGS segmenter $\mathcal{B}$ and a semantic segmenter $\mathcal{S}$. The $\mathcal{B}$ and  $\mathcal{S}$ work together toward real-time and accuracy foreground detection.  In this paper, we choose  SuBSENSE~\cite{st2015subsense} as the benchmark  BGS segmenter and ICNet~\cite{zhao2018icnet} as the benchmark semantic segmenter.

\subsection{SuBSENSE for Background Subtraction}\label{SubSENSEBGS}

%\footnote{\color{blue}{http://www.changedetection.net/}}
We choose SuBSENSE~\cite{st2015subsense} as the benchmark BGS segmenter as it shows state-of-the-art performance among all real-time unsupervised BGS algorithms.
In the past few years, various non-parametric background subtraction methods have been proposed and have shown to outperform many existing methods.
Barnich \textit{et al.}~\cite{barnich2011vibe} proposed the ViBe method whose background model is constructed by  a collection of  $N$ samples randomly selected over time. When a background sample is updated, its neighboring pixels also have a probability to update its background model according to the neighborhood diffusion strategy.
Inspired by  control system theory,  Hofmann \textit{et al.}~\cite{hofmann2012background} proposed an improved   Vibe algorithm  by dynamically adjusting the decision threshold and the learning rate.
Then, St-Charles \textit{et al.}  proposed the SuBSENSE method~\cite{st2015subsense}, which combines color and local binary similarity pattern (LBSP) features to improve the spatial awareness  and using a pixel-level feedback mechanism to dynamically adjust its parameters.
A brief overview of SubSENSE is presented as follows.
We define the background model $\mathcal{M}(x)$ at pixel $x$ as:
\begin{equation}\label{B_x_model}
\mathcal{M}(x) = \{ \mathcal{M}_1(x), \mathcal{M}_2(x), \ldots, \mathcal{M}_N(x)  \}
\end{equation}
which contains $N$ recent background samples.
To classify a pixel $x$ in the $t$-th  input frame $I_t$ as foreground or background, it will be matched against its background samples according to (\ref{classification}):
\begin{equation} \label{classification}
B_t(x) = \begin{cases}
1, & \text{if }  \#\{ dist(I_t(x), \mathcal{M}_n(x)) < R, \forall n     \} < \#_{min} \\
0, & \text{otherwise. }
\end{cases}
\end{equation}
where $B_t(x)$ is the output segmentation result,  $B_t(x)$ = 1 means foreground and  $B_t(x)$ =  0 means background.
$ dist(I_t(x), \mathcal{M}_n(x))   $ returns the distance between the input pixel $I_t(x)$ and a background sample $\mathcal{M}_n(x)$. $R$ is the distance threshold which can be dynamically changed for each pixel over time. If the distance between $I_t(x)$ and $\mathcal{M}_n(x)$  is less than the threshold $R$, a match is found. And $\#_{min}$ is the minimum number of matches required for classifying a pixel as background, usually $\#_{min}$ is fixed as 2.

%
%\begin{figure}[t]
%\centering
%   \includegraphics[width=1.0\linewidth]{mySuBSENSE_new.pdf}
%   \caption{Schematic diagram of the SuBSENSE algorithm.}
%\label{SuBSENSEdiagram}
%\end{figure}

To increase the model robustness and flexibility, the distance threshold $R(x)$ need to be  dynamically  adjusted per-pixel.
A feedback mechanism based on two-pronged background monitoring is proposed. First, to measure the motion entropy of dynamic background, a new controller $D_{min}$ is defined:
\begin{equation}\label{Dmin}
D_{min}(x) = D_{min}(x) \cdot (1- \alpha) + d_t(x) \cdot \alpha
\end{equation}
where $d_t(x)$ is the minimal normalized   distance, and $\alpha$ is the learning rate. For dynamic background region pixels, $D_{min}(x)$ trends to the value 1, and for static background regions, $D_{min}(x)$ trends to 0.
Then, a pixel-level accumulator $v$ is defined to monitor blinking  pixels:
\begin{equation}\label{vx}
v(x) = \begin{cases}
v(x) + v_{incr}, & \text{if }  X_t(x) = 1 \\
v(x) - v_{decr}, & \text{otherwise. }
\end{cases}
\end{equation}
where $v_{incr}$ and $v_{decr}$ are two fixed parameters with the value of 1 and 0.1, respectively.
$X_t(x)$ is the blinking pixel map calculated by an XOR operation between $B_t(x)$ and $B_{t-1}(x)$.
With $v(x)$ and $D_{min}(x)$ defined, the distance threshold $R(x)$ can be recursively adjusted as follows:
\begin{equation}\label{R(x)}
R(x) = \begin{cases}
R(x) + v(x), & \text{if }  R(x) < ( 1 + D_{min}(x)\cdot 2) ^2 \\
R(x) - \frac{1}{v(x)}, & \text{otherwise }
\end{cases}.
\end{equation}

The background update rate parameter $T$ is used to control the speed of the background absorption.
The randomly-picked background samples in $B(x)$ have the probability of $1/T(x)$ to be replaced by $I_t(x)$, if current pixel $x$ belongs to the background. The lower the $T(x)$ is, the higher the update probability, and vice versa.
$T(x)$ is also recursively adjusted by $D_{min}(x)$ and $v(x)$. More specifically, $T(x)$ is defined as follows:
\begin{equation}\label{T(x)}
T(x) = \begin{cases}
T(x) + \frac{1}{v(x)\cdot D_{min}(x)}, & \text{if } B_t(x) = 1 \\
T(x) - \frac{v(x)}{D_{min}(x)},        & \text{if } B_t(x) = 0
\end{cases}.
\end{equation}
The authors also define a lower bound $T_{lower}$ and an upper bound $T_{upper}$ to clamp $T(x)$ to the interval $[T_{lower}, T_{upper}]$.
%From Equation (\ref{T(x)}) we can see that:
%When pixels are classified as foreground with $S_t(x) = 1$, the update rate will increases rapidly if pixels are in a region with very little entropy. Additionally, if pixels are in an unstable region, the background model will update quickly to keep adapting to new observations.

\subsection{ICNet for Real-time Semantic Segmentation}\label{ICNetSeg}
%ICNet~\cite{zhao2018icnet}
We adopt ICNet~\cite{zhao2018icnet} to develop the benchmark semantic segmenter $\mathcal{S}$. The ICNet  achieved an excellent trade-off   between  efficiency and accuracy   for  real-time semantic segmentation. To reduce network computational complexity, a multi-resolution cascade network architecture was proposed. Its core idea is to let the low-resolution image go through the full semantic network first for a coarse segmentation map. Then, a cascade fusion unit introduces  middle- and high-resolution image feature to help improving the coarse  segmentation map gradually.

In order to extract object-level potential foreground semantic information for the Change Detection dataset sequences~\cite{wang2014cdnet}, we first train an ICNet model for semantic segmentation with the  ADE20K dataset~\cite{zhou2017scene}.
%We first train an  ICNet model for semantic segmentation with the  ADE20K dataset~\cite{zhou2017scene}. Then, the trained model is used  to extract object-level potential foreground semantic information on the Change Detection dataset sequences~\cite{wang2014cdnet}.
The ADE20K dataset is a recently released benchmark for scene parsing. It contains 20K/2K/3K high quality pixel-level annotations images for training, validation and testing. The pixel annotations span 150 classes (\textit{e.g.}, person, car, and tree) which frequently occur in diverse scenes.
Therefore, it covers a large number of object categories and scene distributions.
Here, we define $C = \{c_1, c_2, \ldots, c_{N} \}$ to be the set of object classes.

\begin{figure}[ht]
\centering
   \includegraphics[width=1.0\linewidth]{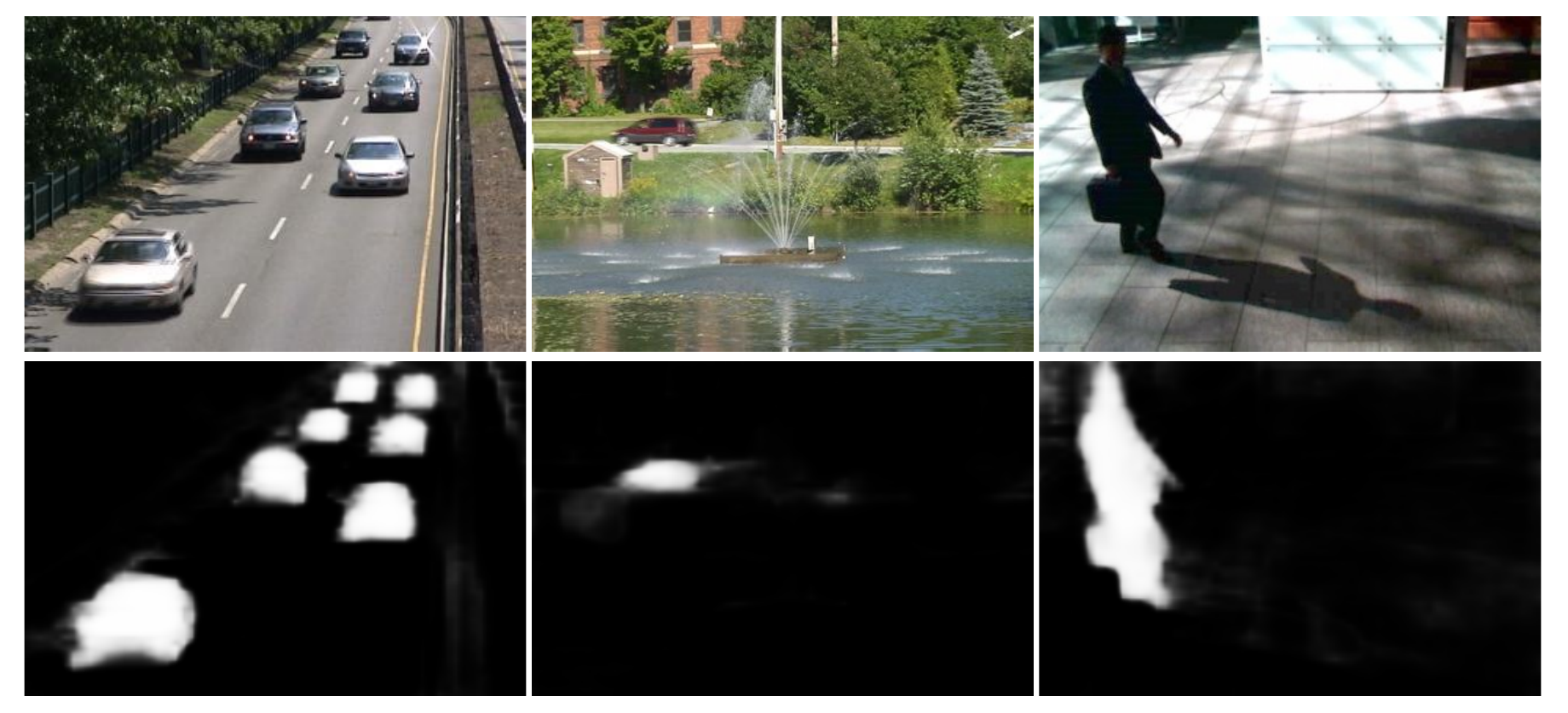}
   \caption{ICNet semantic foreground segmentation results. The first row shows  original images  and the second row shows their corresponding segmentation result. From left to right: \textit{highway, fountain02, peopleInShade}.}
\label{EFRNetresult}
\end{figure}

After the training process,   the trained model is executed on the Change Detection dataset sequences. However, before feeding a frame through the network, we need to make some padding around the image so that it  can adapt  to the input structure of the network  since  sequences from the Change Detection  dataset have a variety of sizes.
After the forward pass,  the last  layer of the model outputs a real value in each pixel for each of the object classes.
We denote the real value vector of pixel $x$ at $t$-th frame for all classes as: $v_t(x) = [v_t^1(x), v_t^2(x), \cdots, v_t^{N}(x)]$, where $v_t^i(x)$ is the predict score for class $c_i$. Then, a softmax function is applied on $v_t(x)$ to get the probability vector $p_t(x) = [p_t^1(x), p_t^2(x), \cdots, p_t^{N}(x)] $  with $p_t^i(x)$ denotes the probability for class $c_i$.
However, since we want to get potential foreground object information for background subtraction problems, only a subset classes from the 150 labels are relevant.
The same with~\cite{braham2017semantic}, we choose the semantic relevant foreground classes as: $F = $  \{person, car, cushion, box, book, boat, bus, truck,
bottle, van, bag, bicycle\} $F \subset C$, which are the most
frequent foreground objects appeared in the Change Detection dataset.
Finally, we compute the semantic foreground probability map $S_t(x)$ as follows (mapping to 0$-$255):
\begin{equation}\label{pstx}
S_t(x)   =  255\cdot\sum_{c_i \in F} p_t^{ c_i}(x)   %= p_t(x \in R)
\end{equation}
Fig.~\ref{EFRNetresult} shows some semantic foreground segmentation results on the Change Detection dataset sequences.

\subsection{RTSS Framework}\label{sectionRTSSframework}

\begin{figure*}[ht]
\centering
   \includegraphics[width=1.0\linewidth]{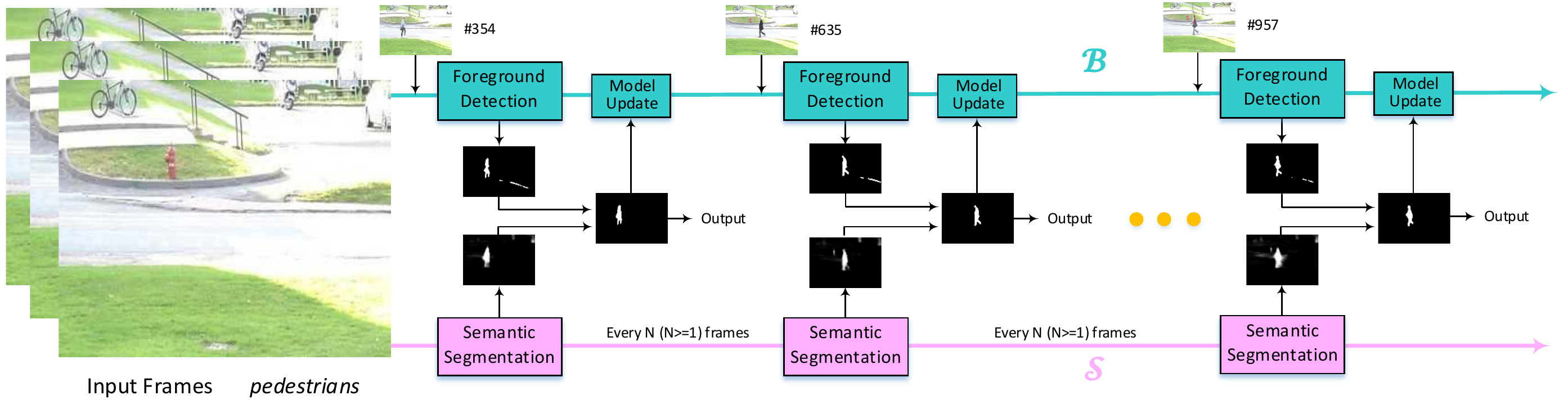}  %RTSSframework
   \caption{Illustration of the RTSS framework in which the BGS segmenter $\mathcal{B}$ and the semantic segmenter $\mathcal{S}$ are processed in two parallel synchronous threads.}
\label{RTSSframework}
\end{figure*}

After introducing the benchmark BGS segmenter $\mathcal{B}$ and the semantic segmenter $\mathcal{S}$,  we will give a detailed description of how  $\mathcal{B}$ and $\mathcal{S}$ cooperate together for real-time and accuracy foreground detection.  Fig.~\ref{RTSSframework} illustrates the proposed RTSS (\textit{background subtraction with real-time semantic segmentation}) framework.

In RTSS, $\mathcal{B}$ and $\mathcal{S}$  run in parallel on two synchronous threads while collaborating  with each other.
The BGS segmenter $\mathcal{B}$ is mainly responsible for the background subtraction process and the semantic segmenter $\mathcal{S}$ is employed to detect the object-level potential foreground object(s).
For each input frame, in the  BGS segmenter thread, $\mathcal{B}$ performs the BG/FG classification process. At the same time, in the semantic segmenter thread, $\mathcal{S}$ makes the semantic segmentation process.
As we mentioned before,  in order to obtain real-time semantic information, we have two strategies to choose from. The first one is to use real-time semantic segmentation algorithms such as ICNet~\cite{zhao2018icnet}, ERFNet~\cite{romera2018erfnet}, LinkNet~\cite{chaurasia2017linknet}, and so on. The second one is to  present the semantic segmentation operation once every N (N $>$ 1) frames if the semantic segmentation algorithms cannot run in real time.
The reason behind this is that  in most cases, two consecutive frames in a video are highly redundant, the semantic segmentation results in two adjacent frames may be very similar,  the  extracted  potential foreground objects may only have  small displacements.
In Fig.~\ref{RTSSframework}, for the   semantic segmenter  $\mathcal{S}$, if a real-time  semantic segmentation algorithm is used, then, N is set to 1. That is to say, every input frame is semantically segmented.
If a non-real-time semantic segmentation algorithm is used, then, N is set to a number greater than 1.  That is to say,  we do the semantic segmentation every N frames. More specifically,   suppose the current frame  $I_t$ is the t-th input frame, and semantic segmentation  is performed to get the semantic foreground probability map  $S_{t}$. However,  we will then skip  the    semantic segmentation operation on  the $I_{t+1}, I_{t+2}, \ldots, I_{t+N-1}$ frames, but use    $S_{t}$  as an alternative semantic foreground probability map  of these  frames.
In this way, the computational time of the semantic segmenter $\mathcal{S}$ can be greatly reduced.

The collaboration process between $\mathcal{B}$ and $\mathcal{S}$ can be summarized as Algorithm~\ref{RTSSframeworkAlrorithm}.
When an input frame $I_t$ is available,  in the $\mathcal{B}$  thread, it runs the BGS algorithm to do the foreground detection first, and gets a preliminary foreground/background (FG/BG) mask $B_t$.
At the same time, in the semantic segmenter thread  $\mathcal{S}$, it forwards $I_t$ to the trained  model and uses the output  result to calculate the semantic foreground probability map ${S}_t$, then, trigger a signal that ${S}_t$ is available. On the other side, in the  $\mathcal{B}$ thread, after the foreground detection process is finished, it will wait until   ${S}_t$ is available. Once ${S}_t$ is available, the  BGS segmenter will combine  ${B}_t$ and ${S}_t$ to produce a  more precise FG/BG mask ${D}_t$. Finally, the refined FG/BG mask is fed back to update the background model.
Go to the next frame and repeat the process until all the frames are processed.

%Algorithm~\ref{RTSSframeworkAlrorithm} summarizes the process of the RTSS framework.
\begin{algorithm}[htbp]
\caption{: The RTSS framework process.}
\begin{algorithmic}[1]
\STATE Initialize the BGS segmenter thread for $\mathcal{B}$.
\STATE Initialize the semantic segmenter thread for $\mathcal{S}$.
\STATE Run $\mathcal{B}$   and $\mathcal{S}$   till the end of sequence.

-----------------------------------------------------------------------
$\mathcal{B}$:\\
\textbf{while} \textit{current frame $I_t$ is valid} \textbf{do} \\
\qquad {BG/FG classification and output the result $B_t$}\\
\qquad \textbf{while} \textit{semantic result $S_t$ is valid} \textbf{do}\\
\qquad \qquad get $S_t$ from $\mathcal{S}$ \\  %and trigger a signal to  $\mathcal{S}$
\qquad \qquad combine  $B_t$ and $S_t$ to get  the final result $D_t$\\
\qquad \qquad use $D_t$ to  update the background model\\
\qquad \textbf{end}\\
\qquad {current frame $\leftarrow$  next frame}\\
\textbf{end}\\

-----------------------------------------------------------------------
$\mathcal{S}$:\\
\textbf{while} \textit{current frame $I_t$ is valid} \textbf{do}\\
%\qquad \textbf{while} \textit{receive a signal from $\mathcal{B}$} \textbf{do}\\
\qquad  semantic segmentation and output the result $S_t$\\
\qquad  trigger a signal  $S_t$ is available\\
%\qquad \textbf{end}\\
\qquad {current frame $\leftarrow$  next frame}\\
\textbf{end}\\

\end{algorithmic}
\label{RTSSframeworkAlrorithm}
\end{algorithm}

\begin{figure}[t]
\centering
   \includegraphics[width=1.0\linewidth]{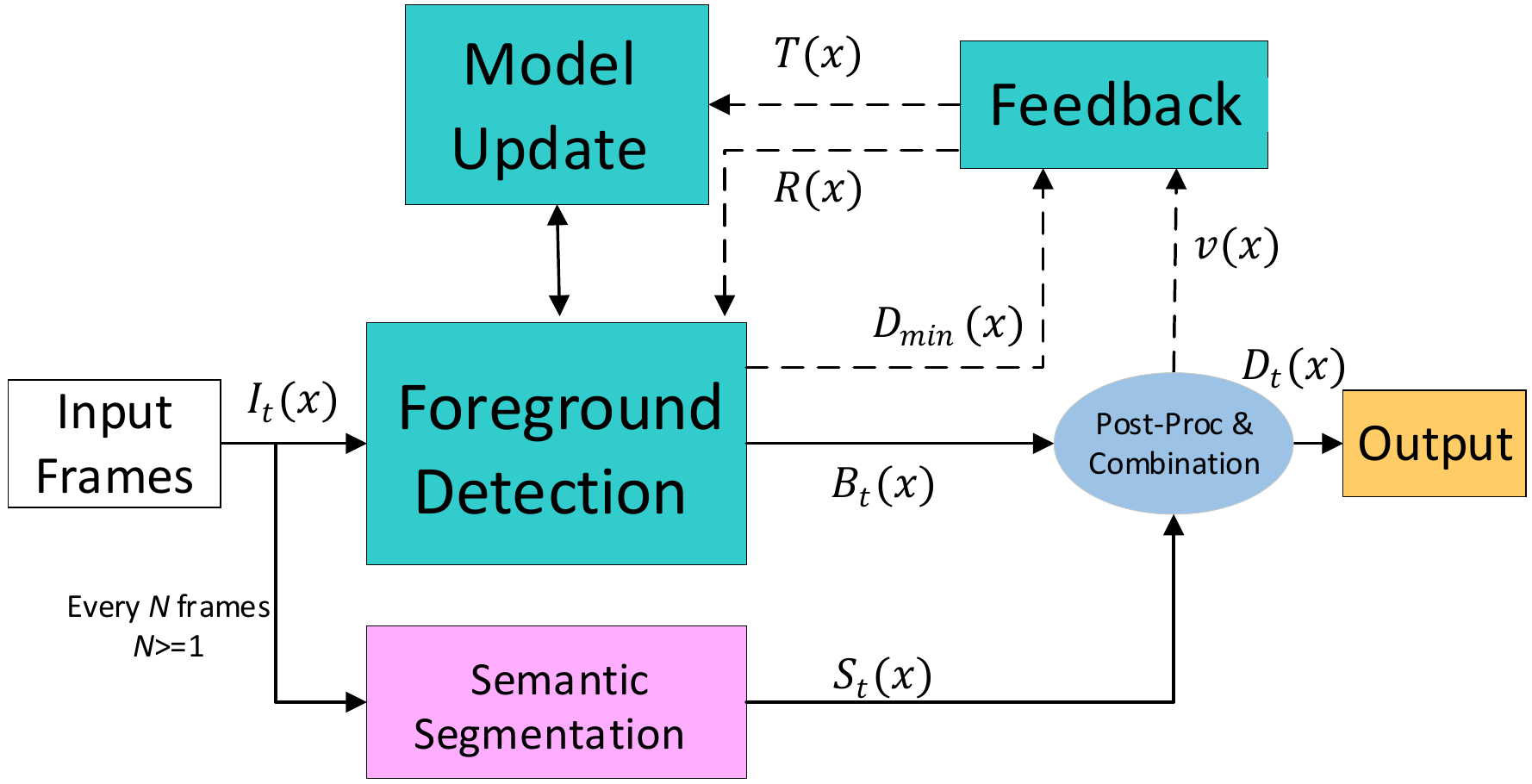}
   \caption{Schematic diagram of the interaction  between the SuBSENSE~\cite{st2015subsense} algorithm and semantic segmentation algorithm.}
\label{SuBSENSEdiagram}
\end{figure}

%An overview of the SuBSENSE algorithm and its  interaction with semantic segmentation algorithm is presented in Fig.~\ref{SuBSENSEdiagram}.
Fig.~\ref{SuBSENSEdiagram} presents the interaction process between the SuBSENSE~\cite{st2015subsense} algorithm and semantic segmentation algorithm. $I_t(x)$ is the current input frame, $T(x)$, $R(x)$, $D_{min}(x)$, and $v(x)$ are internal parameters of SuBSENSE as shown in \ref{SubSENSEBGS}. $B_t(x)$ and $S_t(x)$ are the foreground detection  and semantic segmentation results of current frame, respectively. After  $B_t(x)$ and  $S_t(x)$ are obtained, some specific rules are applied on $B_t(x)$ and $S_t(x)$ to get a more precise FG/BG mask $D_t(x)$. At the same time, the refined FG/BG mask is fed back into the background subtraction process and  guide the background model updating, which
produces a more reliable   background model.  As time goes by, the accuracy of foreground detection result in SuBSENSE can be greatly improved, and finally made the final  result $D(x)$ better and better.

Now, the remaining key problem  is   how can we make use of the semantic  foreground segmentation result $S_t$ to compensate for the errors of the background subtraction result $B_t$ and to produce a more accurate result $D_t$. In this work, we use a similar strategy presented in~\cite{braham2017semantic}.
%Now, the remaining key problem  is   how can we combine the semantic  foreground segmentation result $S_t$ to compensate for the errors of the background subtraction result $B_t$ and to produce a more accurate result $D_t$. In this work, we use a strategy similar to~\cite{braham2017semantic}.
First, we want to extract two useful information from the semantic foreground segmentation $S_t$. One is to point out which pixels have high confidence that belong to the background. The other one shows which pixels have a higher probability of belonging to foreground. We denote them as $S_t^{BG}$ and $S_t^{FG}$ respectively. In this work, we define $S_t^{BG}(x)$ as follows:
\begin{equation}\label{stbg}
S_t^{BG}(x) = S_{t}(x).
\end{equation}
While for $S_t^{FG}(x)$, it is defined as follows:
\begin{equation}\label{stfg}
  S_t^{FG}(x) = S_{t}(x) - M_t(x)
\end{equation}
where $M_t(x)$ is regarded as a semantic background model for pixel $x$ at time $t$.
Firstly, $M_t(x)$ is initialized in the first frame with $M_0(x) = S_{0}(x)$.
Then,  to keep the adaptation of the model to the scene changes, a conservative random updating strategy as introduced in \cite{barnich2011vibe} is  adopted to update $M_t(x)$. More specifically, if $D_t(x)$ is classified as FG, $M_t(x)$ remains unchanged; if $D_t(x)$ is classified as BG, $M_t(x)$ has a probability of $1/\phi$ to be updated by $S_{t}(x)$, where $\phi$  is a time subsampling factor which controls the adaptation speed.
This conservative updating strategy can avoid model corruption due to intermittent and slow moving foreground objects.
This process can be summarized by Algorithm~\ref{mtxupdating}:
\begin{algorithm}[htbp]
\caption{: $M_t(x)$ updating process.} %
\begin{algorithmic}[1]
\STATE  Initialize $M_t(x)$ with $M_0(x)   =   S_{0}(x)$
\STATE \textbf{for} $t \geq 0$
\STATE  \quad \textbf{if} $D_t(x)$ = FG
\STATE  \quad \quad  $M_{t+1}(x) = M_t(x)$;
\STATE  \quad \textbf{if} $D_t(x)$ = BG
\STATE  \quad \quad \textbf{if} $rand()$  \%  $\phi$ = 0
\STATE  \quad \quad \quad $M_{t+1}(x) = S_{t}(x)$;
\STATE  \quad \quad \textbf{else}
\STATE  \quad \quad \quad $M_{t+1}(x) = M_t(x)$;
\STATE \textbf{end for}
\end{algorithmic}
\label{mtxupdating}
\end{algorithm}

\begin{figure}[!ht]
\centering
\includegraphics[width=1.0\linewidth]{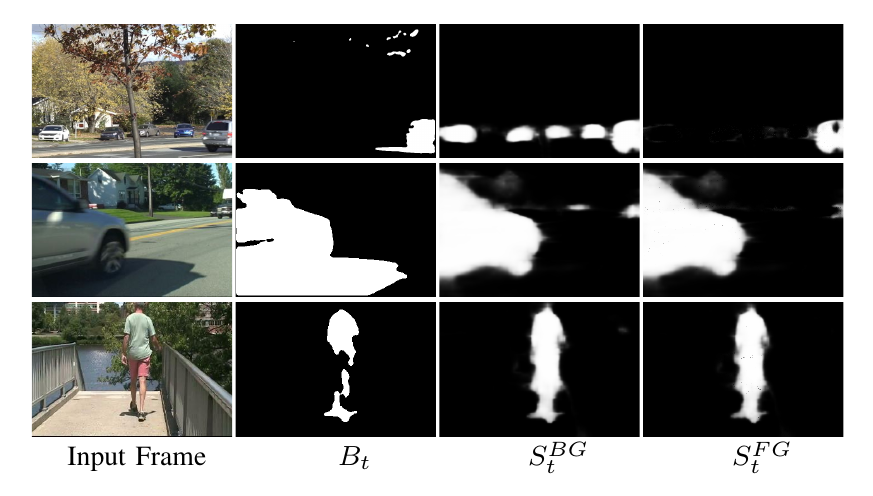}
\vspace{-0.08\linewidth}
\caption{Illustration of  background subtraction result $B_t$, semantic segmentation result $S_t^{BG}$ and $S_t^{FG}$ on some sequences. From top to down: \textit{fall, bungalows, overpass.}}
\label{BtStBgStFg}
\end{figure}

Fig.~\ref{BtStBgStFg} shows some examples of  background subtraction result $B_t$, and semantic segmentation results $S_t^{BG}$ and $S_t^{FG}$ on the Change Detection dataset sequences.
We now need to define some rules for combining $B_t$, $S_t^{BG}$  and $S_t^{FG}$  to get $D_t$.

Firstly, we specify that pixels with a low semantic foreground probability in $S_t^{BG}$ should be classified as  background without considering  $B_t$, as shown as follows:
\begin{equation}\label{rule1}
  \text{If } S_t^{BG} \leq \tau_{BG} \text{, then }  D_t(x) = \text{BG}
\end{equation}
where $\tau_{BG}$ is the  threshold value for background.
As shown in Fig.~\ref{BtStBgStFg}, the BGS segmenter produces  many false positive pixels due to dynamic backgrounds, illumination variations and shadows, which
severely affect the  accuracy of the  foreground detection result. However, rule~(\ref{rule1}) provides a simple way to address these challenges.

Secondly, pixels with a high semantic foreground probability in $S_t^{FG}$ should be classified as  foreground, as shown as follows:
\begin{equation}\label{rule2}
  \text{If } S_t^{FG} \geq \tau_{FG} \text{, then }   D_t(x) = \text{FG}
\end{equation}
where $\tau_{FG}$ denotes the  threshold value for the foreground.
rule~(\ref{rule2}) is mainly focused on correcting false negative detection pixels. In Fig.~\ref{BtStBgStFg}, we can see that when foreground samples and background model share similar features due to camouflage or model absorption,  many ``holes'' appear  in foreground object, while $S_t^{FG}$ can be used to  compensate and correct these errors.

Finally,  if all the previous conditions are not satisfied, this means that the semantic segmentation result cannot provide enough information to make a decision. At this time, we directly take the BGS algorithm result as the final result with $D_t(x) = B_t(x)$.
To sum up, the complete $B_t$, $S_t^{BG}$  and $S_t^{FG}$ combination process can be summarized with Equation (\ref{Dtx}):
\begin{equation} \label{Dtx}
D_t(x) = \begin{cases}
FG,     & \text{if }  S_t^{FG}(x) \geq \tau_{FG}; \\
BG,     & \text{else if }  S_t^{BG}(x) \leq \tau_{BG}; \\
B_t(x), & \text{otherwise. }    \\
\end{cases}
\end{equation}

\section{Experimental Results}\label{Experimental results}

\begin{figure*}[t]
\begin{center}
\subfloat{
\includegraphics[width=0.33\linewidth]{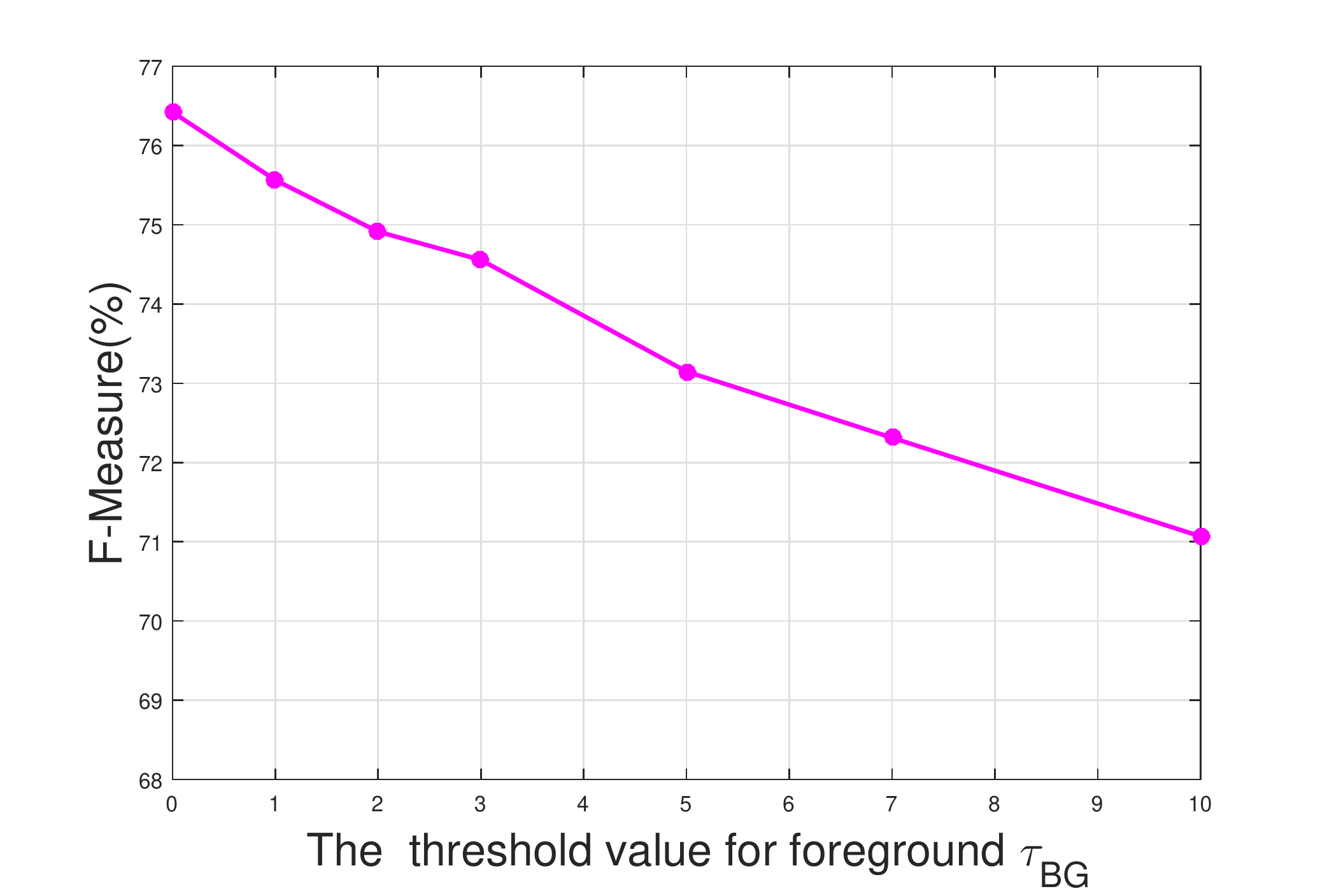}
\includegraphics[width=0.33\linewidth]{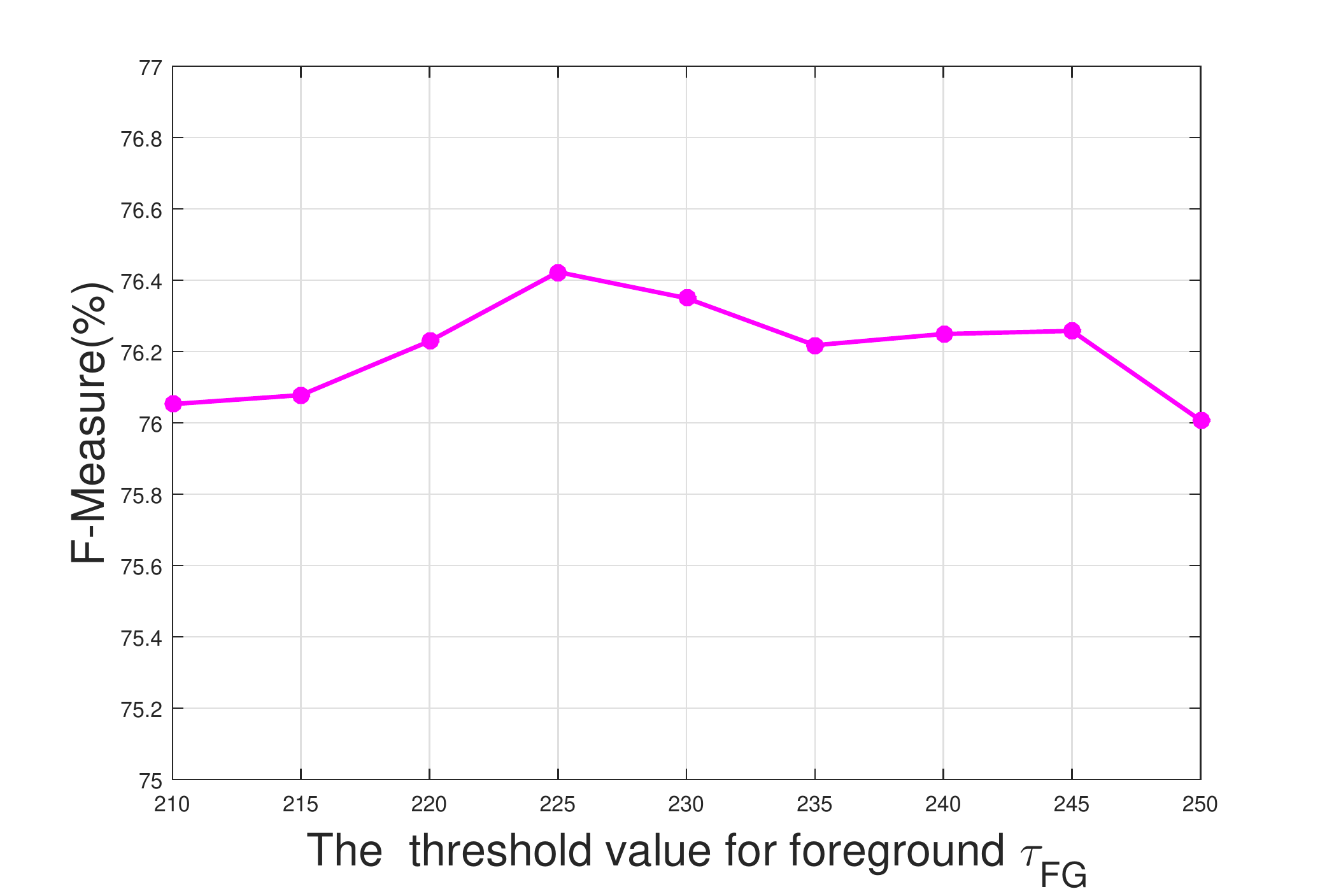}
\includegraphics[width=0.33\linewidth]{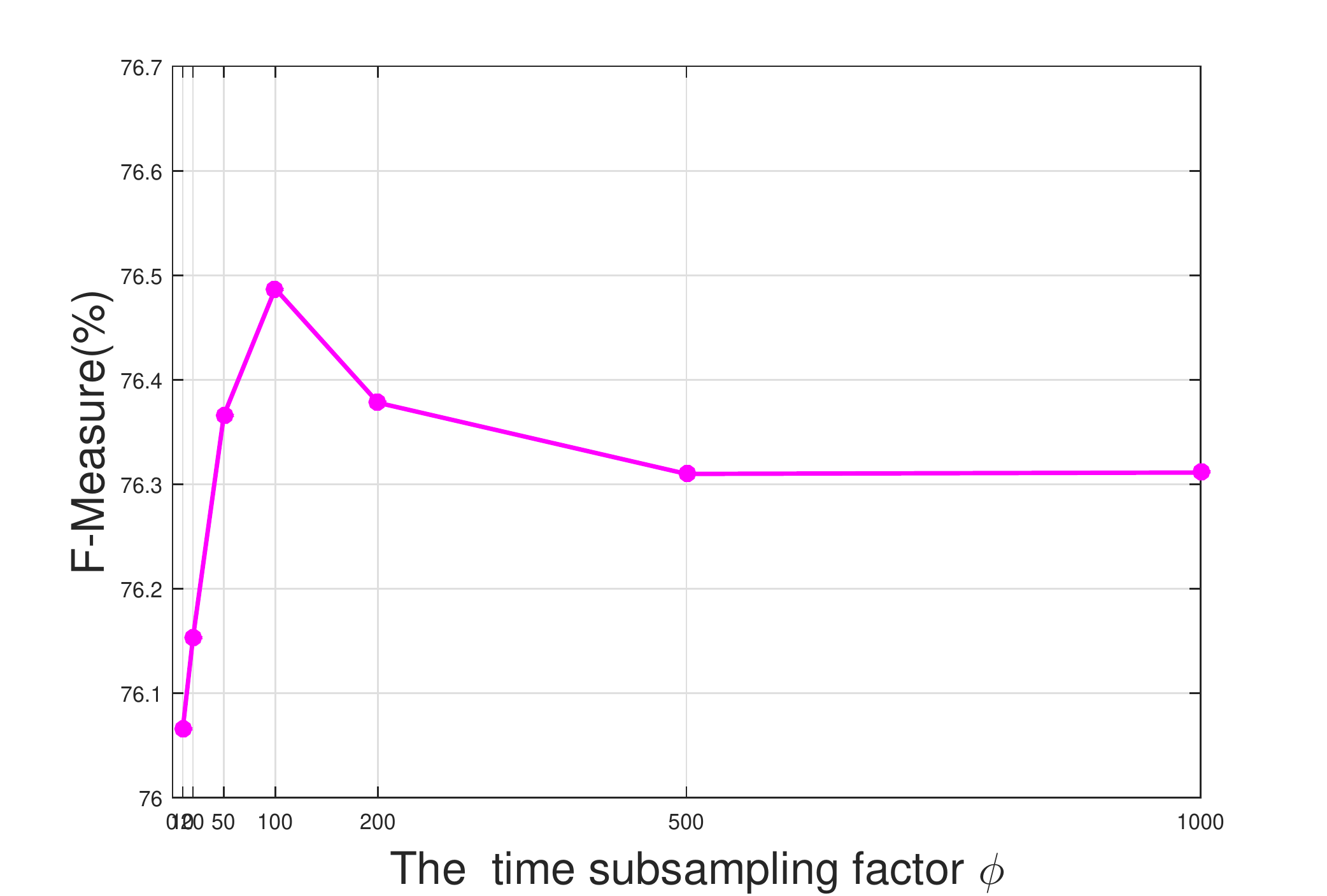}
}
\caption{The relationship between  different parameter values   and F-Measure scores on the CDnet 2014 dataset.}
\label{parameters}
\end{center}
\end{figure*}

In this section,  we first give a brief introduction of  evaluation dataset  and metrics. Then  we provide both quantitative and qualitative comparisons with state-of-the-art background subtraction  methods. We also perform an ablation study of the RTSS framework.

\subsection{{Evaluation Dataset and Metrics}}\label{EvaluationDataset}

We evaluate the performance of the proposed  RTSS framework for background subtraction using the Change Detection (CDnet) 2014 dataset~\cite{wang2014cdnet}.
The CDnet 2014 dataset is a region-level benchmark obtained from realistic scenarios.  Accurate human experts constructed ground truths are available for all sequences.
This dataset contains 53 video sequences which are grouped into 11 different video categories comprising
{``baseline'', ``camera jitter'', ``dynamic background'',  ``intermittent object motion'', ``shadow'', ``thermal'', ``bad weather'', ``low framerate'', ``night videos'', ``pan-tilt-zoom'' } and {``turbulence''}. The resolution of the sequences varies from $240\times320$ to $576\times720$ with hundred to thousand frames.
To our knowledge, this is one of the most comprehensive datasets for background subtraction evaluations.

To make an  exhaustive quantitative comparison between different background subtraction methods,  seven different quantitative metrics have been defined in~\cite{wang2014cdnet}: Recall (\textit{Re}), Specificity (\textit{Sp}), False positive rate (\textit{FPR}), False negative rate (\textit{FNR}), Percentage of wrong classifications (\textit{PWC}), Precision (\textit{Pr}), F-Measure (\textit{FM}).
All these metric values  range from 0 to 1.  For \textit{Re, Sp, Pr} and \textit{FM} metrics, higher values indicate more accuracy while for \textit{PWC, FNR} and \textit{FPR} metrics, lower values indicate better performance.
All the measured metrics depend on \textit{TP} (True Positives), \textit{TN} (True Negatives), \textit{FP} (False Positives), and \textit{FN} (False Negatives).  \textit{TP}  is the correctly detected foreground pixels,  \textit{TN} is the correctly detected background pixels,  \textit{FP} is the number of background pixels detected as foreground and \textit{FN} is the number of foreground pixels detected as background.\\

\textbullet \quad Recall (\textit{Re}) = $\frac{TP}{TP+FN}$

\textbullet \quad Specificity (\textit{Sp}) = $\frac{TN}{TN+FP}$

\textbullet \quad False positive rate (\textit{FPR}) = $\frac{FP}{FP+TN}$

\textbullet \quad False negative rate (\textit{FNR}) = $\frac{FN}{TP+FN}$

\textbullet \quad Percentage of wrong classifications (\textit{PWC}) = $100 \cdot\frac{FN+FP}{TP+FN+FP+TN}$

\textbullet \quad Precision (\textit{Pr}) = $\frac{TP}{TP+FP}$

\textbullet \quad F-Measure (\textit{FM}) = $2\cdot\frac{Re\cdot Pr}{Re+Pr}$   \\

Generally speaking, a background subtraction method is considered good if it gets high recall scores, without sacrificing precision. Therefore, the F-Measure (\textit{FM}) metric which  represents a balance between the {Recall} and {Precision}, is mainly  accepted as a single indicator of the overall performance of  different BGS methods.

\subsection{{Parameter Setting}}\label{ParametersSetting}

As discussed above, the proposed method  contains a few parameters which should be determined before building a complete model:        the  threshold  for background $\tau_{BG}$,  the  threshold  for foreground $\tau_{FG}$ and the time subsampling factor  $\phi$.
The performance with different parameter settings are show in Fig.~\ref{parameters}.
Although we can obtain optimal performance by adjusting parameters in different scenarios, we used a universal parameter set for all sequences to respect the  competition rules of the Change Detection dataset~\cite{wang2014cdnet}.

\begin{itemize}
  \item $\tau_{BG} = 0$:  the  threshold value   in Equation~(\ref{rule1}).
  \item $\tau_{FG} = 225$:  the  threshold value   in Equation~(\ref{rule2}).
  \item $\phi = 100$: the time subsampling factor for controlling the background model   adaptation speed.
\end{itemize}

%Subsense+icnet
\begin{table*}[!ht]
\centering
\begin{threeparttable}
\caption{Complete results of the RTSS framework with \textcolor[rgb]{0,0,1}{SuBSENSE} and \textcolor[rgb]{1,0,1}{ICNet}   on the CDnet 2014 dataset\tnote{1}}
{\tabcolsep3pt\begin{tabular}{cccccccc}
\toprule
Category        &Recall         &Specificity        &FPR            &FNR            &PWC            &Precision           &F-Measure\\
\midrule
baseline
&\textcolor[rgb]{0,0,1}{0.9519}(\textcolor[rgb]{1,0,1}{0.9566} $\Uparrow$)       &\textcolor[rgb]{0,0,1}{0.9982}(\textcolor[rgb]{1,0,1}{0.9946} $\Downarrow$)      &\textcolor[rgb]{0,0,1}{0.0018}(\textcolor[rgb]{1,0,1}{0.0054} $\Downarrow$)     &\textcolor[rgb]{0,0,1}{0.0481}(\textcolor[rgb]{1,0,1}{0.0434} \textcolor[rgb]{1,0,0}{$\Uparrow$})         &\textcolor[rgb]{0,0,1}{0.3639}(\textcolor[rgb]{1,0,1}{0.6310} $\Downarrow$)
&\textcolor[rgb]{0,0,1}{0.9486}(\textcolor[rgb]{1,0,1}{0.9093} $\Downarrow$)     &\textcolor[rgb]{0,0,1}{0.9498}(\textcolor[rgb]{1,0,1}{0.9312} $\Downarrow 2\%$)\\

cameraJ
&\textcolor[rgb]{0,0,1}{0.8319}(\textcolor[rgb]{1,0,1}{0.8131} $\Downarrow$)     &\textcolor[rgb]{0,0,1}{0.9901}(\textcolor[rgb]{1,0,1}{0.9914} \textcolor[rgb]{1,0,0}{$\Uparrow$})      &\textcolor[rgb]{0,0,1}{0.0099}(\textcolor[rgb]{1,0,1}{0.0086} \textcolor[rgb]{1,0,0}{$\Uparrow$})       &\textcolor[rgb]{0,0,1}{0.1681}(\textcolor[rgb]{1,0,1}{0.1869} $\Downarrow$)         &\textcolor[rgb]{0,0,1}{1.6937}(\textcolor[rgb]{1,0,1}{1.6788} \textcolor[rgb]{1,0,0}{$\Uparrow$})
&\textcolor[rgb]{0,0,1}{0.7944}(\textcolor[rgb]{1,0,1}{0.8193} \textcolor[rgb]{1,0,0}{$\Uparrow$})       &\textcolor[rgb]{0,0,1}{0.8096}(\textcolor[rgb]{1,0,1}{0.7995} $\Downarrow 1\%$)\\

dynamic
&\textcolor[rgb]{0,0,1}{0.7739}(\textcolor[rgb]{1,0,1}{0.9191} \textcolor[rgb]{1,0,0}{$\Uparrow$})       &\textcolor[rgb]{0,0,1}{0.9994}(\textcolor[rgb]{1,0,1}{0.9992} $\Downarrow$)      &\textcolor[rgb]{0,0,1}{0.0006}(\textcolor[rgb]{1,0,1}{0.0008} $\Downarrow$)     &\textcolor[rgb]{0,0,1}{0.2261}(\textcolor[rgb]{1,0,1}{0.0809} \textcolor[rgb]{1,0,0}{$\Uparrow$})         &\textcolor[rgb]{0,0,1}{0.4094}(\textcolor[rgb]{1,0,1}{0.1544} \textcolor[rgb]{1,0,0}{$\Uparrow$})
&\textcolor[rgb]{0,0,1}{0.8913}(\textcolor[rgb]{1,0,1}{0.8980} \textcolor[rgb]{1,0,0}{$\Uparrow$})       &\textcolor[rgb]{0,0,1}{0.8159}(\textcolor[rgb]{1,0,1}{0.9053} \textcolor[rgb]{1,0,0}{$\Uparrow11\%$})\\

intermittent
&\textcolor[rgb]{0,0,1}{0.5715}(\textcolor[rgb]{1,0,1}{0.7354} \textcolor[rgb]{1,0,0}{$\Uparrow$})            &\textcolor[rgb]{0,0,1}{0.9953}(\textcolor[rgb]{1,0,1}{0.9963} \textcolor[rgb]{1,0,0}{$\Uparrow$})
&\textcolor[rgb]{0,0,1}{0.0047}(\textcolor[rgb]{1,0,1}{0.0037} \textcolor[rgb]{1,0,0}{$\Uparrow$})            &\textcolor[rgb]{0,0,1}{0.4285}(\textcolor[rgb]{1,0,1}{0.2646} \textcolor[rgb]{1,0,0}{$\Uparrow$})         &\textcolor[rgb]{0,0,1}{4.0811}(\textcolor[rgb]{1,0,1}{3.0055} \textcolor[rgb]{1,0,0}{$\Uparrow$})
&\textcolor[rgb]{0,0,1}{0.8174}(\textcolor[rgb]{1,0,1}{0.8841} \textcolor[rgb]{1,0,0}{$\Uparrow$})            &\textcolor[rgb]{0,0,1}{0.6068}(\textcolor[rgb]{1,0,1}{0.7802} \textcolor[rgb]{1,0,0}{$\Uparrow29\%$})\\

shadow
&\textcolor[rgb]{0,0,1}{0.9441}(\textcolor[rgb]{1,0,1}{0.9590} \textcolor[rgb]{1,0,0}{$\Uparrow$})            &\textcolor[rgb]{0,0,1}{0.9920}(\textcolor[rgb]{1,0,1}{0.9942} \textcolor[rgb]{1,0,0}{$\Uparrow$})
&\textcolor[rgb]{0,0,1}{0.0080}(\textcolor[rgb]{1,0,1}{0.0058} \textcolor[rgb]{1,0,0}{$\Uparrow$})            &\textcolor[rgb]{0,0,1}{0.0559}(\textcolor[rgb]{1,0,1}{0.0410} \textcolor[rgb]{1,0,0}{$\Uparrow$})         &\textcolor[rgb]{0,0,1}{1.0018}(\textcolor[rgb]{1,0,1}{0.7243} \textcolor[rgb]{1,0,0}{$\Uparrow$})
&\textcolor[rgb]{0,0,1}{0.8645}(\textcolor[rgb]{1,0,1}{0.8938} \textcolor[rgb]{1,0,0}{$\Uparrow$})            &\textcolor[rgb]{0,0,1}{0.8995}(\textcolor[rgb]{1,0,1}{0.9242} \textcolor[rgb]{1,0,0}{$\Uparrow3\%$})\\

thermal
&\textcolor[rgb]{0,0,1}{0.8166}(\textcolor[rgb]{1,0,1}{0.6492} $\Downarrow$)            &\textcolor[rgb]{0,0,1}{0.9910}(\textcolor[rgb]{1,0,1}{0.9903} $\Downarrow$)
&\textcolor[rgb]{0,0,1}{0.0090}(\textcolor[rgb]{1,0,1}{0.0097} $\Downarrow$)            &\textcolor[rgb]{0,0,1}{0.1834}(\textcolor[rgb]{1,0,1}{0.3508} $\Downarrow$)         &\textcolor[rgb]{0,0,1}{1.9632}(\textcolor[rgb]{1,0,1}{2.4733} $\Downarrow$)
&\textcolor[rgb]{0,0,1}{0.8319}(\textcolor[rgb]{1,0,1}{0.8626} \textcolor[rgb]{1,0,0}{$\Uparrow$})              &\textcolor[rgb]{0,0,1}{0.8172}(\textcolor[rgb]{1,0,1}{0.7007} $\Downarrow14\%$)\\
\midrule
badWeather
&\textcolor[rgb]{0,0,1}{0.8082}(\textcolor[rgb]{1,0,1}{0.7442} $\Downarrow$)            &\textcolor[rgb]{0,0,1}{0.9991}(\textcolor[rgb]{1,0,1}{0.9993} \textcolor[rgb]{1,0,0}{$\Uparrow$})
&\textcolor[rgb]{0,0,1}{0.0009}(\textcolor[rgb]{1,0,1}{0.0007} \textcolor[rgb]{1,0,0}{$\Uparrow$})              &\textcolor[rgb]{0,0,1}{0.1918}(\textcolor[rgb]{1,0,1}{0.2558} $\Downarrow$)         &\textcolor[rgb]{0,0,1}{0.4807}(\textcolor[rgb]{1,0,1}{0.5061} $\Downarrow$)
&\textcolor[rgb]{0,0,1}{0.9183}(\textcolor[rgb]{1,0,1}{0.9504} \textcolor[rgb]{1,0,0}{$\Uparrow$})              &\textcolor[rgb]{0,0,1}{0.8577}(\textcolor[rgb]{1,0,1}{0.8196} $\Downarrow4\%$)\\

lowFramerate
&\textcolor[rgb]{0,0,1}{0.8267}(\textcolor[rgb]{1,0,1}{0.8050} $\Downarrow$)            &\textcolor[rgb]{0,0,1}{0.9937}(\textcolor[rgb]{1,0,1}{0.9952} \textcolor[rgb]{1,0,0}{$\Uparrow$})
&\textcolor[rgb]{0,0,1}{0.0063}(\textcolor[rgb]{1,0,1}{0.0048} \textcolor[rgb]{1,0,0}{$\Uparrow$})              &\textcolor[rgb]{0,0,1}{0.1733}(\textcolor[rgb]{1,0,1}{0.1950} $\Downarrow$)         &\textcolor[rgb]{0,0,1}{1.2054}(\textcolor[rgb]{1,0,1}{1.1464} \textcolor[rgb]{1,0,0}{$\Uparrow$})
&\textcolor[rgb]{0,0,1}{0.6475}(\textcolor[rgb]{1,0,1}{0.6787} \textcolor[rgb]{1,0,0}{$\Uparrow$})              &\textcolor[rgb]{0,0,1}{0.6659}(\textcolor[rgb]{1,0,1}{0.6849} \textcolor[rgb]{1,0,0}{$\Uparrow3\%$})\\

nightVideos
&\textcolor[rgb]{0,0,1}{0.5896}(\textcolor[rgb]{1,0,1}{0.5341} $\Downarrow$)            &\textcolor[rgb]{0,0,1}{0.9837}(\textcolor[rgb]{1,0,1}{0.9879} \textcolor[rgb]{1,0,0}{$\Uparrow$})
&\textcolor[rgb]{0,0,1}{0.0163}(\textcolor[rgb]{1,0,1}{0.0121} \textcolor[rgb]{1,0,0}{$\Uparrow$})            &\textcolor[rgb]{0,0,1}{0.4104}(\textcolor[rgb]{1,0,1}{0.4659} $\Downarrow$)         &\textcolor[rgb]{0,0,1}{2.5053}(\textcolor[rgb]{1,0,1}{2.1386} \textcolor[rgb]{1,0,0}{$\Uparrow$})
&\textcolor[rgb]{0,0,1}{0.4619}(\textcolor[rgb]{1,0,1}{0.5244} \textcolor[rgb]{1,0,0}{$\Uparrow$})            &\textcolor[rgb]{0,0,1}{0.4895}(\textcolor[rgb]{1,0,1}{0.4933} \textcolor[rgb]{1,0,0}{$\Uparrow1\%$})\\

PTZ
&\textcolor[rgb]{0,0,1}{0.8293}(\textcolor[rgb]{1,0,1}{0.8713} \textcolor[rgb]{1,0,0}{$\Uparrow$})            &\textcolor[rgb]{0,0,1}{0.9649}(\textcolor[rgb]{1,0,1}{0.9865} \textcolor[rgb]{1,0,0}{$\Uparrow$})
&\textcolor[rgb]{0,0,1}{0.0351}(\textcolor[rgb]{1,0,1}{0.0135} \textcolor[rgb]{1,0,0}{$\Uparrow$})            &\textcolor[rgb]{0,0,1}{0.1707}(\textcolor[rgb]{1,0,1}{0.1287} \textcolor[rgb]{1,0,0}{$\Uparrow$})         &\textcolor[rgb]{0,0,1}{3.6426}(\textcolor[rgb]{1,0,1}{1.4582} \textcolor[rgb]{1,0,0}{$\Uparrow$})
&\textcolor[rgb]{0,0,1}{0.3163}(\textcolor[rgb]{1,0,1}{0.4157} \textcolor[rgb]{1,0,0}{$\Uparrow$})            &\textcolor[rgb]{0,0,1}{0.3806}(\textcolor[rgb]{1,0,1}{0.5158} \textcolor[rgb]{1,0,0}{$\Uparrow36\%$})\\

turbulence
&\textcolor[rgb]{0,0,1}{0.8213}(\textcolor[rgb]{1,0,1}{0.8423} \textcolor[rgb]{1,0,0}{$\Uparrow$})            &\textcolor[rgb]{0,0,1}{0.9998}(\textcolor[rgb]{1,0,1}{0.9996} $\Downarrow$)
&\textcolor[rgb]{0,0,1}{0.0002}(\textcolor[rgb]{1,0,1}{0.0004} $\Downarrow$)            &\textcolor[rgb]{0,0,1}{0.1787}(\textcolor[rgb]{1,0,1}{0.1577} \textcolor[rgb]{1,0,0}{$\Uparrow$})         &\textcolor[rgb]{0,0,1}{0.1373}(\textcolor[rgb]{1,0,1}{0.1352} \textcolor[rgb]{1,0,0}{$\Uparrow$})
&\textcolor[rgb]{0,0,1}{0.9318}(\textcolor[rgb]{1,0,1}{0.8788} $\Downarrow$)            &\textcolor[rgb]{0,0,1}{0.8689}(\textcolor[rgb]{1,0,1}{0.8590} $\Downarrow1\%$)\\
\midrule
\bf{Overall}
&\textcolor[rgb]{0,0,1}{0.7968}(\textcolor[rgb]{1,0,1}{0.8027} \textcolor[rgb]{1,0,0}{$\Uparrow$})            &\textcolor[rgb]{0,0,1}{0.9916}(\textcolor[rgb]{1,0,1}{0.9940} \textcolor[rgb]{1,0,0}{$\Uparrow$})
&\textcolor[rgb]{0,0,1}{0.0084}(\textcolor[rgb]{1,0,1}{0.0060} \textcolor[rgb]{1,0,0}{$\Uparrow$})            &\textcolor[rgb]{0,0,1}{0.2032}(\textcolor[rgb]{1,0,1}{0.1973} \textcolor[rgb]{1,0,0}{$\Uparrow$})         &\textcolor[rgb]{0,0,1}{1.5895}(\textcolor[rgb]{1,0,1}{1.2774} \textcolor[rgb]{1,0,0}{$\Uparrow$})
&\textcolor[rgb]{0,0,1}{0.7658}(\textcolor[rgb]{1,0,1}{0.7923} \textcolor[rgb]{1,0,0}{$\Uparrow$})            &\textcolor[rgb]{0,0,1}{0.7420}(\textcolor[rgb]{1,0,1}{0.7649} \textcolor[rgb]{1,0,0}{$\Uparrow3\%$})\\
\bottomrule
\end{tabular}}
\begin{tablenotes}
\item[1] Note that blue entries indicate the original SuBSENSE  results.  In parentheses, the purple entries indicate the RTSS$_{\text{SuBSENSE+ICNet}}$ results, and the arrows show the variation when compared to the original SuBSENSE results.
\end{tablenotes}
\label{resultsOfCD2014ICNet}
\end{threeparttable}
\end{table*}

%subsense+pspnet
\begin{table*}[!ht]
\centering
\begin{threeparttable}
\caption{Complete results of the RTSS framework with \textcolor[rgb]{0,0,1}{SuBSENSE} and \textcolor[rgb]{1,0,1}{PSPNet}   on the CDnet 2014 dataset\tnote{1}}
{\tabcolsep2pt\begin{tabular}{cccccccc}
\toprule
Category        &Recall         &Specificity        &FPR            &FNR            &PWC            &Precision           &F-Measure\\
\midrule
baseline
&\textcolor[rgb]{0,0,1}{0.9519}(\textcolor[rgb]{1,0,1}{0.9659} $\Uparrow$)       &\textcolor[rgb]{0,0,1}{0.9982}(\textcolor[rgb]{1,0,1}{0.9979} $\Downarrow$)      &\textcolor[rgb]{0,0,1}{0.0018}(\textcolor[rgb]{1,0,1}{0.0021} $\Downarrow$)     &\textcolor[rgb]{0,0,1}{0.0481}(\textcolor[rgb]{1,0,1}{0.0341} \textcolor[rgb]{1,0,0}{$\Uparrow$})         &\textcolor[rgb]{0,0,1}{0.3639}(\textcolor[rgb]{1,0,1}{0.3119} \textcolor[rgb]{1,0,0}{$\Uparrow$})
&\textcolor[rgb]{0,0,1}{0.9486}(\textcolor[rgb]{1,0,1}{0.9552} \textcolor[rgb]{1,0,0}{$\Uparrow$})     &\textcolor[rgb]{0,0,1}{0.9498}(\textcolor[rgb]{1,0,1}{0.9603} \textcolor[rgb]{1,0,0}{$\Uparrow 1\%$})\\

cameraJ
&\textcolor[rgb]{0,0,1}{0.8319}(\textcolor[rgb]{1,0,1}{0.8284} $\Downarrow$)     &\textcolor[rgb]{0,0,1}{0.9901}(\textcolor[rgb]{1,0,1}{0.9936} \textcolor[rgb]{1,0,0}{$\Uparrow$})      &\textcolor[rgb]{0,0,1}{0.0099}(\textcolor[rgb]{1,0,1}{0.0064} \textcolor[rgb]{1,0,0}{$\Uparrow$})       &\textcolor[rgb]{0,0,1}{0.1681}(\textcolor[rgb]{1,0,1}{0.1716} $\Downarrow$)         &\textcolor[rgb]{0,0,1}{1.6937}(\textcolor[rgb]{1,0,1}{1.3762} \textcolor[rgb]{1,0,0}{$\Uparrow$})
&\textcolor[rgb]{0,0,1}{0.7944}(\textcolor[rgb]{1,0,1}{0.8780} \textcolor[rgb]{1,0,0}{$\Uparrow$})       &\textcolor[rgb]{0,0,1}{0.8096}(\textcolor[rgb]{1,0,1}{0.8395} \textcolor[rgb]{1,0,0}{$\Uparrow 4\%$})\\

dynamic
&\textcolor[rgb]{0,0,1}{0.7739}(\textcolor[rgb]{1,0,1}{0.9273} \textcolor[rgb]{1,0,0}{$\Uparrow$})       &\textcolor[rgb]{0,0,1}{0.9994}(\textcolor[rgb]{1,0,1}{0.9994} \textcolor[rgb]{1,0,0}{$\Uparrow$})      &\textcolor[rgb]{0,0,1}{0.0006}(\textcolor[rgb]{1,0,1}{0.0006} \textcolor[rgb]{1,0,0}{$\Uparrow$})     &\textcolor[rgb]{0,0,1}{0.2261}(\textcolor[rgb]{1,0,1}{0.0727} \textcolor[rgb]{1,0,0}{$\Uparrow$})         &\textcolor[rgb]{0,0,1}{0.4094}(\textcolor[rgb]{1,0,1}{0.1293} \textcolor[rgb]{1,0,0}{$\Uparrow$})
&\textcolor[rgb]{0,0,1}{0.8913}(\textcolor[rgb]{1,0,1}{0.9390} \textcolor[rgb]{1,0,0}{$\Uparrow$})       &\textcolor[rgb]{0,0,1}{0.8159}(\textcolor[rgb]{1,0,1}{0.9328} \textcolor[rgb]{1,0,0}{$\Uparrow14\%$})\\

intermittent
&\textcolor[rgb]{0,0,1}{0.5715}(\textcolor[rgb]{1,0,1}{0.7473} \textcolor[rgb]{1,0,0}{$\Uparrow$})            &\textcolor[rgb]{0,0,1}{0.9953}(\textcolor[rgb]{1,0,1}{0.9975} \textcolor[rgb]{1,0,0}{$\Uparrow$})
&\textcolor[rgb]{0,0,1}{0.0047}(\textcolor[rgb]{1,0,1}{0.0025} \textcolor[rgb]{1,0,0}{$\Uparrow$})            &\textcolor[rgb]{0,0,1}{0.4285}(\textcolor[rgb]{1,0,1}{0.2527} \textcolor[rgb]{1,0,0}{$\Uparrow$})         &\textcolor[rgb]{0,0,1}{4.0811}(\textcolor[rgb]{1,0,1}{2.8421} \textcolor[rgb]{1,0,0}{$\Uparrow$})
&\textcolor[rgb]{0,0,1}{0.8174}(\textcolor[rgb]{1,0,1}{0.9026} \textcolor[rgb]{1,0,0}{$\Uparrow$})            &\textcolor[rgb]{0,0,1}{0.6068}(\textcolor[rgb]{1,0,1}{0.7938} \textcolor[rgb]{1,0,0}{$\Uparrow31\%$})\\

shadow
&\textcolor[rgb]{0,0,1}{0.9441}(\textcolor[rgb]{1,0,1}{0.9681} \textcolor[rgb]{1,0,0}{$\Uparrow$})            &\textcolor[rgb]{0,0,1}{0.9920}(\textcolor[rgb]{1,0,1}{0.9972} \textcolor[rgb]{1,0,0}{$\Uparrow$})
&\textcolor[rgb]{0,0,1}{0.0080}(\textcolor[rgb]{1,0,1}{0.0028} \textcolor[rgb]{1,0,0}{$\Uparrow$})            &\textcolor[rgb]{0,0,1}{0.0559}(\textcolor[rgb]{1,0,1}{0.0319} \textcolor[rgb]{1,0,0}{$\Uparrow$})         &\textcolor[rgb]{0,0,1}{1.0018}(\textcolor[rgb]{1,0,1}{0.3999} \textcolor[rgb]{1,0,0}{$\Uparrow$})
&\textcolor[rgb]{0,0,1}{0.8645}(\textcolor[rgb]{1,0,1}{0.9441} \textcolor[rgb]{1,0,0}{$\Uparrow$})            &\textcolor[rgb]{0,0,1}{0.8995}(\textcolor[rgb]{1,0,1}{0.9557} \textcolor[rgb]{1,0,0}{$\Uparrow6\%$})\\

thermal
&\textcolor[rgb]{0,0,1}{0.8166}(\textcolor[rgb]{1,0,1}{0.8489} \textcolor[rgb]{1,0,0}{$\Uparrow$})            &\textcolor[rgb]{0,0,1}{0.9910}(\textcolor[rgb]{1,0,1}{0.9907} $\Downarrow$)
&\textcolor[rgb]{0,0,1}{0.0090}(\textcolor[rgb]{1,0,1}{0.0093} $\Downarrow$)            &\textcolor[rgb]{0,0,1}{0.1834}(\textcolor[rgb]{1,0,1}{0.1511} \textcolor[rgb]{1,0,0}{$\Uparrow$})         &\textcolor[rgb]{0,0,1}{1.9632}(\textcolor[rgb]{1,0,1}{1.4590} \textcolor[rgb]{1,0,0}{$\Uparrow$})
&\textcolor[rgb]{0,0,1}{0.8319}(\textcolor[rgb]{1,0,1}{0.8559} \textcolor[rgb]{1,0,0}{$\Uparrow$})              &\textcolor[rgb]{0,0,1}{0.8172}(\textcolor[rgb]{1,0,1}{0.8499} \textcolor[rgb]{1,0,0}{$\Uparrow4\%$})\\
\midrule
badWeather
&\textcolor[rgb]{0,0,1}{0.8082}(\textcolor[rgb]{1,0,1}{0.8028} $\Downarrow$)            &\textcolor[rgb]{0,0,1}{0.9991}(\textcolor[rgb]{1,0,1}{0.9994} \textcolor[rgb]{1,0,0}{$\Uparrow$})
&\textcolor[rgb]{0,0,1}{0.0009}(\textcolor[rgb]{1,0,1}{0.0006} \textcolor[rgb]{1,0,0}{$\Uparrow$})              &\textcolor[rgb]{0,0,1}{0.1918}(\textcolor[rgb]{1,0,1}{0.1972} $\Downarrow$)         &\textcolor[rgb]{0,0,1}{0.4807}(\textcolor[rgb]{1,0,1}{0.4314} \textcolor[rgb]{1,0,0}{$\Uparrow$})
&\textcolor[rgb]{0,0,1}{0.9183}(\textcolor[rgb]{1,0,1}{0.9475} \textcolor[rgb]{1,0,0}{$\Uparrow$})              &\textcolor[rgb]{0,0,1}{0.8577}(\textcolor[rgb]{1,0,1}{0.8657} \textcolor[rgb]{1,0,0}{$\Uparrow1\%$})\\

lowFramerate
&\textcolor[rgb]{0,0,1}{0.8267}(\textcolor[rgb]{1,0,1}{0.8224} $\Downarrow$)            &\textcolor[rgb]{0,0,1}{0.9937}(\textcolor[rgb]{1,0,1}{0.9956} \textcolor[rgb]{1,0,0}{$\Uparrow$})
&\textcolor[rgb]{0,0,1}{0.0063}(\textcolor[rgb]{1,0,1}{0.0044} \textcolor[rgb]{1,0,0}{$\Uparrow$})              &\textcolor[rgb]{0,0,1}{0.1733}(\textcolor[rgb]{1,0,1}{0.1776} $\Downarrow$)         &\textcolor[rgb]{0,0,1}{1.2054}(\textcolor[rgb]{1,0,1}{1.0276} \textcolor[rgb]{1,0,0}{$\Uparrow$})
&\textcolor[rgb]{0,0,1}{0.6475}(\textcolor[rgb]{1,0,1}{0.6861} \textcolor[rgb]{1,0,0}{$\Uparrow$})              &\textcolor[rgb]{0,0,1}{0.6659}(\textcolor[rgb]{1,0,1}{0.6970} \textcolor[rgb]{1,0,0}{$\Uparrow5\%$})\\

nightVideos
&\textcolor[rgb]{0,0,1}{0.5896}(\textcolor[rgb]{1,0,1}{0.4710} $\Downarrow$)            &\textcolor[rgb]{0,0,1}{0.9837}(\textcolor[rgb]{1,0,1}{0.9916} \textcolor[rgb]{1,0,0}{$\Uparrow$})
&\textcolor[rgb]{0,0,1}{0.0163}(\textcolor[rgb]{1,0,1}{0.0084} \textcolor[rgb]{1,0,0}{$\Uparrow$})            &\textcolor[rgb]{0,0,1}{0.4104}(\textcolor[rgb]{1,0,1}{0.5290} $\Downarrow$)         &\textcolor[rgb]{0,0,1}{2.5053}(\textcolor[rgb]{1,0,1}{1.9308} \textcolor[rgb]{1,0,0}{$\Uparrow$})
&\textcolor[rgb]{0,0,1}{0.4619}(\textcolor[rgb]{1,0,1}{0.6068} \textcolor[rgb]{1,0,0}{$\Uparrow$})            &\textcolor[rgb]{0,0,1}{0.4895}(\textcolor[rgb]{1,0,1}{0.4868} $\Downarrow1\%$)\\

PTZ
&\textcolor[rgb]{0,0,1}{0.8293}(\textcolor[rgb]{1,0,1}{0.8735} \textcolor[rgb]{1,0,0}{$\Uparrow$})            &\textcolor[rgb]{0,0,1}{0.9649}(\textcolor[rgb]{1,0,1}{0.9917} \textcolor[rgb]{1,0,0}{$\Uparrow$})
&\textcolor[rgb]{0,0,1}{0.0351}(\textcolor[rgb]{1,0,1}{0.0083} \textcolor[rgb]{1,0,0}{$\Uparrow$})            &\textcolor[rgb]{0,0,1}{0.1707}(\textcolor[rgb]{1,0,1}{0.1265} \textcolor[rgb]{1,0,0}{$\Uparrow$})         &\textcolor[rgb]{0,0,1}{3.6426}(\textcolor[rgb]{1,0,1}{0.9339} \textcolor[rgb]{1,0,0}{$\Uparrow$})
&\textcolor[rgb]{0,0,1}{0.3163}(\textcolor[rgb]{1,0,1}{0.5349} \textcolor[rgb]{1,0,0}{$\Uparrow$})            &\textcolor[rgb]{0,0,1}{0.3806}(\textcolor[rgb]{1,0,1}{0.6323} \textcolor[rgb]{1,0,0}{$\Uparrow66\%$})\\

turbulence
&\textcolor[rgb]{0,0,1}{0.8213}(\textcolor[rgb]{1,0,1}{0.8528} \textcolor[rgb]{1,0,0}{$\Uparrow$})            &\textcolor[rgb]{0,0,1}{0.9998}(\textcolor[rgb]{1,0,1}{0.9994} $\Downarrow$)
&\textcolor[rgb]{0,0,1}{0.0002}(\textcolor[rgb]{1,0,1}{0.0006} $\Downarrow$)            &\textcolor[rgb]{0,0,1}{0.1787}(\textcolor[rgb]{1,0,1}{0.1472} \textcolor[rgb]{1,0,0}{$\Uparrow$})         &\textcolor[rgb]{0,0,1}{0.1373}(\textcolor[rgb]{1,0,1}{0.1579} $\Downarrow$)
&\textcolor[rgb]{0,0,1}{0.9318}(\textcolor[rgb]{1,0,1}{0.8102} $\Downarrow$)            &\textcolor[rgb]{0,0,1}{0.8689}(\textcolor[rgb]{1,0,1}{0.8191} $\Downarrow5\%$)\\
\midrule
\bf{Overall}
&\textcolor[rgb]{0,0,1}{0.7968}(\textcolor[rgb]{1,0,1}{0.8280} \textcolor[rgb]{1,0,0}{$\Uparrow$})            &\textcolor[rgb]{0,0,1}{0.9916}(\textcolor[rgb]{1,0,1}{0.9958} \textcolor[rgb]{1,0,0}{$\Uparrow$})
&\textcolor[rgb]{0,0,1}{0.0084}(\textcolor[rgb]{1,0,1}{0.0042} \textcolor[rgb]{1,0,0}{$\Uparrow$})            &\textcolor[rgb]{0,0,1}{0.2032}(\textcolor[rgb]{1,0,1}{0.1720} \textcolor[rgb]{1,0,0}{$\Uparrow$})         &\textcolor[rgb]{0,0,1}{1.5895}(\textcolor[rgb]{1,0,1}{1.0000} \textcolor[rgb]{1,0,0}{$\Uparrow$})
&\textcolor[rgb]{0,0,1}{0.7658}(\textcolor[rgb]{1,0,1}{0.8237} \textcolor[rgb]{1,0,0}{$\Uparrow$})            &\textcolor[rgb]{0,0,1}{0.7420}(\textcolor[rgb]{1,0,1}{0.8030} \textcolor[rgb]{1,0,0}{$\Uparrow8\%$})\\
\bottomrule
\end{tabular}}
\begin{tablenotes}
\item[1] Note that blue entries indicate the original SuBSENSE results.  In parentheses, the purple entries indicate the RTSS$_{\text{SuBSENSE+PSPNet}}$ results, and the arrows show the variation when compared to the original SuBSENSE results.
\end{tablenotes}
\label{resultsOfCD2014PSPNet}
\end{threeparttable}
\end{table*}

%各种语义分割算法结果

\begin{table*}[ht]
\centering
\begin{threeparttable}
\caption{Overall and per-category F-Measure  scores when SuBSENSE cooperate with different semantic segmentation algorithms}%\tnote{1}
{\tabcolsep3pt\begin{tabular}{cccccccccccccc}
\toprule
Method  & Overall  & $F_{baseline}$  & $F_{cam. jitt.}$   & $F_{dyn. bg.}$  & $F_{int. mot.}$ & $F_{shadow}$  & $F_{thermal}$   & $F_{bad. wea.}$
 & $F_{low. fr.}$   & $F_{night}$         & $F_{PTZ}$     & $F_{trubul.}$              \\
\midrule
ERFNet
&   {0.7570}&   {0.9549}&   {0.8104}&   {0.8908}&   {0.7708}&   {0.9297}&   {0.6298}&   {0.8474}&   {0.6728}&   {0.5276}&   {0.4965}&   {0.7989}\\
LinkNet
&   {0.7593}&   {0.9541}&   {0.8118}&   {0.8532}&   {0.7203}&   {0.9145}&   {0.7548}&   {0.8627}&   {0.6869}&   {0.5052}&   {0.4610}&   {0.8278}\\
PSPNet$_\text{N=3}$
&   {0.7755}&   {0.9526}&   {0.8178}&   {0.9263}&   {0.7892}&   {0.9394}&   {0.8496}&   {0.8544}&   {0.5958}&   {0.4615}&   {0.5477}&   {0.7962}\\
PSPNet$_\text{N=2}$
&   {0.7892}&   {0.9574}&   {0.8345}&   {0.9322}&   {0.7929}&   {0.9500}&   {0.8488}&   {0.8628}&   {0.6249}&   {0.4741}&   {0.5917}&   {0.8120}\\
%PSPNet$_\text{N=1}$
%&   {0.8030}&   {0.9603}&   {0.8395}&   {0.9328}&   {0.7938}&   {0.9557}&   {0.8499}&   {0.8657}&   {0.6970}&   {0.4868}&   {0.6323}&   {0.8191}\\
MFCN$_\text{supervised}$
&   {0.9570}&   {0.9798}&   {0.9740}&   {0.9781}&   {0.9385}&   {0.9859}&   {0.9699}&   {0.9693}&   {0.9048}&   {0.9173}&   {0.9516}&   {0.9582}\\
\bottomrule
\end{tabular}}
%\begin{tablenotes}
%\item[1] The methods in {\textcolor{orange}{orange}} color are the {supervised}  methods, while the others are unsupervised.
%\end{tablenotes}
\label{ablationstudysemantic}
\end{threeparttable}
\end{table*}

\subsection{Ablation Analysis}

\begin{table*}[ht]
\begin{center}
\caption{Ablation study for BGS algorithms on the CDnet2014 dataset}
{\tabcolsep13pt\begin{tabular}{cccccccc}
\toprule
 Method                                         & Recall   & Specificity  & FPR      & FNR      & PWC      & Precision   & F-Measure\\
\midrule
{GMM~\cite{stauffer1999adaptive}}                                & \textcolor[rgb]{0,0,1}{0.6195}   & \textcolor[rgb]{0,0,1}{0.9870}
                                                & \textcolor[rgb]{0,0,1}{0.0130}   & \textcolor[rgb]{0,0,1}{0.3805}   & \textcolor[rgb]{0,0,1}{2.9030}
                                                & \textcolor[rgb]{0,0,1}{0.6171}   & \textcolor[rgb]{0,0,1}{0.5390}\\ %ok
{RTSS$_{\text{GMM+ICNet}}$}                    & \textcolor[rgb]{1,0,1}{0.6792}   & \textcolor[rgb]{1,0,1}{0.9920}
                                                & \textcolor[rgb]{1,0,1}{0.0080}   & \textcolor[rgb]{1,0,1}{0.3208}   & \textcolor[rgb]{1,0,1}{2.0622}
                                                & \textcolor[rgb]{1,0,1}{0.6992}   & \textcolor[rgb]{1,0,1}{0.6302} (\textcolor[rgb]{1,0,0}{$\Uparrow17\%$})\\ %ok
{RTSS$_{\text{GMM+PSPNet}}$}                    & \textcolor[rgb]{1,0,1}{0.7146}   & \textcolor[rgb]{1,0,1}{0.9948}
                                                & \textcolor[rgb]{1,0,1}{0.0052}   & \textcolor[rgb]{1,0,1}{0.2854}   & \textcolor[rgb]{1,0,1}{1.6941}
                                                & \textcolor[rgb]{1,0,1}{0.7517}   & \textcolor[rgb]{1,0,1}{0.6957} (\textcolor[rgb]{1,0,0}{$\Uparrow29\%$})\\ %ok
\midrule
{ViBe ~\cite{barnich2011vibe}}                              & \textcolor[rgb]{0,0,1}{0.6146}   & \textcolor[rgb]{0,0,1}{0.9738}
                                                & \textcolor[rgb]{0,0,1}{0.0262}   & \textcolor[rgb]{0,0,1}{0.3854}   & \textcolor[rgb]{0,0,1}{3.8341}
                                                & \textcolor[rgb]{0,0,1}{0.6614}   & \textcolor[rgb]{0,0,1}{0.5761}\\ %ok
{RTSS$_{\text{ViBe+ICNet}}$}                   & \textcolor[rgb]{1,0,1}{0.6465}   & \textcolor[rgb]{1,0,1}{0.9882}
                                                & \textcolor[rgb]{1,0,1}{0.0118}   & \textcolor[rgb]{1,0,1}{0.3535}   & \textcolor[rgb]{1,0,1}{2.2954}
                                                & \textcolor[rgb]{1,0,1}{0.7210}   & \textcolor[rgb]{1,0,1}{0.6307} (\textcolor[rgb]{1,0,0}{$\Uparrow9\%$})\\ %ok
{RTSS$_{\text{ViBe+PSPNet}}$}                   & \textcolor[rgb]{1,0,1}{0.6872}   & \textcolor[rgb]{1,0,1}{0.9940}
                                                & \textcolor[rgb]{1,0,1}{0.0060}   & \textcolor[rgb]{1,0,1}{0.3128}   & \textcolor[rgb]{1,0,1}{1.5994}
                                                & \textcolor[rgb]{1,0,1}{0.7789}   & \textcolor[rgb]{1,0,1}{0.6978} (\textcolor[rgb]{1,0,0}{$\Uparrow21\%$})\\ %ok
\midrule

{PBAS~\cite{hofmann2012background}}                              & \textcolor[rgb]{0,0,1}{0.7268}   & \textcolor[rgb]{0,0,1}{0.9542}
                                                & \textcolor[rgb]{0,0,1}{0.0458}   & \textcolor[rgb]{0,0,1}{0.2732}      & \textcolor[rgb]{0,0,1}{5.2080}
                                                & \textcolor[rgb]{0,0,1}{0.7019}   & \textcolor[rgb]{0,0,1}{0.6426}\\ %ok
{RTSS$_{\text{PBAS+ICNet}}$}                   & \textcolor[rgb]{1,0,1}{0.7265}   & \textcolor[rgb]{1,0,1}{0.9824}
                                                & \textcolor[rgb]{1,0,1}{0.0176}   & \textcolor[rgb]{1,0,1}{0.2735}    & \textcolor[rgb]{1,0,1}{2.4598}
                                                & \textcolor[rgb]{1,0,1}{0.7327}   & \textcolor[rgb]{1,0,1}{0.6733} (\textcolor[rgb]{1,0,0}{$\Uparrow5\%$})\\ %ok
{RTSS$_{\text{PBAS+PSPNet}}$}                   & \textcolor[rgb]{1,0,1}{0.7630}   & \textcolor[rgb]{1,0,1}{0.9914}
                                                & \textcolor[rgb]{1,0,1}{0.0086}   & \textcolor[rgb]{1,0,1}{0.2370}    & \textcolor[rgb]{1,0,1}{1.4363}
                                                & \textcolor[rgb]{1,0,1}{0.7925}   & \textcolor[rgb]{1,0,1}{0.7411} (\textcolor[rgb]{1,0,0}{$\Uparrow15\%$})\\ %ok
\bottomrule
\end{tabular}}
\label{AblationStudyForBGSAlgorithms}
\end{center}
\end{table*}

\textbf{Ablation study for semantic segmentation algorithms}:   %   \textit{$\text{RTSS}_{\text{SuBSENSE+ICNet}}$}
In Table \ref{resultsOfCD2014ICNet}, we present the complete quantitative evaluation results of the proposed method on the CDnet 2014 dataset, seven metrics and overall scores are reported. The blue entries indicate the original SuBSENSE results, while the proposed RTSS framework which combines SuBSENSE and ICNet  (RTSS$_{\text{SuBSENSE+ICNet}}$) results are shown in the parentheses with purple entries. The arrows indicate the variations between the two. We can see that RTSS$_{\text{SuBSENSE+ICNet}}$ achieves an over F-Measure score of 0.7649, which is a 3\% improvement  compared to the original SuBSENSE. This improvement is mainly benefited from ``dynamic background'', ``intermittent object motion'', ``shadow'', ``low framerate'', and ``pan-tilt-zoom'' categories.
The dynamic background category contains six challenging videos depicting outdoor scenarios, where there is dynamic motion in the background such as trees shaking in the wind, boats moving on the river and cars passing behind  rippling fountains.
However, RTSS$_{\text{SuBSENSE+ICNet}}$ improves the F-Measure by a significant margin of 11\% on this category, which demonstrates its tolerance  for dealing with background changes.
The intermittent object motion category includes six videos, each of which contains objects with intermittent motion such as parked cars suddenly start moving and objects are abandoned, which resulting ghost artifacts in the detected result. Due to the  object-level semantic information provided by the semantic segmenter  $\mathcal{S}$, the proposed method has a remarkable 29\% improvement on the F-Measure score.
The same can be said for the pan-tilt-zoom (PTZ) category, which contains videos captured with moving camera, making it hard for most algorithms to adapt to the scene changes.
Due to the use of  potential foreground semantic segmentation, the proposed method can  distinguish the  potential foreground and background information quickly  despite camera motions,  which makes an amazing 36\% improvement for this category.
For the shadow category which comprised videos exhibiting shadows and reflections       and the low framerate category which included four outdoor videos with the frame rate varies from 0.17 to 1 FPS,  RTSS$_{\text{SuBSENSE+ICNet}}$ also has a 3\% improvement of the F-Measure score.

However, from Table~\ref{resultsOfCD2014ICNet}, we can also find that in some categories such as baseline, camera jitter, thermal, bad weather and turbulence, compared to the original SuBSENSE,  RTSS$_{\text{SuBSENSE+ICNet}}$ performs worse than the original algorithm.
%We can't help asking why this is the case? Our first thought is that  the precision of the semantic segmenter  $\mathcal{S}$  is  not accurate enough. Therefore, we try to introduce the state-of-the-art semantic segmentation method PSPNet~\cite{zhao2017pyramid} into the RTSS framework to replace the ICNet. Here, we call the new algorithm RTSS$_{\text{SuBSENSE+PSPNet}}$, %(We should notice that \textit{$\text{RTSS}_{\text{SuBSENSE+PSPNet}}$} can not run in real-time).
Our hypothesis is that the precision of the semantic segmenter $\mathcal{S}$ is not accurate enough. Because  similar phenomena can be found  when other real-time semantic segmentation algorithms are used, such as ERFNet~\cite{romera2018erfnet}, LinkNet~\cite{chaurasia2017linknet}, as shown in Table~\ref{ablationstudysemantic} (more complete and detailed results are reported in Appendix).
Therefore, we introduce the state-of-the-art semantic segmentation method PSPNet  into the RTSS framework to replace the ICNet. We call the new algorithm RTSS$_{\text{SuBSENSE+PSPNet}}$, %(We should notice that \textit{$\text{RTSS}_{\text{SuBSENSE+PSPNet}}$} can not run in real-time).
repeat the same experimental process as before, and the final results are shown in Table~\ref{resultsOfCD2014PSPNet}, which illustrates that our hypothesis seems correct.

We can see that RTSS$_{\text{SuBSENSE+PSPNet}}$ achieves an overall F-Measure of 0.8030 on the whole dataset, about 8\% improvement compared to the 3\% of RTSS$_{\text{SuBSENSE+ICNet}}$, which is a significant margin. In addition to the night videos and turbulence categories, all the remaining categories have a significant performance improvement especially in the  dynamic background, intermittent object motion, shadow and PTZ categories. The trend of these performance improvements is consistent with the RTSS$_{\text{SuBSENSE+ICNet}}$ results.
In the turbulence category, the videos are captured with long distance thermal lens. The most prominent characteristic of this category is that the sequences suffer from air turbulence under high temperature environments, which resulting distortion in each frame.
Since the semantic segmentation network model is trained on the  ADE20K dataset~\cite{zhou2017scene}, images from the turbulence category have a big difference  with the ADE20K dataset, thus the semantic foreground/background segmentation  for these sequences are unstable  and inaccurate.
So we can see that both RTSS$_{\text{SuBSENSE+ICNet}}$ and RTSS$_{\text{SuBSENSE+PSPNet}}$ have  the F-Measure decrease in this category.
If the semantic segmenter  $\mathcal{S}$ can provide more accuracy semantic information,  we may get better results.
However, we should also notice that the high computational complexity of PSPNet makes it difficult to achieve real-time performance if we perform semantic segmentation for every frame in the RTSS framework.    So, as we mentioned before,   consecutive frames in a video are highly redundant, we can do the semantic segmentation every N frames.
Table~\ref{ablationstudysemantic} shows the results of PSPNet with N = 2 and N = 3 (more complete and detailed results are reported in Appendix).  Although their results are not as good as N = 1, they also show a big improvement compared with the original SuBSENSE.

Finally, we use a  deep learning-based supervised BGS algorithm  MFCN~\cite{zeng2018multiscale} as the semantic segmenter (since we cannot get a more precise  semantic segmentation algorithm than PSPNet at present). For each sequence in the CDnet 2014 dataset, a subset of 200 input frames with their corresponding ground truths is randomly and manually selected to train the  model. Then we take the output of MFCN on each frame as the semantic information.
Repeat the same experimental process as before, we get an overall  F-Measure of 0.9570, this again verifies our  speculations.

\vspace{0.03\linewidth}

\textbf{Ablation study for BGS algorithms}:
We now make an ablation study for BGS algorithms on the proposed RTSS framework. Three classic BGS algorithms GMM~\cite{stauffer1999adaptive}, ViBe~\cite{barnich2011vibe} and PBAS~\cite{hofmann2012background} are introduced into the RTSS framework to replace SuBSENSE.
We report the experimental results in Table~\ref{AblationStudyForBGSAlgorithms}.  The blue entries indicate the original BGS algorithm results, the purple entries indicate the RTSS results, and the arrows show the F-Measure variation when compared to the original algorithm.
We can see that all of them have a remarkable improvement when compared with their original BGS algorithms.  The RTSS$_{\text{GMM+ICNet}}$ which combines GMM  and ICNet improves from 0.5390 to 0.6302, about 17\% F-Measure improvement, and RTSS$_{\text{ViBe+ICNet}}$ is about 9\%, RTSS$_{\text{PBAS+ICNet}}$ is about 5\%. When more accurate semantic segmentation algorithm is used, the promotion effect is more obvious,  we can find that
RTSS$_{\text{GMM+PSPNet}}$, RTSS$_{\text{ViBe+PSPNet}}$, RTSS$_{\text{PBAS+PSPNet}}$ have 29\%, 21\%, 15\% promotion, respectively. %But we should notice that the PSPNet cannot run in real-time at present.

Based on above analysis, we can find that the proposed RTSS framework is very flexible,  and can be applied to many BGS algorithms.
The choices of BGS segmenter $\mathcal{B}$   and semantic segmenter $\mathcal{S}$ may depend on application scenarios and computational resources.
We also believe that the proposed framework  has great potential for  future improvement with the emergence of  faster and more accurate BGS algorithms and semantic segmentation algorithms.

\begin{table*}[ht]
\begin{center}
\caption{ Comparison of the results on the CDnet 2014 dataset between SemanticBGS and RTSS}
{\tabcolsep12pt\begin{tabular}{cccccccc}
\toprule
 Method                                                     & Recall        & Specificity   & FPR           & FNR           & PWC           & Precision     & F-Measure\\
\midrule
SuBSENSE~\cite{st2015subsense}                                    &\textcolor[rgb]{0,0,1}{0.7968}       &\textcolor[rgb]{0,0,1}{0.9916}
                                                            &\textcolor[rgb]{0,0,1}{0.0084}       &\textcolor[rgb]{0,0,1}{0.2032}       &\textcolor[rgb]{0,0,1}{1.5895}
                                                            &\textcolor[rgb]{0,0,1}{0.7658}       &\textcolor[rgb]{0,0,1}{0.7420}\\
\midrule
{SemanticBGS$_{\text{SuBSENSE+ICNet}}$}    &{0.7797}       &{0.9939}       &{0.0061}       &{0.2203}       &{1.4529}       &{0.7862}       &{0.7490} (\textcolor[rgb]{1,0,0}{$\Uparrow1\%$})\\

{RTSS$_{\text{SuBSENSE+ICNet}}$}                           &\bf{0.8027}    & \bf{0.9940}   & \bf{0.0060}   &\bf{0.1973}    & \bf{1.2774}   & \bf{0.7923}   & \bf{0.7649 (\textcolor[rgb]{1,0,0}{$\Uparrow3\%$})}\\
\midrule
{SemanticBGS$_{\text{SuBSENSE+PSPNet}}$}    &{0.8079}       & {0.9955}      &{0.0045}       &{0.1926}       &{1.1294}       &{0.8153}       &{0.7874} (\textcolor[rgb]{1,0,0}{$\Uparrow6\%$})\\
{RTSS$_{\text{SuBSENSE+PSPNet}}$}                           &\bf{0.8280}    & \bf{0.9958}   & \bf{0.0042}   &\bf{0.1720}    & \bf{1.0000}   & \bf{0.8237}   & \bf{0.8030 (\textcolor[rgb]{1,0,0}{$\Uparrow8\%$})}\\
\bottomrule
\end{tabular}}{}
\label{SemanticBGSVSRTSS}
\end{center}
\end{table*}

\begin{table*}[ht]
\centering
\begin{threeparttable}
\caption{Overall and per-category F-Measure scores  on the CDnet 2014 dataset by different BGS methods\tnote{1}}
{\tabcolsep2pt\begin{tabular}{cccccccccccccc}
\toprule
Method  & Overall  & $F_{baseline}$  & $F_{cam. jitt.}$   & $F_{dyn. bg.}$  & $F_{int. mot.}$ & $F_{shadow}$  & $F_{thermal}$   & $F_{bad. wea.}$
 & $F_{low. fr.}$   & $F_{night}$         & $F_{PTZ}$     & $F_{trubul.}$              \\
\midrule
{\textcolor{orange}{FgSegNet~\cite{lim2018foreground}}}
&   {0.9770}&   {0.9973} &  {0.9954}&   {0.9958}&   {0.9951}&   {0.9937}&   {0.9921}&   {0.9845}&   {0.8786}&   {0.9655}&   {0.9843}&   {0.9648}\\
{\textcolor{orange}{CascadeCNN~\cite{wang2017interactive}}}
&   {0.9209}&   {0.9786}&   {0.9758}&   {0.9658}&   {0.8505}&   {0.9593}&   {0.8958}&   {0.9431}&   {0.8370}&   {0.8965}&   {0.9168}&   {0.9108}\\
\textcolor{blue}{{RTSS$_{\text{SuBSENSE+PSPNet}}$}   }
&   {0.7917}&   {0.9597}&   {0.8396}&   {0.9325}&   {0.7864}&   {0.9551}&   {0.8510}&   {0.8662}&   {0.6771}&   {0.5295}&   {0.5489}&   {0.7630}\\
%\textcolor{blue}{{RTSS$_{\text{SuBSENSE+PSPNet(N=2)}}$}   }
%&   {0.7917}&   {0.9597}&   {0.8396}&   {0.9325}&   {0.7864}&   {0.9551}&   {0.8510}&   {0.8662}&   {0.6771}&   {0.5295}&   {0.5489}&   {0.7630}\\
%\textcolor{blue}{{RTSS$_{\text{SuBSENSE+PSPNet(N=3)}}$}   }
%&   {0.7917}&   {0.9597}&   {0.8396}&   {0.9325}&   {0.7864}&   {0.9551}&   {0.8510}&   {0.8662}&   {0.6771}&   {0.5295}&   {0.5489}&   {0.7630}\\
{\textcolor{orange}{IUTIS-5~\cite{bianco2017far}}}
&   {0.7717}&   {0.9567}&   {0.8332}&   {0.8902}&   {0.7296}&   {0.8766}&   {0.8303}&   {0.8248}&   {0.7743}&   {0.5290}&   {0.4282}&   {0.7836}\\
\textcolor{blue}{{RTSS$_{\text{SuBSENSE+ICNet}}$}  }
&   {0.7571}&   {0.9312}&   {0.7995}&   {0.9053}&   {0.7802}&   {0.9242}&   {0.7007}&   {0.8454}&   {0.6643}&   {0.5479}&   {0.4324}&   {0.7969}\\
\textcolor{blue}{{RTSS$_{\text{SuBSENSE+LinkNet}}$}  }
&   {0.7502}&   {0.9541}&   {0.8118}&   {0.8532}&   {0.7203}&   {0.9145}&   {0.7548}&   {0.8612}&   {0.6687}&   {0.5480}&   {0.3911}&   {0.7744}\\
\textcolor{blue}{SharedModel~\cite{chen2015learning}}
&   {0.7474}&   {0.9522}&   {0.8141}&   {0.8222}&   {0.6727}&   {0.8898}&   {0.8319}&   {0.8480}&   {0.7286}&   {0.5419}&   {0.3860}&   {0.7339}\\
{\textcolor{orange}{DeepBS~\cite{babaee2018deep}}}
&   {0.7458}&   {0.9580}&   {0.8990}&   {0.8761}&   {0.6098}&   {0.9304}&   {0.7583}&   {0.8301}&   {0.6002}&   {0.5835}&   {0.3133}&   {0.8455}\\
\textcolor{blue}{{RTSS$_{\text{SuBSENSE+ERFNet}}$}  }
&   {0.7450}&   {0.9549}&   {0.8104}&   {0.8908}&   {0.7708}&   {0.9276}&   {0.6298}&   {0.8457}&   {0.6512}&   {0.5455}&   {0.4151}&   {0.7528}\\
\textcolor{blue}{WeSamBE ~\cite{jiang2017wesambe}}
&   {0.7446}&   {0.9413}&   {0.7976}&   {0.7440}&   {0.7392}&   {0.8999}&   {0.7962}&   {0.8608}&   {0.6602}&   {0.5929}&   {0.3844}&   {0.7737}\\
\textcolor{blue}{SuBSENSE ~\cite{st2015subsense}}
&   {0.7408}&   {0.9503}&   {0.8152}&   {0.8177}&   {0.6569}&   {0.8986}&   {0.8171}&   {0.8619}&   {0.6445}&   {0.5599}&   {0.3476}&   {0.7792}\\
\textcolor{blue}{PAWCS ~\cite{st2016universal}}
&   {0.7403}&   {0.9397}&   {0.8137}&   {0.8938}&   {0.7764}&   {0.8913}&   {0.8324}&   {0.8152}&   {0.6588}&   {0.4152}&   {0.4615}&   {0.6450}\\
\textcolor{blue}{C-EFIC ~\cite{allebosch2015c}}
&   {0.7307}&   {0.9309}&   {0.8248}&   {0.5627}&   {0.6229}&   {0.8778}&   {0.8349}&   {0.7867}&   {0.6806}&   {0.6677}&   {0.6207}&   {0.6275}\\
\bottomrule
\end{tabular}}
\begin{tablenotes}
\item[1] The methods in \textcolor{blue}{blue} color are the \emph{unsupervised} BGS methods, like ours. For completeness, we also show in {\textcolor{orange}{orange}} color  the \emph{supervised} BGS methods.  %the deep learning and genetic programming  based
\end{tablenotes}
\label{stateoftheartcomparsion}
\end{threeparttable}
\end{table*}

\subsection{Comparisons with  State-of-the-art Methods}

\textbf{Quantitative Evaluation}:
Firstly, to demonstrate one of our key contribution, the proposed RTSS framework is preferable to the method presented in~\cite{braham2017semantic} by  taking a semantic segmentation  as a post-processing operation to  refine the BGS segmentation result (we call this algorithm SemanticBGS),  in Table~\ref{SemanticBGSVSRTSS}, we present  the  performance comparison results. The first row shows the seven metric results of the original SuBSENSE BGS algorithm, the second row gives the comparison results of SemanticBGS and RTSS when cooperated with the ICNet~\cite{zhao2018icnet} semantic segmentor, and the third row shows the results when using the PSPNet~\cite{zhao2017pyramid}.
For a specific metric, if the method obtains a better performance, the corresponding  value is highlighted in bold.
From Table~\ref{SemanticBGSVSRTSS}, we can see that the proposed method RTSS surpasses the SemanticBGS method on all seven metrics. The overall F-Measure score of {SemanticBGS$_{\text{SuBSENSE+ICNet}}$} is only with  nearly 1\% improvement compared with the proposed RTSS of 3\%. When combined with the PSPNet,  the SemanticBGS has a 6\% promotion while RTSS obtains 8\% improvement, the gap is obvious.
Therefore, this fully demonstrates the effectiveness of the algorithm proposed in this paper, that is, feeding back the refined FG/BG mask produced by the BGS segmenter and semantic segmenter to  the background subtraction process to guide the background model updating can significantly improve the final accuracy.

We also compare our proposed method with other state-of-the-art algorithms listed on \textit{{\color{blue}{\url{www.changedetection.net}}}}. In Table~\ref{stateoftheartcomparsion}, we give a detailed overall and per-category F-Measure comparison, as well as the average ranking. The results are reported by the online evaluation server\footnote{{\color{blue}{http://jacarini.dinf.usherbrooke.ca/results2014/589/}} (589-592).   For  some sequences in the CDnet 2014 dataset, their ground truths are not public available, so the local evaluation results presented in the previous tables may have some difference with the online server reported results.}. Due to space limitation, we only show the top-ranked methods: FgSegNet~\cite{lim2018foreground}, CascadeCNN~\cite{wang2017interactive}, IUTIS-5~\cite{bianco2017far}, SharedModel ~\cite{chen2015learning}, DeepBS~\cite{babaee2018deep}, WeSamBE~\cite{jiang2017wesambe}, SuBSENSE~\cite{st2015subsense}, PAWCS~\cite{st2016universal} and C-EFIC~\cite{allebosch2015c}. Among them, FgSegNet, CascadeCNN and DeepBS are deep learning based supervised methods.
%Although IUTIS-5 is not deep learning based, however,  it is a combined algorithm based on five other state-of-the-art methods, and this method also involves a training process with ground truths, therefore, we regard it as a supervised method.
Since IUTIS involves a training process with ground truths, here, we also regard it as a supervised method.
In Table~\ref{stateoftheartcomparsion}, these supervised methods are marked with orange and the unsupervised ones with blue.
Firstly, it can be observed that the top two methods are all deep learning based, FgSegNet even achieves an  overall F-Measure of 0.9770, which is much higher than  the proposed RTSS algorithms.
However, as we stated earlier, from the point of view of applications, BGS algorithms for video surveillance should be unsupervised. Therefore, it could be argued whether these algorithms can
be applied to scenarios like the ones we consider in this paper.
Secondly, we can see that the proposed RTSS framework achieves state-of-the-art performance among all unsupervised methods and in fact even performs better than some deep learning based method such as DeepBS.
It is worth noting that RTSS is a very flexible framework, allowing for easy changing the components if required.

\begin{figure*}[!htbp]
\centering
\includegraphics[width=1.0\linewidth]{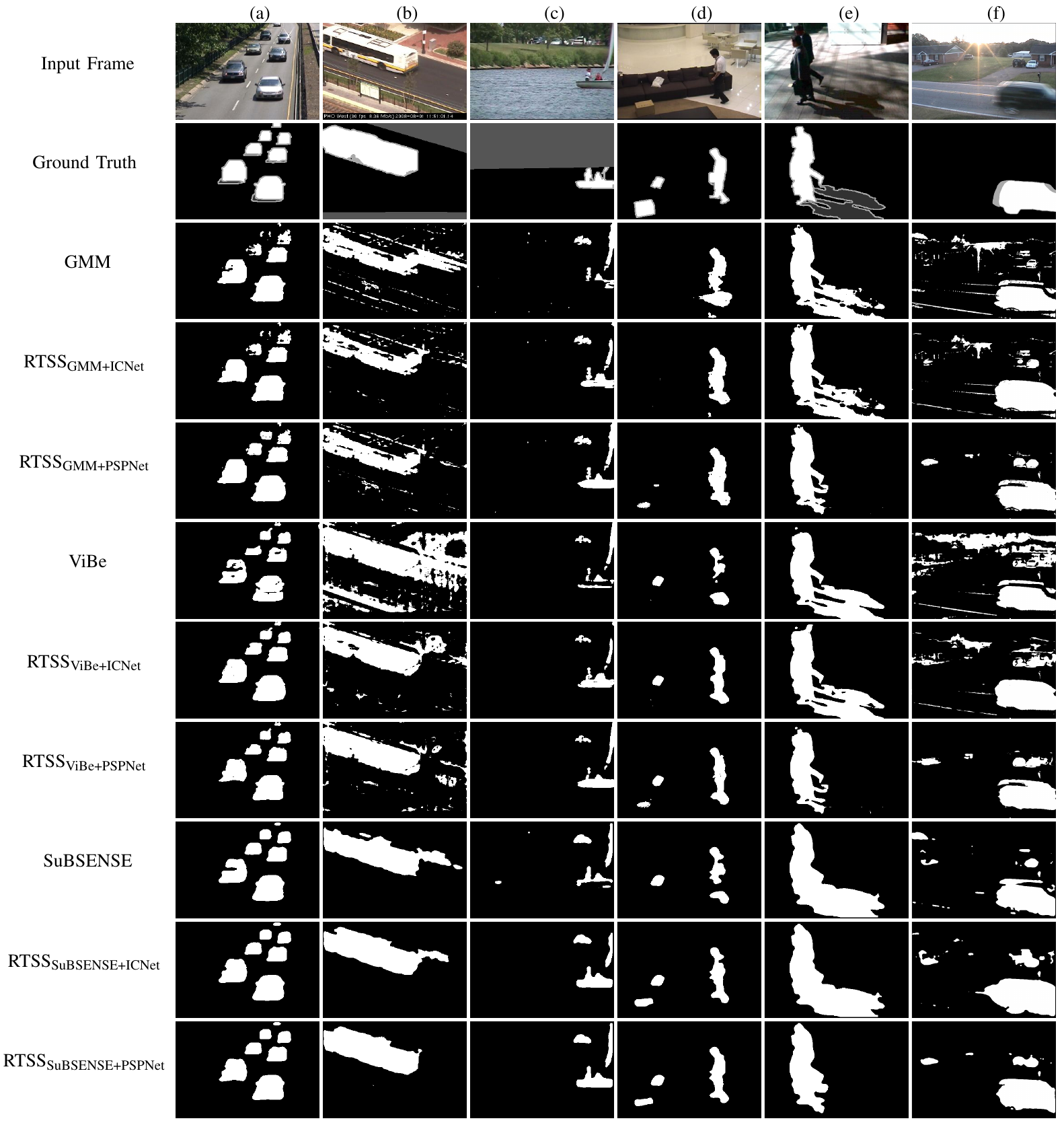}%qualitative
\vspace{-0.02\linewidth}
\caption{Comparison of the qualitative results on various sequences from CDnet 2014 dataset~\cite{wang2014cdnet}. From left to right: (a)  Sequence \textit{highway} from category ``baseline''.   (b) Sequence \textit{boulevard} from category ``camera jitter''.  (c) Sequence \textit{boats} from category ``dynamic background''. (d) Sequence \textit{sofa} from category ``intermittent object motion''. (e) Sequence \textit{peopleInShade} from category ``shadow''. (f) Sequence \textit{continuousPan} from category ``pan-tilt-zoom''.
From top to bottom: Input Frame, Ground Truth, GMM~\cite{stauffer1999adaptive}, RTSS$_{\text{GMM+ICNet}}$, RTSS$_{\text{GMM+PSPNet}}$, ViBe~\cite{barnich2011vibe}, RTSS$_{\text{ViBe+ICNet}}$,
RTSS$_{\text{ViBe+PSPNet}}$, SuBSENSE~\cite{st2015subsense}, RTSS$_{\text{SuBSENSE+ICNet}}$   and RTSS$_{\text{SuBSENSE+PSPNet}}$ foreground segmentation results.
}
\label{CDnet2014QualitativeComparison}
\end{figure*}

\textbf{Qualitative evaluation}:
Visual comparison of the foreground segmentation results are reported with 6 challenging sequences from the CDnet 2014 dataset~\cite{wang2014cdnet}: \textit{highway} ($t$ = 818)  from the ``baseline'' category, \textit{boulevard} ($t$ = 2318)  from the ``camera jitter'' category, \textit{boats} ($t$ = 7914)  from the ``dynamic background'' category,  \textit{sofa} ($t$ = 1169)  from the ``intermittent object motion'' category, \textit{peopleInShade} ($t$ = 1098)  from the ``shadow'' category and \textit{continuousPan} ($t$ = 989)  from the ``pan-tilt-zoom'' category.  Fig.~\ref{CDnet2014QualitativeComparison} provides qualitative comparison results with different methods.
Visually, we can see that the results of the proposed RTSS method  look  much  better than the corresponding baseline method, which shows good agreement with  above quantitative evaluation results.

In the \textit{highway} and \textit{boats} sequences, which contain dynamic backgrounds such as tree branches moving with the wind  and rippling water in the river, the proposed method can  extract  complete foreground objects and shows good robustness to the  background disturbance.
In the  \textit{boulevard} sequences, due to the repetitive jitter of the camera, many false positives are produced by other BGS methods. However, our foreground detection result is very accurate and complete.
In the \textit{sofa} sequence, which contains  challenge about intermittent object motion, many false negatives are produced by  other methods due to model absorption,
but we can see that the proposed method  performs well.
In the \textit{peopleInShade} sequence from the  ``shadow'' category, we can find that although the shadow is strong, it only has a small impact on our results.
In the \textit{continuousPan}, although the camera is not static, RTSS also detects cars perfectly   and shows  good robustness for the motion background.
All the above analysis again demonstrates the effectiveness of the proposed RTSS in  variety challenging situations.

\subsection{Processing Speed}

We know that background subtraction is often the first step in many computer vision applications. Processing speed is a critical factor.
In order to analyze the computational cost of the proposed method,  we report the processing  time of our method runs on the \textit{highway} ($240\times320$) sequence. The results are shown in Table~\ref{runningSpeed1}.  % and~\ref{runningSpeed2}.
The experiment is performed on a computer with an Intel Core-i7 8700K processor  with 32 GB memory  and  an NVIDIA Titan Xp GPU. The  operating system is  Ubuntu 16.04. We do not consider I/O time. According to the results, we can see that the average processing time of {RTSS$_{\text{SuBSENSE+ICNet}}$}  is about 38 ms per frame, which shows real-time performance. If we use the more accurate PSPNet, the average processing time is about 51 ms per frame with N = 3.
However, the flexibility of the proposed RTSS framework allows the inclusion of  faster and more accurate semantic segmentation algorithms, once they emerged.

\begin{table}[ht]
\centering
\caption{RTSS processing  time (ms).}
{\tabcolsep20pt
\begin{tabular}{c|c} \hline
RTSS$_{\text{SuBSENSE+ICNet}}$  &   RTSS$_{\text{SuBSENSE+PSPNet (N=3)}}$   \\ \hline
38                              &   51               \\  \hline
\end{tabular}
}
\label{runningSpeed1}
\end{table}

\section{Conclusion}\label{CONCLUSION}

We proposed a novel background subtraction framework  called \textit{background subtraction with real-time semantic segmentation} (RTSS), which consists of two components: a  BGS segmenter $\mathcal{B}$ and a  semantic segmenter $\mathcal{S}$, which work in parallel on two threads interactively for real-time and accurate foreground segmentation.
Experimental results obtained on the CDnet 2014 dataset demonstrated the effectiveness of the proposed algorithm. Compared to other background subtraction methods, our method achieves state-of-the-art performance among all unsupervised background subtraction methods while operating at real-time, and even performs better than some deep learning based supervised algorithms. Moreover, it is worth noting that the proposed framework is very flexible, allowing for both easy adaption of future segmentation improvements and application driven component changes.

A number of improvements can be considered for our method. Our future work will focus on more efficient fusion strategy between the outputs of semantic segmentation and background subtraction, as well as more accurate and faster BGS and semantic segmentation algorithms.

\section*{Acknowledgment}
%We would like to thank the anonymous reviewers for their helpful feedback.
This research is supported by the National Science Foundation of China under Grant No.61401425.
We gratefully acknowledge the support of NVIDIA Corporation with the donation of the Titan Xp GPU used for this research.

%The authors would like to thank...

\section*{Appendix}\label{appendexTables}
Table~\ref{resultsOfCD2014PSPNetN2}-\ref{resultsOfCD2014mfcn} show the complete and detailed quantitative evaluation results of RTSS with different semantic segmentation algorithms.

% Can use something like this to put references on a page
% by themselves when using endfloat and the captionsoff option.
\ifCLASSOPTIONcaptionsoff
  \newpage
\fi

\clearpage
% !TEX root = paper_main.tex
\begin{appendices}

%N=2
\begin{table*}[!ht]
\centering
\begin{threeparttable}
\caption{Complete results of the RTSS framework with \textcolor[rgb]{0,0,1}{SuBSENSE} and \textcolor[rgb]{1,0,1}{PSPNet (N=2)}   on the CDnet 2014 dataset\tnote{1}}
{\tabcolsep2pt\begin{tabular}{cccccccc}
\toprule
Category        &Recall         &Specificity        &FPR            &FNR            &PWC            &Precision           &F-Measure\\
\midrule
baseline  %baseline :	0.9677	0.9977	0.0023	0.0323	0.3186	0.9478	0.9574
&\textcolor[rgb]{0,0,1}{0.9519}(\textcolor[rgb]{1,0,1}{0.9677} \textcolor[rgb]{1,0,0}{$\Uparrow$})    &\textcolor[rgb]{0,0,1}{0.9982}(\textcolor[rgb]{1,0,1}{0.9977} $\Downarrow$)      &\textcolor[rgb]{0,0,1}{0.0018}(\textcolor[rgb]{1,0,1}{0.0023} $\Downarrow$)     &\textcolor[rgb]{0,0,1}{0.0481}(\textcolor[rgb]{1,0,1}{0.0323} \textcolor[rgb]{1,0,0}{$\Uparrow$})         &\textcolor[rgb]{0,0,1}{0.3639}(\textcolor[rgb]{1,0,1}{0.3186} \textcolor[rgb]{1,0,0}{$\Uparrow$})
&\textcolor[rgb]{0,0,1}{0.9486}(\textcolor[rgb]{1,0,1}{0.9478} $\Downarrow$)     &\textcolor[rgb]{0,0,1}{0.9498}(\textcolor[rgb]{1,0,1}{0.9574} \textcolor[rgb]{1,0,0}{$\Uparrow 1\%$})\\

cameraJ   %cameraJ :	0.8292	0.9930	0.0070	0.1708	1.4362	0.8663	0.8345
&\textcolor[rgb]{0,0,1}{0.8319}(\textcolor[rgb]{1,0,1}{0.8292} $\Downarrow$)     &\textcolor[rgb]{0,0,1}{0.9901}(\textcolor[rgb]{1,0,1}{0.9930} \textcolor[rgb]{1,0,0}{$\Uparrow$})      &\textcolor[rgb]{0,0,1}{0.0099}(\textcolor[rgb]{1,0,1}{0.0070} \textcolor[rgb]{1,0,0}{$\Uparrow$})       &\textcolor[rgb]{0,0,1}{0.1681}(\textcolor[rgb]{1,0,1}{0.1708} $\Downarrow$)         &\textcolor[rgb]{0,0,1}{1.6937}(\textcolor[rgb]{1,0,1}{1.4362} \textcolor[rgb]{1,0,0}{$\Uparrow$})
&\textcolor[rgb]{0,0,1}{0.7944}(\textcolor[rgb]{1,0,1}{0.8663} \textcolor[rgb]{1,0,0}{$\Uparrow$})       &\textcolor[rgb]{0,0,1}{0.8096}(\textcolor[rgb]{1,0,1}{0.8345} \textcolor[rgb]{1,0,0}{$\Uparrow 3\%$})\\

dynamic   %dynamic :	0.9315	0.9993	0.0007	0.0685	0.1361	0.9335	0.9322
&\textcolor[rgb]{0,0,1}{0.7739}(\textcolor[rgb]{1,0,1}{0.9315} \textcolor[rgb]{1,0,0}{$\Uparrow$})       &\textcolor[rgb]{0,0,1}{0.9994}(\textcolor[rgb]{1,0,1}{0.9993} $\Downarrow$)      &\textcolor[rgb]{0,0,1}{0.0006}(\textcolor[rgb]{1,0,1}{0.0007} $\Downarrow$)     &\textcolor[rgb]{0,0,1}{0.2261}(\textcolor[rgb]{1,0,1}{0.0685} \textcolor[rgb]{1,0,0}{$\Uparrow$})         &\textcolor[rgb]{0,0,1}{0.4094}(\textcolor[rgb]{1,0,1}{0.1361} \textcolor[rgb]{1,0,0}{$\Uparrow$})
&\textcolor[rgb]{0,0,1}{0.8913}(\textcolor[rgb]{1,0,1}{0.9335} \textcolor[rgb]{1,0,0}{$\Uparrow$})       &\textcolor[rgb]{0,0,1}{0.8159}(\textcolor[rgb]{1,0,1}{0.9322} \textcolor[rgb]{1,0,0}{$\Uparrow14\%$})\\

intermittent  %intermi :	0.7524	0.9971	0.0029	0.2476	2.8627	0.8941	0.7929
&\textcolor[rgb]{0,0,1}{0.5715}(\textcolor[rgb]{1,0,1}{0.7524} \textcolor[rgb]{1,0,0}{$\Uparrow$})            &\textcolor[rgb]{0,0,1}{0.9953}(\textcolor[rgb]{1,0,1}{0.9971} \textcolor[rgb]{1,0,0}{$\Uparrow$})
&\textcolor[rgb]{0,0,1}{0.0047}(\textcolor[rgb]{1,0,1}{0.0029} \textcolor[rgb]{1,0,0}{$\Uparrow$})            &\textcolor[rgb]{0,0,1}{0.4285}(\textcolor[rgb]{1,0,1}{0.2476} \textcolor[rgb]{1,0,0}{$\Uparrow$})
&\textcolor[rgb]{0,0,1}{4.0811}(\textcolor[rgb]{1,0,1}{2.8627} \textcolor[rgb]{1,0,0}{$\Uparrow$})
&\textcolor[rgb]{0,0,1}{0.8174}(\textcolor[rgb]{1,0,1}{0.8941} \textcolor[rgb]{1,0,0}{$\Uparrow$})            &\textcolor[rgb]{0,0,1}{0.6068}(\textcolor[rgb]{1,0,1}{0.7929} \textcolor[rgb]{1,0,0}{$\Uparrow31\%$})\\

shadow  %shadow :	0.9660	0.9966	0.0034	0.0340	0.4710	0.9351	0.9500
&\textcolor[rgb]{0,0,1}{0.9441}(\textcolor[rgb]{1,0,1}{0.9660} \textcolor[rgb]{1,0,0}{$\Uparrow$})            &\textcolor[rgb]{0,0,1}{0.9920}(\textcolor[rgb]{1,0,1}{0.9966} \textcolor[rgb]{1,0,0}{$\Uparrow$})
&\textcolor[rgb]{0,0,1}{0.0080}(\textcolor[rgb]{1,0,1}{0.0034} \textcolor[rgb]{1,0,0}{$\Uparrow$})            &\textcolor[rgb]{0,0,1}{0.0559}(\textcolor[rgb]{1,0,1}{0.0340} \textcolor[rgb]{1,0,0}{$\Uparrow$})
&\textcolor[rgb]{0,0,1}{1.0018}(\textcolor[rgb]{1,0,1}{0.4710} \textcolor[rgb]{1,0,0}{$\Uparrow$})
&\textcolor[rgb]{0,0,1}{0.8645}(\textcolor[rgb]{1,0,1}{0.9351} \textcolor[rgb]{1,0,0}{$\Uparrow$})            &\textcolor[rgb]{0,0,1}{0.8995}(\textcolor[rgb]{1,0,1}{0.9500} \textcolor[rgb]{1,0,0}{$\Uparrow6\%$})\\

thermal %thermal :	0.8547	0.9886	0.0114	0.1453	1.5637	0.8478	0.8488
&\textcolor[rgb]{0,0,1}{0.8166}(\textcolor[rgb]{1,0,1}{0.8547} \textcolor[rgb]{1,0,0}{$\Uparrow$})            &\textcolor[rgb]{0,0,1}{0.9910}(\textcolor[rgb]{1,0,1}{0.9886} $\Downarrow$)
&\textcolor[rgb]{0,0,1}{0.0090}(\textcolor[rgb]{1,0,1}{0.0114} $\Downarrow$)
&\textcolor[rgb]{0,0,1}{0.1834}(\textcolor[rgb]{1,0,1}{0.1453} \textcolor[rgb]{1,0,0}{$\Uparrow$})
&\textcolor[rgb]{0,0,1}{1.9632}(\textcolor[rgb]{1,0,1}{1.5637} \textcolor[rgb]{1,0,0}{$\Uparrow$})
&\textcolor[rgb]{0,0,1}{0.8319}(\textcolor[rgb]{1,0,1}{0.8478} \textcolor[rgb]{1,0,0}{$\Uparrow$})              &\textcolor[rgb]{0,0,1}{0.8172}(\textcolor[rgb]{1,0,1}{0.8488} \textcolor[rgb]{1,0,0}{$\Uparrow4\%$})\\
\midrule
badWeather  %badWeat :	0.8033	0.9991	0.0009	0.1967	0.4556	0.9405	0.8628
&\textcolor[rgb]{0,0,1}{0.8082}(\textcolor[rgb]{1,0,1}{0.8033} $\Downarrow$)
&\textcolor[rgb]{0,0,1}{0.9991}(\textcolor[rgb]{1,0,1}{0.9991} \textcolor[rgb]{1,0,0}{$\Uparrow$})
&\textcolor[rgb]{0,0,1}{0.0009}(\textcolor[rgb]{1,0,1}{0.0009} \textcolor[rgb]{1,0,0}{$\Uparrow$})              &\textcolor[rgb]{0,0,1}{0.1918}(\textcolor[rgb]{1,0,1}{0.1967} $\Downarrow$)
&\textcolor[rgb]{0,0,1}{0.4807}(\textcolor[rgb]{1,0,1}{0.4556} \textcolor[rgb]{1,0,0}{$\Uparrow$})
&\textcolor[rgb]{0,0,1}{0.9183}(\textcolor[rgb]{1,0,1}{0.9405} \textcolor[rgb]{1,0,0}{$\Uparrow$})              &\textcolor[rgb]{0,0,1}{0.8577}(\textcolor[rgb]{1,0,1}{0.8628} \textcolor[rgb]{1,0,0}{$\Uparrow1\%$})\\

lowFramerate  %lowFram :	0.7218	0.9938	0.0062	0.2782	1.5281	0.6269	0.6249
&\textcolor[rgb]{0,0,1}{0.8267}(\textcolor[rgb]{1,0,1}{0.7218} $\Downarrow$)
&\textcolor[rgb]{0,0,1}{0.9937}(\textcolor[rgb]{1,0,1}{0.9938} \textcolor[rgb]{1,0,0}{$\Uparrow$})
&\textcolor[rgb]{0,0,1}{0.0063}(\textcolor[rgb]{1,0,1}{0.0062} \textcolor[rgb]{1,0,0}{$\Uparrow$})              &\textcolor[rgb]{0,0,1}{0.1733}(\textcolor[rgb]{1,0,1}{0.2782} $\Downarrow$)
&\textcolor[rgb]{0,0,1}{1.2054}(\textcolor[rgb]{1,0,1}{1.5281} $\Downarrow$)
&\textcolor[rgb]{0,0,1}{0.6475}(\textcolor[rgb]{1,0,1}{0.6269} $\Downarrow$)
&\textcolor[rgb]{0,0,1}{0.6659}(\textcolor[rgb]{1,0,1}{0.6249} $\Downarrow6\%$)\\

nightVideos  %nightVi :	0.4566	0.9915	0.0085	0.5434	1.9630	0.6048	0.4741
&\textcolor[rgb]{0,0,1}{0.5896}(\textcolor[rgb]{1,0,1}{0.4566} $\Downarrow$)            &\textcolor[rgb]{0,0,1}{0.9837}(\textcolor[rgb]{1,0,1}{0.9915} \textcolor[rgb]{1,0,0}{$\Uparrow$})
&\textcolor[rgb]{0,0,1}{0.0163}(\textcolor[rgb]{1,0,1}{0.0085} \textcolor[rgb]{1,0,0}{$\Uparrow$})            &\textcolor[rgb]{0,0,1}{0.4104}(\textcolor[rgb]{1,0,1}{0.5434} $\Downarrow$)         &\textcolor[rgb]{0,0,1}{2.5053}(\textcolor[rgb]{1,0,1}{1.9630} \textcolor[rgb]{1,0,0}{$\Uparrow$})
&\textcolor[rgb]{0,0,1}{0.4619}(\textcolor[rgb]{1,0,1}{0.6048} \textcolor[rgb]{1,0,0}{$\Uparrow$})            &\textcolor[rgb]{0,0,1}{0.4895}(\textcolor[rgb]{1,0,1}{0.4741} $\Downarrow1\%$)\\

PTZ   %PTZ :		0.7919	0.9911	0.0089	0.2081	1.0605	0.5070	0.5917
&\textcolor[rgb]{0,0,1}{0.8293}(\textcolor[rgb]{1,0,1}{0.7919} $\Downarrow$)
&\textcolor[rgb]{0,0,1}{0.9649}(\textcolor[rgb]{1,0,1}{0.9911} \textcolor[rgb]{1,0,0}{$\Uparrow$})
&\textcolor[rgb]{0,0,1}{0.0351}(\textcolor[rgb]{1,0,1}{0.0089} \textcolor[rgb]{1,0,0}{$\Uparrow$})            &\textcolor[rgb]{0,0,1}{0.1707}(\textcolor[rgb]{1,0,1}{0.2081} $\Downarrow$)
&\textcolor[rgb]{0,0,1}{3.6426}(\textcolor[rgb]{1,0,1}{1.0605} \textcolor[rgb]{1,0,0}{$\Uparrow$})
&\textcolor[rgb]{0,0,1}{0.3163}(\textcolor[rgb]{1,0,1}{0.5070} \textcolor[rgb]{1,0,0}{$\Uparrow$})            &\textcolor[rgb]{0,0,1}{0.3806}(\textcolor[rgb]{1,0,1}{0.5917} \textcolor[rgb]{1,0,0}{$\Uparrow55\%$})\\

turbulence  %turbule :	0.8553	0.9993	0.0007	0.1447	0.1654	0.8000	0.8120
&\textcolor[rgb]{0,0,1}{0.8213}(\textcolor[rgb]{1,0,1}{0.8553} \textcolor[rgb]{1,0,0}{$\Uparrow$})            &\textcolor[rgb]{0,0,1}{0.9998}(\textcolor[rgb]{1,0,1}{0.9993} $\Downarrow$)
&\textcolor[rgb]{0,0,1}{0.0002}(\textcolor[rgb]{1,0,1}{0.0007} $\Downarrow$)
&\textcolor[rgb]{0,0,1}{0.1787}(\textcolor[rgb]{1,0,1}{0.1447} \textcolor[rgb]{1,0,0}{$\Uparrow$})
&\textcolor[rgb]{0,0,1}{0.1373}(\textcolor[rgb]{1,0,1}{0.1654} $\Downarrow$)
&\textcolor[rgb]{0,0,1}{0.9318}(\textcolor[rgb]{1,0,1}{0.8000} $\Downarrow$)
&\textcolor[rgb]{0,0,1}{0.8689}(\textcolor[rgb]{1,0,1}{0.8120} $\Downarrow6\%$)\\
\midrule
\bf{Overall}  %Overall :	0.8118	0.9952	0.0048	0.1882	1.0874	0.8094	0.7892
&\textcolor[rgb]{0,0,1}{0.7968}(\textcolor[rgb]{1,0,1}{0.8118} \textcolor[rgb]{1,0,0}{$\Uparrow$})            &\textcolor[rgb]{0,0,1}{0.9916}(\textcolor[rgb]{1,0,1}{0.9952} \textcolor[rgb]{1,0,0}{$\Uparrow$})
&\textcolor[rgb]{0,0,1}{0.0084}(\textcolor[rgb]{1,0,1}{0.0048} \textcolor[rgb]{1,0,0}{$\Uparrow$})            &\textcolor[rgb]{0,0,1}{0.2032}(\textcolor[rgb]{1,0,1}{0.1882} \textcolor[rgb]{1,0,0}{$\Uparrow$})
&\textcolor[rgb]{0,0,1}{1.5895}(\textcolor[rgb]{1,0,1}{1.0874} \textcolor[rgb]{1,0,0}{$\Uparrow$})
&\textcolor[rgb]{0,0,1}{0.7658}(\textcolor[rgb]{1,0,1}{0.8094} \textcolor[rgb]{1,0,0}{$\Uparrow$})            &\textcolor[rgb]{0,0,1}{0.7420}(\textcolor[rgb]{1,0,1}{0.7892} \textcolor[rgb]{1,0,0}{$\Uparrow6\%$})\\
\bottomrule
\end{tabular}}
\begin{tablenotes}
\item[1] Note that blue entries indicate the original SuBSENSE results.  In parentheses, the purple entries indicate the RTSS$_{\text{SuBSENSE+PSPNet (N=2)}}$ results, and the arrows show the variation when compared to the original SuBSENSE results.
\end{tablenotes}
\label{resultsOfCD2014PSPNetN2}
\end{threeparttable}
\end{table*}

%N=3 【未完成】
\begin{table*}[!ht]
\centering
\begin{threeparttable}
\caption{Complete results of the RTSS framework with \textcolor[rgb]{0,0,1}{SuBSENSE} and \textcolor[rgb]{1,0,1}{PSPNet (N=3)}   on the CDnet 2014 dataset\tnote{1}}
{\tabcolsep2pt\begin{tabular}{cccccccc}
\toprule
Category        &Recall         &Specificity        &FPR            &FNR            &PWC            &Precision           &F-Measure\\
\midrule
baseline  %baseline :	0.9612	0.9978	0.0022	0.0388	0.3210	0.9447	0.9526
&\textcolor[rgb]{0,0,1}{0.9519}(\textcolor[rgb]{1,0,1}{0.9612} \textcolor[rgb]{1,0,0}{$\Uparrow$})
&\textcolor[rgb]{0,0,1}{0.9982}(\textcolor[rgb]{1,0,1}{0.9978} $\Downarrow$)
&\textcolor[rgb]{0,0,1}{0.0018}(\textcolor[rgb]{1,0,1}{0.0022} $\Downarrow$)
&\textcolor[rgb]{0,0,1}{0.0481}(\textcolor[rgb]{1,0,1}{0.0388} \textcolor[rgb]{1,0,0}{$\Uparrow$})
&\textcolor[rgb]{0,0,1}{0.3639}(\textcolor[rgb]{1,0,1}{0.3210} \textcolor[rgb]{1,0,0}{$\Uparrow$})
&\textcolor[rgb]{0,0,1}{0.9486}(\textcolor[rgb]{1,0,1}{0.9447} $\Downarrow$)
&\textcolor[rgb]{0,0,1}{0.9498}(\textcolor[rgb]{1,0,1}{0.9526} \textcolor[rgb]{1,0,0}{$\Uparrow 1\%$})\\

cameraJ   %cameraJ :	0.8090	0.9924	0.0076	0.1910	1.5926	0.8574	0.8178
&\textcolor[rgb]{0,0,1}{0.8319}(\textcolor[rgb]{1,0,1}{0.8090} $\Downarrow$)
&\textcolor[rgb]{0,0,1}{0.9901}(\textcolor[rgb]{1,0,1}{0.9924} \textcolor[rgb]{1,0,0}{$\Uparrow$})
&\textcolor[rgb]{0,0,1}{0.0099}(\textcolor[rgb]{1,0,1}{0.0076} \textcolor[rgb]{1,0,0}{$\Uparrow$})
&\textcolor[rgb]{0,0,1}{0.1681}(\textcolor[rgb]{1,0,1}{0.1910} $\Downarrow$)
&\textcolor[rgb]{0,0,1}{1.6937}(\textcolor[rgb]{1,0,1}{1.5926} \textcolor[rgb]{1,0,0}{$\Uparrow$})
&\textcolor[rgb]{0,0,1}{0.7944}(\textcolor[rgb]{1,0,1}{0.8574} \textcolor[rgb]{1,0,0}{$\Uparrow$})
&\textcolor[rgb]{0,0,1}{0.8096}(\textcolor[rgb]{1,0,1}{0.8178} \textcolor[rgb]{1,0,0}{$\Uparrow 1\%$})\\

dynamic   %dynamic :	0.9231	0.9992	0.0008	0.0769	0.1483	0.9304	0.9263
&\textcolor[rgb]{0,0,1}{0.7739}(\textcolor[rgb]{1,0,1}{0.9231} \textcolor[rgb]{1,0,0}{$\Uparrow$})
&\textcolor[rgb]{0,0,1}{0.9994}(\textcolor[rgb]{1,0,1}{0.9992} $\Downarrow$)
&\textcolor[rgb]{0,0,1}{0.0006}(\textcolor[rgb]{1,0,1}{0.0008} $\Downarrow$)
&\textcolor[rgb]{0,0,1}{0.2261}(\textcolor[rgb]{1,0,1}{0.0769} \textcolor[rgb]{1,0,0}{$\Uparrow$})
&\textcolor[rgb]{0,0,1}{0.4094}(\textcolor[rgb]{1,0,1}{0.1483} \textcolor[rgb]{1,0,0}{$\Uparrow$})
&\textcolor[rgb]{0,0,1}{0.8913}(\textcolor[rgb]{1,0,1}{0.9304} \textcolor[rgb]{1,0,0}{$\Uparrow$})
&\textcolor[rgb]{0,0,1}{0.8159}(\textcolor[rgb]{1,0,1}{0.9263} \textcolor[rgb]{1,0,0}{$\Uparrow13\%$})\\

intermittent  %intermi :	0.7405	0.9972	0.0028	0.2595	2.9163	0.8999	0.7892
&\textcolor[rgb]{0,0,1}{0.5715}(\textcolor[rgb]{1,0,1}{0.7405} \textcolor[rgb]{1,0,0}{$\Uparrow$})
&\textcolor[rgb]{0,0,1}{0.9953}(\textcolor[rgb]{1,0,1}{0.9972} \textcolor[rgb]{1,0,0}{$\Uparrow$})
&\textcolor[rgb]{0,0,1}{0.0047}(\textcolor[rgb]{1,0,1}{0.0028} \textcolor[rgb]{1,0,0}{$\Uparrow$})
&\textcolor[rgb]{0,0,1}{0.4285}(\textcolor[rgb]{1,0,1}{0.2595} \textcolor[rgb]{1,0,0}{$\Uparrow$})
&\textcolor[rgb]{0,0,1}{4.0811}(\textcolor[rgb]{1,0,1}{2.9163} \textcolor[rgb]{1,0,0}{$\Uparrow$})
&\textcolor[rgb]{0,0,1}{0.8174}(\textcolor[rgb]{1,0,1}{0.8999} \textcolor[rgb]{1,0,0}{$\Uparrow$})
&\textcolor[rgb]{0,0,1}{0.6068}(\textcolor[rgb]{1,0,1}{0.7892} \textcolor[rgb]{1,0,0}{$\Uparrow30\%$})\\

shadow  %shadow :	0.9552	0.9960	0.0040	0.0448	0.5801	0.9246	0.9394
&\textcolor[rgb]{0,0,1}{0.9441}(\textcolor[rgb]{1,0,1}{0.9552} \textcolor[rgb]{1,0,0}{$\Uparrow$})
&\textcolor[rgb]{0,0,1}{0.9920}(\textcolor[rgb]{1,0,1}{0.9960} \textcolor[rgb]{1,0,0}{$\Uparrow$})
&\textcolor[rgb]{0,0,1}{0.0080}(\textcolor[rgb]{1,0,1}{0.0040} \textcolor[rgb]{1,0,0}{$\Uparrow$})
&\textcolor[rgb]{0,0,1}{0.0559}(\textcolor[rgb]{1,0,1}{0.0448} \textcolor[rgb]{1,0,0}{$\Uparrow$})
&\textcolor[rgb]{0,0,1}{1.0018}(\textcolor[rgb]{1,0,1}{0.5801} \textcolor[rgb]{1,0,0}{$\Uparrow$})
&\textcolor[rgb]{0,0,1}{0.8645}(\textcolor[rgb]{1,0,1}{0.9246} \textcolor[rgb]{1,0,0}{$\Uparrow$})
&\textcolor[rgb]{0,0,1}{0.8995}(\textcolor[rgb]{1,0,1}{0.9394} \textcolor[rgb]{1,0,0}{$\Uparrow4\%$})\\

thermal  %thermal :	0.8527	0.9896	0.0104	0.1473	1.5115	0.8513	0.8496
&\textcolor[rgb]{0,0,1}{0.8166}(\textcolor[rgb]{1,0,1}{0.8527} \textcolor[rgb]{1,0,0}{$\Uparrow$})
&\textcolor[rgb]{0,0,1}{0.9910}(\textcolor[rgb]{1,0,1}{0.9896} $\Downarrow$)
&\textcolor[rgb]{0,0,1}{0.0090}(\textcolor[rgb]{1,0,1}{0.0104} $\Downarrow$)
&\textcolor[rgb]{0,0,1}{0.1834}(\textcolor[rgb]{1,0,1}{0.1473} \textcolor[rgb]{1,0,0}{$\Uparrow$})
&\textcolor[rgb]{0,0,1}{1.9632}(\textcolor[rgb]{1,0,1}{1.5115} \textcolor[rgb]{1,0,0}{$\Uparrow$})
&\textcolor[rgb]{0,0,1}{0.8319}(\textcolor[rgb]{1,0,1}{0.8513} \textcolor[rgb]{1,0,0}{$\Uparrow$})
&\textcolor[rgb]{0,0,1}{0.8172}(\textcolor[rgb]{1,0,1}{0.8496} \textcolor[rgb]{1,0,0}{$\Uparrow4\%$})\\
\midrule
badWeather  %badWeat :	0.7942	0.9988	0.0012	0.2058	0.4985	0.9336	0.8544
&\textcolor[rgb]{0,0,1}{0.8082}(\textcolor[rgb]{1,0,1}{0.7942} $\Downarrow$)
&\textcolor[rgb]{0,0,1}{0.9991}(\textcolor[rgb]{1,0,1}{0.9988} $\Downarrow$)
&\textcolor[rgb]{0,0,1}{0.0009}(\textcolor[rgb]{1,0,1}{0.0012} $\Downarrow$)
&\textcolor[rgb]{0,0,1}{0.1918}(\textcolor[rgb]{1,0,1}{0.2058} $\Downarrow$)
&\textcolor[rgb]{0,0,1}{0.4807}(\textcolor[rgb]{1,0,1}{0.4985} \textcolor[rgb]{1,0,0}{$\Uparrow$})
&\textcolor[rgb]{0,0,1}{0.9183}(\textcolor[rgb]{1,0,1}{0.9336} \textcolor[rgb]{1,0,0}{$\Uparrow$})
&\textcolor[rgb]{0,0,1}{0.8577}(\textcolor[rgb]{1,0,1}{0.8544} $\Downarrow 1\%$)\\

lowFramerate   %lowFram :	0.6815	0.9937	0.0063	0.3185	1.6851	0.6142	0.5958
&\textcolor[rgb]{0,0,1}{0.8267}(\textcolor[rgb]{1,0,1}{0.6815} $\Downarrow$)
&\textcolor[rgb]{0,0,1}{0.9937}(\textcolor[rgb]{1,0,1}{0.9937} \textcolor[rgb]{1,0,0}{$\Uparrow$})
&\textcolor[rgb]{0,0,1}{0.0063}(\textcolor[rgb]{1,0,1}{0.0063} \textcolor[rgb]{1,0,0}{$\Uparrow$})
&\textcolor[rgb]{0,0,1}{0.1733}(\textcolor[rgb]{1,0,1}{0.3185} $\Downarrow$)
&\textcolor[rgb]{0,0,1}{1.2054}(\textcolor[rgb]{1,0,1}{1.6851} $\Downarrow$)
&\textcolor[rgb]{0,0,1}{0.6475}(\textcolor[rgb]{1,0,1}{0.6142} $\Downarrow$)
&\textcolor[rgb]{0,0,1}{0.6659}(\textcolor[rgb]{1,0,1}{0.5958} $\Downarrow10\%$)\\

nightVideos  %nightVi :	0.4415	0.9916	0.0084	0.5585	1.9825	0.6025	0.4615
&\textcolor[rgb]{0,0,1}{0.5896}(\textcolor[rgb]{1,0,1}{0.4415} $\Downarrow$)
&\textcolor[rgb]{0,0,1}{0.9837}(\textcolor[rgb]{1,0,1}{0.9916} \textcolor[rgb]{1,0,0}{$\Uparrow$})
&\textcolor[rgb]{0,0,1}{0.0163}(\textcolor[rgb]{1,0,1}{0.0084} \textcolor[rgb]{1,0,0}{$\Uparrow$})
&\textcolor[rgb]{0,0,1}{0.4104}(\textcolor[rgb]{1,0,1}{0.5585} $\Downarrow$)
&\textcolor[rgb]{0,0,1}{2.5053}(\textcolor[rgb]{1,0,1}{1.9825} \textcolor[rgb]{1,0,0}{$\Uparrow$})
&\textcolor[rgb]{0,0,1}{0.4619}(\textcolor[rgb]{1,0,1}{0.6025} \textcolor[rgb]{1,0,0}{$\Uparrow$})
&\textcolor[rgb]{0,0,1}{0.4895}(\textcolor[rgb]{1,0,1}{0.4615} $\Downarrow1\%$)\\

PTZ  %PTZ :		0.7115	0.9910	0.0090	0.2885	1.1595	0.4771	0.5477
&\textcolor[rgb]{0,0,1}{0.8293}(\textcolor[rgb]{1,0,1}{0.7115} $\Downarrow$)
&\textcolor[rgb]{0,0,1}{0.9649}(\textcolor[rgb]{1,0,1}{0.9910} \textcolor[rgb]{1,0,0}{$\Uparrow$})
&\textcolor[rgb]{0,0,1}{0.0351}(\textcolor[rgb]{1,0,1}{0.0090} \textcolor[rgb]{1,0,0}{$\Uparrow$})
&\textcolor[rgb]{0,0,1}{0.1707}(\textcolor[rgb]{1,0,1}{0.2885} $\Downarrow$)
&\textcolor[rgb]{0,0,1}{3.6426}(\textcolor[rgb]{1,0,1}{1.1595} \textcolor[rgb]{1,0,0}{$\Uparrow$})
&\textcolor[rgb]{0,0,1}{0.3163}(\textcolor[rgb]{1,0,1}{0.4771} \textcolor[rgb]{1,0,0}{$\Uparrow$})
&\textcolor[rgb]{0,0,1}{0.3806}(\textcolor[rgb]{1,0,1}{0.5477} \textcolor[rgb]{1,0,0}{$\Uparrow43\%$})\\

turbulence  %turbule :	0.8552	0.9991	0.0009	0.1448	0.1795	0.7819	0.7962
&\textcolor[rgb]{0,0,1}{0.8213}(\textcolor[rgb]{1,0,1}{0.8552} \textcolor[rgb]{1,0,0}{$\Uparrow$})
&\textcolor[rgb]{0,0,1}{0.9998}(\textcolor[rgb]{1,0,1}{0.9991} $\Downarrow$)
&\textcolor[rgb]{0,0,1}{0.0002}(\textcolor[rgb]{1,0,1}{0.0009} $\Downarrow$)
&\textcolor[rgb]{0,0,1}{0.1787}(\textcolor[rgb]{1,0,1}{0.1448} \textcolor[rgb]{1,0,0}{$\Uparrow$})
&\textcolor[rgb]{0,0,1}{0.1373}(\textcolor[rgb]{1,0,1}{0.1795} $\Downarrow$)
&\textcolor[rgb]{0,0,1}{0.9318}(\textcolor[rgb]{1,0,1}{0.7819} $\Downarrow$)
&\textcolor[rgb]{0,0,1}{0.8689}(\textcolor[rgb]{1,0,1}{0.7962} $\Downarrow8\%$)\\
\midrule
\bf{Overall}  %Overall :	0.7932	0.9951	0.0049	0.2068	1.1432	0.8016	0.7755
&\textcolor[rgb]{0,0,1}{0.7968}(\textcolor[rgb]{1,0,1}{0.7932} $\Downarrow$)
&\textcolor[rgb]{0,0,1}{0.9916}(\textcolor[rgb]{1,0,1}{0.9951} \textcolor[rgb]{1,0,0}{$\Uparrow$})
&\textcolor[rgb]{0,0,1}{0.0084}(\textcolor[rgb]{1,0,1}{0.0049} \textcolor[rgb]{1,0,0}{$\Uparrow$})
&\textcolor[rgb]{0,0,1}{0.2032}(\textcolor[rgb]{1,0,1}{0.2068} $\Downarrow$)
&\textcolor[rgb]{0,0,1}{1.5895}(\textcolor[rgb]{1,0,1}{1.1432} \textcolor[rgb]{1,0,0}{$\Uparrow$})
&\textcolor[rgb]{0,0,1}{0.7658}(\textcolor[rgb]{1,0,1}{0.8016} \textcolor[rgb]{1,0,0}{$\Uparrow$})
&\textcolor[rgb]{0,0,1}{0.7420}(\textcolor[rgb]{1,0,1}{0.7755} \textcolor[rgb]{1,0,0}{$\Uparrow4\%$})\\
\bottomrule
\end{tabular}}
\begin{tablenotes}
\item[1] Note that blue entries indicate the original SuBSENSE results.  In parentheses, the purple entries indicate the RTSS$_{\text{SuBSENSE+PSPNet (N=3)}}$ results, and the arrows show the variation when compared to the original SuBSENSE results.
\end{tablenotes}
\label{resultsOfCD2014PSPNetN3}
\end{threeparttable}
\end{table*}

%ERFNet
\begin{table*}[!ht]
\centering
\begin{threeparttable}
\caption{Complete results of the RTSS framework with \textcolor[rgb]{0,0,1}{SuBSENSE} and \textcolor[rgb]{1,0,1}{ERFNet}   on the CDnet 2014 dataset\tnote{1}}
{\tabcolsep2pt\begin{tabular}{cccccccc}
\toprule
Category        &Recall         &Specificity        &FPR            &FNR            &PWC            &Precision           &F-Measure\\
\midrule
baseline  %baseline :	0.9623	0.9977	0.0023	0.0377	0.3571	0.9483	0.9549
&\textcolor[rgb]{0,0,1}{0.9519}(\textcolor[rgb]{1,0,1}{0.9623} \textcolor[rgb]{1,0,0}{$\Uparrow$})
&\textcolor[rgb]{0,0,1}{0.9982}(\textcolor[rgb]{1,0,1}{0.9977} $\Downarrow$)
&\textcolor[rgb]{0,0,1}{0.0018}(\textcolor[rgb]{1,0,1}{0.0023} $\Downarrow$)
&\textcolor[rgb]{0,0,1}{0.0481}(\textcolor[rgb]{1,0,1}{0.0377} \textcolor[rgb]{1,0,0}{$\Uparrow$})
&\textcolor[rgb]{0,0,1}{0.3639}(\textcolor[rgb]{1,0,1}{0.3571} \textcolor[rgb]{1,0,0}{$\Uparrow$})
&\textcolor[rgb]{0,0,1}{0.9486}(\textcolor[rgb]{1,0,1}{0.9483} $\Downarrow$)
&\textcolor[rgb]{0,0,1}{0.9498}(\textcolor[rgb]{1,0,1}{0.9549} \textcolor[rgb]{1,0,0}{$\Uparrow 1\%$})\\

cameraJ   %cameraJ :	0.7953	0.9929	0.0071	0.2047	1.6554	0.8457	0.8104
&\textcolor[rgb]{0,0,1}{0.8319}(\textcolor[rgb]{1,0,1}{0.7953} $\Downarrow$)
&\textcolor[rgb]{0,0,1}{0.9901}(\textcolor[rgb]{1,0,1}{0.9929} \textcolor[rgb]{1,0,0}{$\Uparrow$})
&\textcolor[rgb]{0,0,1}{0.0099}(\textcolor[rgb]{1,0,1}{0.0071} \textcolor[rgb]{1,0,0}{$\Uparrow$})
&\textcolor[rgb]{0,0,1}{0.1681}(\textcolor[rgb]{1,0,1}{0.2047} $\Downarrow$)
&\textcolor[rgb]{0,0,1}{1.6937}(\textcolor[rgb]{1,0,1}{1.6554} \textcolor[rgb]{1,0,0}{$\Uparrow$})
&\textcolor[rgb]{0,0,1}{0.7944}(\textcolor[rgb]{1,0,1}{0.8457} \textcolor[rgb]{1,0,0}{$\Uparrow$})
&\textcolor[rgb]{0,0,1}{0.8096}(\textcolor[rgb]{1,0,1}{0.8104} \textcolor[rgb]{1,0,0}{$\Uparrow 1\%$})\\

dynamic  %dynamic :	0.8693	0.9995	0.0005	0.1307	0.1950	0.9243	0.8908
&\textcolor[rgb]{0,0,1}{0.7739}(\textcolor[rgb]{1,0,1}{0.8693} \textcolor[rgb]{1,0,0}{$\Uparrow$})
&\textcolor[rgb]{0,0,1}{0.9994}(\textcolor[rgb]{1,0,1}{0.9995} \textcolor[rgb]{1,0,0}{$\Uparrow$})
&\textcolor[rgb]{0,0,1}{0.0006}(\textcolor[rgb]{1,0,1}{0.0005} \textcolor[rgb]{1,0,0}{$\Uparrow$})
&\textcolor[rgb]{0,0,1}{0.2261}(\textcolor[rgb]{1,0,1}{0.1307} \textcolor[rgb]{1,0,0}{$\Uparrow$})
&\textcolor[rgb]{0,0,1}{0.4094}(\textcolor[rgb]{1,0,1}{0.1950} \textcolor[rgb]{1,0,0}{$\Uparrow$})
&\textcolor[rgb]{0,0,1}{0.8913}(\textcolor[rgb]{1,0,1}{0.9243} \textcolor[rgb]{1,0,0}{$\Uparrow$})
&\textcolor[rgb]{0,0,1}{0.8159}(\textcolor[rgb]{1,0,1}{0.8908} \textcolor[rgb]{1,0,0}{$\Uparrow 9\%$})\\

intermittent  %intermi :	0.7275	0.9974	0.0026	0.2725	2.9674	0.8884	0.7708
&\textcolor[rgb]{0,0,1}{0.5715}(\textcolor[rgb]{1,0,1}{0.7275} \textcolor[rgb]{1,0,0}{$\Uparrow$})
&\textcolor[rgb]{0,0,1}{0.9953}(\textcolor[rgb]{1,0,1}{0.9974} \textcolor[rgb]{1,0,0}{$\Uparrow$})
&\textcolor[rgb]{0,0,1}{0.0047}(\textcolor[rgb]{1,0,1}{0.0026} \textcolor[rgb]{1,0,0}{$\Uparrow$})
&\textcolor[rgb]{0,0,1}{0.4285}(\textcolor[rgb]{1,0,1}{0.2725} \textcolor[rgb]{1,0,0}{$\Uparrow$})
&\textcolor[rgb]{0,0,1}{4.0811}(\textcolor[rgb]{1,0,1}{2.9674} \textcolor[rgb]{1,0,0}{$\Uparrow$})
&\textcolor[rgb]{0,0,1}{0.8174}(\textcolor[rgb]{1,0,1}{0.8884} \textcolor[rgb]{1,0,0}{$\Uparrow$})
&\textcolor[rgb]{0,0,1}{0.6068}(\textcolor[rgb]{1,0,1}{0.7708} \textcolor[rgb]{1,0,0}{$\Uparrow27\%$})\\

shadow  %shadow :	0.9657	0.9936	0.0064	0.0343	0.7609	0.8955	0.9276
&\textcolor[rgb]{0,0,1}{0.9441}(\textcolor[rgb]{1,0,1}{0.9657} \textcolor[rgb]{1,0,0}{$\Uparrow$})
&\textcolor[rgb]{0,0,1}{0.9920}(\textcolor[rgb]{1,0,1}{0.9936} \textcolor[rgb]{1,0,0}{$\Uparrow$})
&\textcolor[rgb]{0,0,1}{0.0080}(\textcolor[rgb]{1,0,1}{0.0064} \textcolor[rgb]{1,0,0}{$\Uparrow$})
&\textcolor[rgb]{0,0,1}{0.0559}(\textcolor[rgb]{1,0,1}{0.0343} \textcolor[rgb]{1,0,0}{$\Uparrow$})
&\textcolor[rgb]{0,0,1}{1.0018}(\textcolor[rgb]{1,0,1}{0.7609} \textcolor[rgb]{1,0,0}{$\Uparrow$})
&\textcolor[rgb]{0,0,1}{0.8645}(\textcolor[rgb]{1,0,1}{0.8955} \textcolor[rgb]{1,0,0}{$\Uparrow$})
&\textcolor[rgb]{0,0,1}{0.8995}(\textcolor[rgb]{1,0,1}{0.9276} \textcolor[rgb]{1,0,0}{$\Uparrow 3\%$})\\

thermal %thermal :	0.5398	0.9940	0.0060	0.4602	3.0182	0.8174	0.6298
&\textcolor[rgb]{0,0,1}{0.8166}(\textcolor[rgb]{1,0,1}{0.5398} $\Downarrow$)
&\textcolor[rgb]{0,0,1}{0.9910}(\textcolor[rgb]{1,0,1}{0.9940} \textcolor[rgb]{1,0,0}{$\Uparrow$})
&\textcolor[rgb]{0,0,1}{0.0090}(\textcolor[rgb]{1,0,1}{0.0060} \textcolor[rgb]{1,0,0}{$\Uparrow$})
&\textcolor[rgb]{0,0,1}{0.1834}(\textcolor[rgb]{1,0,1}{0.4602} $\Downarrow$)
&\textcolor[rgb]{0,0,1}{1.9632}(\textcolor[rgb]{1,0,1}{3.0182} $\Downarrow$)
&\textcolor[rgb]{0,0,1}{0.8319}(\textcolor[rgb]{1,0,1}{0.8174} $\Downarrow$)
&\textcolor[rgb]{0,0,1}{0.8172}(\textcolor[rgb]{1,0,1}{0.6298} $\Downarrow22\%$)\\
\midrule
badWeather %badWeat :	0.7743	0.9994	0.0006	0.2257	0.5371	0.9413	0.8474
&\textcolor[rgb]{0,0,1}{0.8082}(\textcolor[rgb]{1,0,1}{0.7743} $\Downarrow$)
&\textcolor[rgb]{0,0,1}{0.9991}(\textcolor[rgb]{1,0,1}{0.9994} \textcolor[rgb]{1,0,0}{$\Uparrow$})
&\textcolor[rgb]{0,0,1}{0.0009}(\textcolor[rgb]{1,0,1}{0.0006} \textcolor[rgb]{1,0,0}{$\Uparrow$})
&\textcolor[rgb]{0,0,1}{0.1918}(\textcolor[rgb]{1,0,1}{0.2257} $\Downarrow$)
&\textcolor[rgb]{0,0,1}{0.4807}(\textcolor[rgb]{1,0,1}{0.5371} $\Downarrow$)
&\textcolor[rgb]{0,0,1}{0.9183}(\textcolor[rgb]{1,0,1}{0.9413} \textcolor[rgb]{1,0,0}{$\Uparrow$})
&\textcolor[rgb]{0,0,1}{0.8577}(\textcolor[rgb]{1,0,1}{0.8474} $\Downarrow1\%$)\\

lowFramerate %lowFram :	0.8130	0.9933	0.0067	0.1870	1.3043	0.6555	0.6728
&\textcolor[rgb]{0,0,1}{0.8267}(\textcolor[rgb]{1,0,1}{0.8130} $\Downarrow$)
&\textcolor[rgb]{0,0,1}{0.9937}(\textcolor[rgb]{1,0,1}{0.9933} $\Downarrow$)
&\textcolor[rgb]{0,0,1}{0.0063}(\textcolor[rgb]{1,0,1}{0.0067} $\Downarrow$)
&\textcolor[rgb]{0,0,1}{0.1733}(\textcolor[rgb]{1,0,1}{0.1870} $\Downarrow$)
&\textcolor[rgb]{0,0,1}{1.2054}(\textcolor[rgb]{1,0,1}{1.3043} $\Downarrow$)
&\textcolor[rgb]{0,0,1}{0.6475}(\textcolor[rgb]{1,0,1}{0.6555} \textcolor[rgb]{1,0,0}{$\Uparrow$})
&\textcolor[rgb]{0,0,1}{0.6659}(\textcolor[rgb]{1,0,1}{0.6728} \textcolor[rgb]{1,0,0}{$\Uparrow1\%$})\\

nightVideos %nightVi :	0.4895	0.9939	0.0061	0.5105	1.6965	0.6196	0.5276
&\textcolor[rgb]{0,0,1}{0.5896}(\textcolor[rgb]{1,0,1}{0.4895} $\Downarrow$)
&\textcolor[rgb]{0,0,1}{0.9837}(\textcolor[rgb]{1,0,1}{0.9939} \textcolor[rgb]{1,0,0}{$\Uparrow$})
&\textcolor[rgb]{0,0,1}{0.0163}(\textcolor[rgb]{1,0,1}{0.0061} \textcolor[rgb]{1,0,0}{$\Uparrow$})
&\textcolor[rgb]{0,0,1}{0.4104}(\textcolor[rgb]{1,0,1}{0.5105} $\Downarrow$)
&\textcolor[rgb]{0,0,1}{2.5053}(\textcolor[rgb]{1,0,1}{1.6965} \textcolor[rgb]{1,0,0}{$\Uparrow$})
&\textcolor[rgb]{0,0,1}{0.4619}(\textcolor[rgb]{1,0,1}{0.6196} \textcolor[rgb]{1,0,0}{$\Uparrow$})
&\textcolor[rgb]{0,0,1}{0.4895}(\textcolor[rgb]{1,0,1}{0.5276} \textcolor[rgb]{1,0,0}{$\Uparrow1\%$})\\

PTZ  %PTZ :	0.8108	0.9864	0.0136	0.1892	1.5402	0.4056	0.4965
&\textcolor[rgb]{0,0,1}{0.8293}(\textcolor[rgb]{1,0,1}{0.8108} $\Downarrow$)
&\textcolor[rgb]{0,0,1}{0.9649}(\textcolor[rgb]{1,0,1}{0.9864} \textcolor[rgb]{1,0,0}{$\Uparrow$})
&\textcolor[rgb]{0,0,1}{0.0351}(\textcolor[rgb]{1,0,1}{0.0136} \textcolor[rgb]{1,0,0}{$\Uparrow$})
&\textcolor[rgb]{0,0,1}{0.1707}(\textcolor[rgb]{1,0,1}{0.1892} $\Downarrow$)
&\textcolor[rgb]{0,0,1}{3.6426}(\textcolor[rgb]{1,0,1}{1.5402} \textcolor[rgb]{1,0,0}{$\Uparrow$})
&\textcolor[rgb]{0,0,1}{0.3163}(\textcolor[rgb]{1,0,1}{0.4056} \textcolor[rgb]{1,0,0}{$\Uparrow$})
&\textcolor[rgb]{0,0,1}{0.3806}(\textcolor[rgb]{1,0,1}{0.4965} \textcolor[rgb]{1,0,0}{$\Uparrow31\%$})\\

turbulence  %turbule :	0.7804	0.9981	0.0019	0.2196	0.2909	0.8399	0.7989
&\textcolor[rgb]{0,0,1}{0.8213}(\textcolor[rgb]{1,0,1}{0.7804} $\Downarrow$)
&\textcolor[rgb]{0,0,1}{0.9998}(\textcolor[rgb]{1,0,1}{0.9981} $\Downarrow$)
&\textcolor[rgb]{0,0,1}{0.0002}(\textcolor[rgb]{1,0,1}{0.0019} $\Downarrow$)
&\textcolor[rgb]{0,0,1}{0.1787}(\textcolor[rgb]{1,0,1}{0.2196} $\Downarrow$)
&\textcolor[rgb]{0,0,1}{0.1373}(\textcolor[rgb]{1,0,1}{0.2909} $\Downarrow$)
&\textcolor[rgb]{0,0,1}{0.9318}(\textcolor[rgb]{1,0,1}{0.8399} $\Downarrow$)
&\textcolor[rgb]{0,0,1}{0.8689}(\textcolor[rgb]{1,0,1}{0.7989} $\Downarrow8\%$)\\
\midrule
\bf{Overall}  %Overall :	0.7753	0.9951	0.0049	0.2247	1.3021	0.7983	0.7570
&\textcolor[rgb]{0,0,1}{0.7968}(\textcolor[rgb]{1,0,1}{0.7753} $\Downarrow$)
&\textcolor[rgb]{0,0,1}{0.9916}(\textcolor[rgb]{1,0,1}{0.9951} \textcolor[rgb]{1,0,0}{$\Uparrow$})
&\textcolor[rgb]{0,0,1}{0.0084}(\textcolor[rgb]{1,0,1}{0.0049} \textcolor[rgb]{1,0,0}{$\Uparrow$})
&\textcolor[rgb]{0,0,1}{0.2032}(\textcolor[rgb]{1,0,1}{0.2247} $\Downarrow$)
&\textcolor[rgb]{0,0,1}{1.5895}(\textcolor[rgb]{1,0,1}{1.3021} \textcolor[rgb]{1,0,0}{$\Uparrow$})
&\textcolor[rgb]{0,0,1}{0.7658}(\textcolor[rgb]{1,0,1}{0.7983} \textcolor[rgb]{1,0,0}{$\Uparrow$})
&\textcolor[rgb]{0,0,1}{0.7420}(\textcolor[rgb]{1,0,1}{0.7570} \textcolor[rgb]{1,0,0}{$\Uparrow2\%$})\\
\bottomrule
\end{tabular}}
\begin{tablenotes}
\item[1] Note that blue entries indicate the original SuBSENSE results.  In parentheses, the purple entries indicate the RTSS$_{\text{SuBSENSE+ERFNet}}$ results, and the arrows show the variation when compared to the original SuBSENSE results.
\end{tablenotes}
\label{resultsOfCD2014erfnet}
\end{threeparttable}
\end{table*}

%LinkNet 【完成】
\begin{table*}[!ht]
\centering
\begin{threeparttable}
\caption{Complete results of the RTSS framework with \textcolor[rgb]{0,0,1}{SuBSENSE} and \textcolor[rgb]{1,0,1}{LinkNet}   on the CDnet 2014 dataset\tnote{1}}
{\tabcolsep2pt\begin{tabular}{cccccccc}
\toprule
Category        &Recall         &Specificity        &FPR            &FNR            &PWC            &Precision           &F-Measure\\
\midrule
baseline  %baseline :	0.9564	0.9981	0.0019	0.0436	0.3434	0.9527	0.9541
&\textcolor[rgb]{0,0,1}{0.9519}(\textcolor[rgb]{1,0,1}{0.9564} \textcolor[rgb]{1,0,0}{$\Uparrow$})
&\textcolor[rgb]{0,0,1}{0.9982}(\textcolor[rgb]{1,0,1}{0.9981} $\Downarrow$)
&\textcolor[rgb]{0,0,1}{0.0018}(\textcolor[rgb]{1,0,1}{0.0019} $\Downarrow$)
&\textcolor[rgb]{0,0,1}{0.0481}(\textcolor[rgb]{1,0,1}{0.0436} \textcolor[rgb]{1,0,0}{$\Uparrow$})
&\textcolor[rgb]{0,0,1}{0.3639}(\textcolor[rgb]{1,0,1}{0.3434} \textcolor[rgb]{1,0,0}{$\Uparrow$})
&\textcolor[rgb]{0,0,1}{0.9486}(\textcolor[rgb]{1,0,1}{0.9527} \textcolor[rgb]{1,0,0}{$\Uparrow$})
&\textcolor[rgb]{0,0,1}{0.9498}(\textcolor[rgb]{1,0,1}{0.9541} \textcolor[rgb]{1,0,0}{$\Uparrow 1\%$})\\

cameraJ   %cameraJ :	0.	0.	0.	0.		0.	0.
&\textcolor[rgb]{0,0,1}{0.8319}(\textcolor[rgb]{1,0,1}{0.8180} $\Downarrow$)
&\textcolor[rgb]{0,0,1}{0.9901}(\textcolor[rgb]{1,0,1}{0.9915} \textcolor[rgb]{1,0,0}{$\Uparrow$})
&\textcolor[rgb]{0,0,1}{0.0099}(\textcolor[rgb]{1,0,1}{0.0085} \textcolor[rgb]{1,0,0}{$\Uparrow$})
&\textcolor[rgb]{0,0,1}{0.1681}(\textcolor[rgb]{1,0,1}{0.1820} $\Downarrow$)
&\textcolor[rgb]{0,0,1}{1.6937}(\textcolor[rgb]{1,0,1}{1.6517} \textcolor[rgb]{1,0,0}{$\Uparrow$})
&\textcolor[rgb]{0,0,1}{0.7944}(\textcolor[rgb]{1,0,1}{0.8241} \textcolor[rgb]{1,0,0}{$\Uparrow$})
&\textcolor[rgb]{0,0,1}{0.8096}(\textcolor[rgb]{1,0,1}{0.8118} \textcolor[rgb]{1,0,0}{$\Uparrow 1\%$})\\

dynamic   %dynamic :	0.	0.	0.	0.	0.	0.	0.
&\textcolor[rgb]{0,0,1}{0.7739}(\textcolor[rgb]{1,0,1}{0.8384} \textcolor[rgb]{1,0,0}{$\Uparrow$})
&\textcolor[rgb]{0,0,1}{0.9994}(\textcolor[rgb]{1,0,1}{0.9992} $\Downarrow$)
&\textcolor[rgb]{0,0,1}{0.0006}(\textcolor[rgb]{1,0,1}{0.0008} $\Downarrow$)
&\textcolor[rgb]{0,0,1}{0.2261}(\textcolor[rgb]{1,0,1}{0.1616} \textcolor[rgb]{1,0,0}{$\Uparrow$})
&\textcolor[rgb]{0,0,1}{0.4094}(\textcolor[rgb]{1,0,1}{0.2378} \textcolor[rgb]{1,0,0}{$\Uparrow$})
&\textcolor[rgb]{0,0,1}{0.8913}(\textcolor[rgb]{1,0,1}{0.8935} \textcolor[rgb]{1,0,0}{$\Uparrow$})
&\textcolor[rgb]{0,0,1}{0.8159}(\textcolor[rgb]{1,0,1}{0.8532} \textcolor[rgb]{1,0,0}{$\Uparrow 5\%$})\\

intermittent  %intermi :	0.7043	0.9958	0.0042	0.2957	3.3249	0.8430	0.7203
&\textcolor[rgb]{0,0,1}{0.5715}(\textcolor[rgb]{1,0,1}{0.7043} \textcolor[rgb]{1,0,0}{$\Uparrow$})
&\textcolor[rgb]{0,0,1}{0.9953}(\textcolor[rgb]{1,0,1}{0.9958} \textcolor[rgb]{1,0,0}{$\Uparrow$})
&\textcolor[rgb]{0,0,1}{0.0047}(\textcolor[rgb]{1,0,1}{0.0042} \textcolor[rgb]{1,0,0}{$\Uparrow$})
&\textcolor[rgb]{0,0,1}{0.4285}(\textcolor[rgb]{1,0,1}{0.2957} \textcolor[rgb]{1,0,0}{$\Uparrow$})
&\textcolor[rgb]{0,0,1}{4.0811}(\textcolor[rgb]{1,0,1}{3.3249} \textcolor[rgb]{1,0,0}{$\Uparrow$})
&\textcolor[rgb]{0,0,1}{0.8174}(\textcolor[rgb]{1,0,1}{0.8430} \textcolor[rgb]{1,0,0}{$\Uparrow$})
&\textcolor[rgb]{0,0,1}{0.6068}(\textcolor[rgb]{1,0,1}{0.7203} \textcolor[rgb]{1,0,0}{$\Uparrow19\%$})\\

shadow  %shadow :	0.9486	0.9936	0.0064	0.0514	0.8447	0.8863	0.9145
&\textcolor[rgb]{0,0,1}{0.9441}(\textcolor[rgb]{1,0,1}{0.9486} \textcolor[rgb]{1,0,0}{$\Uparrow$})
&\textcolor[rgb]{0,0,1}{0.9920}(\textcolor[rgb]{1,0,1}{0.9936} \textcolor[rgb]{1,0,0}{$\Uparrow$})
&\textcolor[rgb]{0,0,1}{0.0080}(\textcolor[rgb]{1,0,1}{0.0064} \textcolor[rgb]{1,0,0}{$\Uparrow$})
&\textcolor[rgb]{0,0,1}{0.0559}(\textcolor[rgb]{1,0,1}{0.0514} \textcolor[rgb]{1,0,0}{$\Uparrow$})
&\textcolor[rgb]{0,0,1}{1.0018}(\textcolor[rgb]{1,0,1}{0.8447} \textcolor[rgb]{1,0,0}{$\Uparrow$})
&\textcolor[rgb]{0,0,1}{0.8645}(\textcolor[rgb]{1,0,1}{0.8863} \textcolor[rgb]{1,0,0}{$\Uparrow$})
&\textcolor[rgb]{0,0,1}{0.8995}(\textcolor[rgb]{1,0,1}{0.9145} \textcolor[rgb]{1,0,0}{$\Uparrow2\%$})\\

thermal %thermal :	0.6903	0.9928	0.0072	0.3097	2.6456	0.8448	0.7548
&\textcolor[rgb]{0,0,1}{0.8166}(\textcolor[rgb]{1,0,1}{0.6903} $\Downarrow$)
&\textcolor[rgb]{0,0,1}{0.9910}(\textcolor[rgb]{1,0,1}{0.9928} \textcolor[rgb]{1,0,0}{$\Uparrow$})
&\textcolor[rgb]{0,0,1}{0.0090}(\textcolor[rgb]{1,0,1}{0.0072} \textcolor[rgb]{1,0,0}{$\Uparrow$})
&\textcolor[rgb]{0,0,1}{0.1834}(\textcolor[rgb]{1,0,1}{0.3097} $\Downarrow$)
&\textcolor[rgb]{0,0,1}{1.9632}(\textcolor[rgb]{1,0,1}{2.6456} $\Downarrow$)
&\textcolor[rgb]{0,0,1}{0.8319}(\textcolor[rgb]{1,0,1}{0.8448} \textcolor[rgb]{1,0,0}{$\Uparrow$})
&\textcolor[rgb]{0,0,1}{0.8172}(\textcolor[rgb]{1,0,1}{0.7548} $\Downarrow7\%$)\\
\midrule
badWeather  %badWeat :	0.8024	0.9993	0.0007	0.1976	0.4748	0.9378	0.8627
&\textcolor[rgb]{0,0,1}{0.8082}(\textcolor[rgb]{1,0,1}{0.8024} $\Downarrow$)
&\textcolor[rgb]{0,0,1}{0.9991}(\textcolor[rgb]{1,0,1}{0.9993} \textcolor[rgb]{1,0,0}{$\Uparrow$})
&\textcolor[rgb]{0,0,1}{0.0009}(\textcolor[rgb]{1,0,1}{0.0007} \textcolor[rgb]{1,0,0}{$\Uparrow$})
&\textcolor[rgb]{0,0,1}{0.1918}(\textcolor[rgb]{1,0,1}{0.1976} $\Downarrow$)
&\textcolor[rgb]{0,0,1}{0.4807}(\textcolor[rgb]{1,0,1}{0.4748} \textcolor[rgb]{1,0,0}{$\Uparrow$})
&\textcolor[rgb]{0,0,1}{0.9183}(\textcolor[rgb]{1,0,1}{0.9378} \textcolor[rgb]{1,0,0}{$\Uparrow$})
&\textcolor[rgb]{0,0,1}{0.8577}(\textcolor[rgb]{1,0,1}{0.8627} \textcolor[rgb]{1,0,0}{$\Uparrow1\%$})\\

lowFramerate  %lowFram :	0.7983	0.9956	0.0044	0.2017	1.1033	0.6813	0.6869
&\textcolor[rgb]{0,0,1}{0.8267}(\textcolor[rgb]{1,0,1}{0.7983} $\Downarrow$)
&\textcolor[rgb]{0,0,1}{0.9937}(\textcolor[rgb]{1,0,1}{0.9956} \textcolor[rgb]{1,0,0}{$\Uparrow$})
&\textcolor[rgb]{0,0,1}{0.0063}(\textcolor[rgb]{1,0,1}{0.0044} \textcolor[rgb]{1,0,0}{$\Uparrow$})
&\textcolor[rgb]{0,0,1}{0.1733}(\textcolor[rgb]{1,0,1}{0.2017} $\Downarrow$)
&\textcolor[rgb]{0,0,1}{1.2054}(\textcolor[rgb]{1,0,1}{1.1033} \textcolor[rgb]{1,0,0}{$\Uparrow$})
&\textcolor[rgb]{0,0,1}{0.6475}(\textcolor[rgb]{1,0,1}{0.6813} \textcolor[rgb]{1,0,0}{$\Uparrow$})
&\textcolor[rgb]{0,0,1}{0.6659}(\textcolor[rgb]{1,0,1}{0.6869} \textcolor[rgb]{1,0,0}{$\Uparrow 3\%$})\\

nightVideos  %nightVi :	0.5563	0.9875	0.0125	0.4437	2.1636	0.4992	0.5052
&\textcolor[rgb]{0,0,1}{0.5896}(\textcolor[rgb]{1,0,1}{0.5563} $\Downarrow$)
&\textcolor[rgb]{0,0,1}{0.9837}(\textcolor[rgb]{1,0,1}{0.9875} \textcolor[rgb]{1,0,0}{$\Uparrow$})
&\textcolor[rgb]{0,0,1}{0.0163}(\textcolor[rgb]{1,0,1}{0.0125} \textcolor[rgb]{1,0,0}{$\Uparrow$})
&\textcolor[rgb]{0,0,1}{0.4104}(\textcolor[rgb]{1,0,1}{0.4437} $\Downarrow$)
&\textcolor[rgb]{0,0,1}{2.5053}(\textcolor[rgb]{1,0,1}{2.1636} \textcolor[rgb]{1,0,0}{$\Uparrow$})
&\textcolor[rgb]{0,0,1}{0.4619}(\textcolor[rgb]{1,0,1}{0.4992} \textcolor[rgb]{1,0,0}{$\Uparrow$})
&\textcolor[rgb]{0,0,1}{0.4895}(\textcolor[rgb]{1,0,1}{0.5052} \textcolor[rgb]{1,0,0}{$\Uparrow 1\%$})\\

PTZ %PTZ :		0.8156	0.9846	0.0154	0.1844	1.7126	0.3796	0.4610
&\textcolor[rgb]{0,0,1}{0.8293}(\textcolor[rgb]{1,0,1}{0.8156} $\Downarrow$)
&\textcolor[rgb]{0,0,1}{0.9649}(\textcolor[rgb]{1,0,1}{0.9846} \textcolor[rgb]{1,0,0}{$\Uparrow$})
&\textcolor[rgb]{0,0,1}{0.0351}(\textcolor[rgb]{1,0,1}{0.0154} \textcolor[rgb]{1,0,0}{$\Uparrow$})
&\textcolor[rgb]{0,0,1}{0.1707}(\textcolor[rgb]{1,0,1}{0.1844} $\Downarrow$)
&\textcolor[rgb]{0,0,1}{3.6426}(\textcolor[rgb]{1,0,1}{1.7126} \textcolor[rgb]{1,0,0}{$\Uparrow$})
&\textcolor[rgb]{0,0,1}{0.3163}(\textcolor[rgb]{1,0,1}{0.3796} \textcolor[rgb]{1,0,0}{$\Uparrow$})
&\textcolor[rgb]{0,0,1}{0.3806}(\textcolor[rgb]{1,0,1}{0.4610} \textcolor[rgb]{1,0,0}{$\Uparrow21\%$})\\

turbulence %turbule :	0.7786	0.9997	0.0003	0.2214	0.1401	0.8917	0.8278
&\textcolor[rgb]{0,0,1}{0.8213}(\textcolor[rgb]{1,0,1}{0.7786} $\Downarrow$)
&\textcolor[rgb]{0,0,1}{0.9998}(\textcolor[rgb]{1,0,1}{0.9997} $\Downarrow$)
&\textcolor[rgb]{0,0,1}{0.0002}(\textcolor[rgb]{1,0,1}{0.0003} $\Downarrow$)
&\textcolor[rgb]{0,0,1}{0.1787}(\textcolor[rgb]{1,0,1}{0.2214} $\Downarrow$)
&\textcolor[rgb]{0,0,1}{0.1373}(\textcolor[rgb]{1,0,1}{0.1401} $\Downarrow$)
&\textcolor[rgb]{0,0,1}{0.9318}(\textcolor[rgb]{1,0,1}{0.8917} $\Downarrow$)
&\textcolor[rgb]{0,0,1}{0.8689}(\textcolor[rgb]{1,0,1}{0.8278} $\Downarrow4\%$)\\
\midrule
\bf{Overall} %Overall :	0.7915	0.9943	0.0057	0.2085	1.3311	0.7849	0.7593
&\textcolor[rgb]{0,0,1}{0.7968}(\textcolor[rgb]{1,0,1}{0.7915} $\Downarrow$)
&\textcolor[rgb]{0,0,1}{0.9916}(\textcolor[rgb]{1,0,1}{0.9943} \textcolor[rgb]{1,0,0}{$\Uparrow$})
&\textcolor[rgb]{0,0,1}{0.0084}(\textcolor[rgb]{1,0,1}{0.0057} \textcolor[rgb]{1,0,0}{$\Uparrow$})
&\textcolor[rgb]{0,0,1}{0.2032}(\textcolor[rgb]{1,0,1}{0.2085} \textcolor[rgb]{1,0,0}{$\Uparrow$})
&\textcolor[rgb]{0,0,1}{1.5895}(\textcolor[rgb]{1,0,1}{1.3311} \textcolor[rgb]{1,0,0}{$\Uparrow$})
&\textcolor[rgb]{0,0,1}{0.7658}(\textcolor[rgb]{1,0,1}{0.7849} \textcolor[rgb]{1,0,0}{$\Uparrow$})
&\textcolor[rgb]{0,0,1}{0.7420}(\textcolor[rgb]{1,0,1}{0.7593} \textcolor[rgb]{1,0,0}{$\Uparrow2\%$})\\
\bottomrule
\end{tabular}}
\begin{tablenotes}
\item[1] Note that blue entries indicate the original SuBSENSE results.  In parentheses, the purple entries indicate the RTSS$_{\text{SuBSENSE+LinkNet}}$ results, and the arrows show the variation when compared to the original SuBSENSE results.
\end{tablenotes}
\label{resultsOfCD2014linknet}
\end{threeparttable}
\end{table*}

%MFCN
\begin{table*}[!ht]
\centering
\begin{threeparttable}
\caption{Complete results of the RTSS framework with \textcolor[rgb]{0,0,1}{SuBSENSE} and \textcolor[rgb]{1,0,1}{MFCN}   on the CDnet 2014 dataset\tnote{1}}
{\tabcolsep2pt\begin{tabular}{cccccccc}
\toprule
Category        &Recall         &Specificity        &FPR            &FNR            &PWC            &Precision           &F-Measure\\
\midrule
baseline  %baseline :	0.9653	0.9999	0.0001	0.0347	0.1178	0.9950	0.9798
&\textcolor[rgb]{0,0,1}{0.9519}(\textcolor[rgb]{1,0,1}{0.9653} \textcolor[rgb]{1,0,0}{$\Uparrow$})
&\textcolor[rgb]{0,0,1}{0.9982}(\textcolor[rgb]{1,0,1}{0.9999} \textcolor[rgb]{1,0,0}{$\Uparrow$})
&\textcolor[rgb]{0,0,1}{0.0018}(\textcolor[rgb]{1,0,1}{0.0001} \textcolor[rgb]{1,0,0}{$\Uparrow$})
&\textcolor[rgb]{0,0,1}{0.0481}(\textcolor[rgb]{1,0,1}{0.0347} \textcolor[rgb]{1,0,0}{$\Uparrow$})
&\textcolor[rgb]{0,0,1}{0.3639}(\textcolor[rgb]{1,0,1}{0.1178} \textcolor[rgb]{1,0,0}{$\Uparrow$})
&\textcolor[rgb]{0,0,1}{0.9486}(\textcolor[rgb]{1,0,1}{0.9950} \textcolor[rgb]{1,0,0}{$\Uparrow$})
&\textcolor[rgb]{0,0,1}{0.9498}(\textcolor[rgb]{1,0,1}{0.9798} \textcolor[rgb]{1,0,0}{$\Uparrow 3\%$})\\

cameraJ   %cameraJ :	0.9563	0.9997	0.0003	0.0437	0.1979	0.9928	0.9740
&\textcolor[rgb]{0,0,1}{0.8319}(\textcolor[rgb]{1,0,1}{0.9563} \textcolor[rgb]{1,0,0}{$\Uparrow$})
&\textcolor[rgb]{0,0,1}{0.9901}(\textcolor[rgb]{1,0,1}{0.9997} \textcolor[rgb]{1,0,0}{$\Uparrow$})
&\textcolor[rgb]{0,0,1}{0.0099}(\textcolor[rgb]{1,0,1}{0.0003} \textcolor[rgb]{1,0,0}{$\Uparrow$})
&\textcolor[rgb]{0,0,1}{0.1681}(\textcolor[rgb]{1,0,1}{0.0437} \textcolor[rgb]{1,0,0}{$\Uparrow$})
&\textcolor[rgb]{0,0,1}{1.6937}(\textcolor[rgb]{1,0,1}{0.1979} \textcolor[rgb]{1,0,0}{$\Uparrow$})
&\textcolor[rgb]{0,0,1}{0.7944}(\textcolor[rgb]{1,0,1}{0.9928} \textcolor[rgb]{1,0,0}{$\Uparrow$})
&\textcolor[rgb]{0,0,1}{0.8096}(\textcolor[rgb]{1,0,1}{0.9740} \textcolor[rgb]{1,0,0}{$\Uparrow20\%$})\\

dynamic   %dynamic :	0.9583	1.0000	0.0000	0.0417	0.0412	0.9988	0.9781
&\textcolor[rgb]{0,0,1}{0.7739}(\textcolor[rgb]{1,0,1}{0.9583} \textcolor[rgb]{1,0,0}{$\Uparrow$})
&\textcolor[rgb]{0,0,1}{0.9994}(\textcolor[rgb]{1,0,1}{1.0000} \textcolor[rgb]{1,0,0}{$\Uparrow$})
&\textcolor[rgb]{0,0,1}{0.0006}(\textcolor[rgb]{1,0,1}{0.0000} \textcolor[rgb]{1,0,0}{$\Uparrow$})
&\textcolor[rgb]{0,0,1}{0.2261}(\textcolor[rgb]{1,0,1}{0.0417} \textcolor[rgb]{1,0,0}{$\Uparrow$})
&\textcolor[rgb]{0,0,1}{0.4094}(\textcolor[rgb]{1,0,1}{0.0412} \textcolor[rgb]{1,0,0}{$\Uparrow$})
&\textcolor[rgb]{0,0,1}{0.8913}(\textcolor[rgb]{1,0,1}{0.9988} \textcolor[rgb]{1,0,0}{$\Uparrow$})
&\textcolor[rgb]{0,0,1}{0.8159}(\textcolor[rgb]{1,0,1}{0.9781} \textcolor[rgb]{1,0,0}{$\Uparrow20\%$})\\

intermittent  %intermi :	0.8864	1.0000	0.0000	0.1136	0.8456	0.9994	0.9385
&\textcolor[rgb]{0,0,1}{0.5715}(\textcolor[rgb]{1,0,1}{0.8864} \textcolor[rgb]{1,0,0}{$\Uparrow$})
&\textcolor[rgb]{0,0,1}{0.9953}(\textcolor[rgb]{1,0,1}{1.0000} \textcolor[rgb]{1,0,0}{$\Uparrow$})
&\textcolor[rgb]{0,0,1}{0.0047}(\textcolor[rgb]{1,0,1}{0.0000} \textcolor[rgb]{1,0,0}{$\Uparrow$})
&\textcolor[rgb]{0,0,1}{0.4285}(\textcolor[rgb]{1,0,1}{0.1136} \textcolor[rgb]{1,0,0}{$\Uparrow$})
&\textcolor[rgb]{0,0,1}{4.0811}(\textcolor[rgb]{1,0,1}{0.8456} \textcolor[rgb]{1,0,0}{$\Uparrow$})
&\textcolor[rgb]{0,0,1}{0.8174}(\textcolor[rgb]{1,0,1}{0.9994} \textcolor[rgb]{1,0,0}{$\Uparrow$})
&\textcolor[rgb]{0,0,1}{0.6068}(\textcolor[rgb]{1,0,1}{0.9385} \textcolor[rgb]{1,0,0}{$\Uparrow55\%$})\\

shadow  %shadow :	0.9756	0.9998	0.0002	0.0244	0.1087	0.9965	0.9859
&\textcolor[rgb]{0,0,1}{0.9441}(\textcolor[rgb]{1,0,1}{0.9756} \textcolor[rgb]{1,0,0}{$\Uparrow$})
&\textcolor[rgb]{0,0,1}{0.9920}(\textcolor[rgb]{1,0,1}{0.9998} \textcolor[rgb]{1,0,0}{$\Uparrow$})
&\textcolor[rgb]{0,0,1}{0.0080}(\textcolor[rgb]{1,0,1}{0.0002} \textcolor[rgb]{1,0,0}{$\Uparrow$})
&\textcolor[rgb]{0,0,1}{0.0559}(\textcolor[rgb]{1,0,1}{0.0244} \textcolor[rgb]{1,0,0}{$\Uparrow$})
&\textcolor[rgb]{0,0,1}{1.0018}(\textcolor[rgb]{1,0,1}{0.1087} \textcolor[rgb]{1,0,0}{$\Uparrow$})
&\textcolor[rgb]{0,0,1}{0.8645}(\textcolor[rgb]{1,0,1}{0.9965} \textcolor[rgb]{1,0,0}{$\Uparrow$})
&\textcolor[rgb]{0,0,1}{0.8995}(\textcolor[rgb]{1,0,1}{0.9859} \textcolor[rgb]{1,0,0}{$\Uparrow10\%$})\\

thermal %thermal :	0.9470	0.9997	0.0003	0.0530	0.2581	0.9945	0.9699
&\textcolor[rgb]{0,0,1}{0.8166}(\textcolor[rgb]{1,0,1}{0.9470} \textcolor[rgb]{1,0,0}{$\Uparrow$})
&\textcolor[rgb]{0,0,1}{0.9910}(\textcolor[rgb]{1,0,1}{0.9997} \textcolor[rgb]{1,0,0}{$\Uparrow$})
&\textcolor[rgb]{0,0,1}{0.0090}(\textcolor[rgb]{1,0,1}{0.0003} \textcolor[rgb]{1,0,0}{$\Uparrow$})
&\textcolor[rgb]{0,0,1}{0.1834}(\textcolor[rgb]{1,0,1}{0.0530} \textcolor[rgb]{1,0,0}{$\Uparrow$})
&\textcolor[rgb]{0,0,1}{1.9632}(\textcolor[rgb]{1,0,1}{0.2581} \textcolor[rgb]{1,0,0}{$\Uparrow$})
&\textcolor[rgb]{0,0,1}{0.8319}(\textcolor[rgb]{1,0,1}{0.9945} \textcolor[rgb]{1,0,0}{$\Uparrow$})
&\textcolor[rgb]{0,0,1}{0.8172}(\textcolor[rgb]{1,0,1}{0.9699} \textcolor[rgb]{1,0,0}{$\Uparrow19\%$})\\
\midrule
badWeather  %badWeat :	0.9488	0.9999	0.0001	0.0512	0.1172	0.9913	0.9693
&\textcolor[rgb]{0,0,1}{0.8082}(\textcolor[rgb]{1,0,1}{0.9488} \textcolor[rgb]{1,0,0}{$\Uparrow$})
&\textcolor[rgb]{0,0,1}{0.9991}(\textcolor[rgb]{1,0,1}{0.9999} \textcolor[rgb]{1,0,0}{$\Uparrow$})
&\textcolor[rgb]{0,0,1}{0.0009}(\textcolor[rgb]{1,0,1}{0.0001} \textcolor[rgb]{1,0,0}{$\Uparrow$})
&\textcolor[rgb]{0,0,1}{0.1918}(\textcolor[rgb]{1,0,1}{0.0512} \textcolor[rgb]{1,0,0}{$\Uparrow$})
&\textcolor[rgb]{0,0,1}{0.4807}(\textcolor[rgb]{1,0,1}{0.1172} \textcolor[rgb]{1,0,0}{$\Uparrow$})
&\textcolor[rgb]{0,0,1}{0.9183}(\textcolor[rgb]{1,0,1}{0.9913} \textcolor[rgb]{1,0,0}{$\Uparrow$})
&\textcolor[rgb]{0,0,1}{0.8577}(\textcolor[rgb]{1,0,1}{0.9693} \textcolor[rgb]{1,0,0}{$\Uparrow13\%$})\\

lowFramerate %lowFram :	0.9476	0.9999	0.0001	0.0524	0.2100	0.8883	0.9048
&\textcolor[rgb]{0,0,1}{0.8267}(\textcolor[rgb]{1,0,1}{0.9476} \textcolor[rgb]{1,0,0}{$\Uparrow$})
&\textcolor[rgb]{0,0,1}{0.9937}(\textcolor[rgb]{1,0,1}{0.9999} \textcolor[rgb]{1,0,0}{$\Uparrow$})
&\textcolor[rgb]{0,0,1}{0.0063}(\textcolor[rgb]{1,0,1}{0.0001} \textcolor[rgb]{1,0,0}{$\Uparrow$})
&\textcolor[rgb]{0,0,1}{0.1733}(\textcolor[rgb]{1,0,1}{0.0524} \textcolor[rgb]{1,0,0}{$\Uparrow$})
&\textcolor[rgb]{0,0,1}{1.2054}(\textcolor[rgb]{1,0,1}{0.2100} \textcolor[rgb]{1,0,0}{$\Uparrow$})
&\textcolor[rgb]{0,0,1}{0.6475}(\textcolor[rgb]{1,0,1}{0.8883} \textcolor[rgb]{1,0,0}{$\Uparrow$})
&\textcolor[rgb]{0,0,1}{0.6659}(\textcolor[rgb]{1,0,1}{0.9048} \textcolor[rgb]{1,0,0}{$\Uparrow36\%$})\\

nightVideos  %nightVi :	0.8638	0.9996	0.0004	0.1362	0.2962	0.9845	0.9173
&\textcolor[rgb]{0,0,1}{0.5896}(\textcolor[rgb]{1,0,1}{0.8638} \textcolor[rgb]{1,0,0}{$\Uparrow$})
&\textcolor[rgb]{0,0,1}{0.9837}(\textcolor[rgb]{1,0,1}{0.9996} \textcolor[rgb]{1,0,0}{$\Uparrow$})
&\textcolor[rgb]{0,0,1}{0.0163}(\textcolor[rgb]{1,0,1}{0.0004} \textcolor[rgb]{1,0,0}{$\Uparrow$})
&\textcolor[rgb]{0,0,1}{0.4104}(\textcolor[rgb]{1,0,1}{0.1362} \textcolor[rgb]{1,0,0}{$\Uparrow$})
&\textcolor[rgb]{0,0,1}{2.5053}(\textcolor[rgb]{1,0,1}{0.2962} \textcolor[rgb]{1,0,0}{$\Uparrow$})
&\textcolor[rgb]{0,0,1}{0.4619}(\textcolor[rgb]{1,0,1}{0.9845} \textcolor[rgb]{1,0,0}{$\Uparrow$})
&\textcolor[rgb]{0,0,1}{0.4895}(\textcolor[rgb]{1,0,1}{0.9173} \textcolor[rgb]{1,0,0}{$\Uparrow87\%$})\\

PTZ %PTZ :		0.9304	0.9998	0.0002	0.0696	0.0594	0.9780	0.9516
&\textcolor[rgb]{0,0,1}{0.8293}(\textcolor[rgb]{1,0,1}{0.9304} \textcolor[rgb]{1,0,0}{$\Uparrow$})
&\textcolor[rgb]{0,0,1}{0.9649}(\textcolor[rgb]{1,0,1}{0.9998} \textcolor[rgb]{1,0,0}{$\Uparrow$})
&\textcolor[rgb]{0,0,1}{0.0351}(\textcolor[rgb]{1,0,1}{0.0002} \textcolor[rgb]{1,0,0}{$\Uparrow$})
&\textcolor[rgb]{0,0,1}{0.1707}(\textcolor[rgb]{1,0,1}{0.0696} \textcolor[rgb]{1,0,0}{$\Uparrow$})
&\textcolor[rgb]{0,0,1}{3.6426}(\textcolor[rgb]{1,0,1}{0.0594} \textcolor[rgb]{1,0,0}{$\Uparrow$})
&\textcolor[rgb]{0,0,1}{0.3163}(\textcolor[rgb]{1,0,1}{0.9780} \textcolor[rgb]{1,0,0}{$\Uparrow$})
&\textcolor[rgb]{0,0,1}{0.3806}(\textcolor[rgb]{1,0,1}{0.9516} \textcolor[rgb]{1,0,0}{$\Uparrow150\%$})\\

turbulence %turbule :	0.9301	0.9999	0.0001	0.0699	0.0393	0.9891	0.9582
&\textcolor[rgb]{0,0,1}{0.8213}(\textcolor[rgb]{1,0,1}{0.9301} \textcolor[rgb]{1,0,0}{$\Uparrow$})
&\textcolor[rgb]{0,0,1}{0.9998}(\textcolor[rgb]{1,0,1}{0.9999} \textcolor[rgb]{1,0,0}{$\Uparrow$})
&\textcolor[rgb]{0,0,1}{0.0002}(\textcolor[rgb]{1,0,1}{0.0001} \textcolor[rgb]{1,0,0}{$\Uparrow$})
&\textcolor[rgb]{0,0,1}{0.1787}(\textcolor[rgb]{1,0,1}{0.0699} \textcolor[rgb]{1,0,0}{$\Uparrow$})
&\textcolor[rgb]{0,0,1}{0.1373}(\textcolor[rgb]{1,0,1}{0.0393} \textcolor[rgb]{1,0,0}{$\Uparrow$})
&\textcolor[rgb]{0,0,1}{0.9318}(\textcolor[rgb]{1,0,1}{0.9891} \textcolor[rgb]{1,0,0}{$\Uparrow$})
&\textcolor[rgb]{0,0,1}{0.8689}(\textcolor[rgb]{1,0,1}{0.9582} \textcolor[rgb]{1,0,0}{$\Uparrow10\%$})\\
\midrule
\bf{Overall}  %Overall :	0.9372	0.9998	0.0002	0.0628	0.2083	0.9826	0.9570
&\textcolor[rgb]{0,0,1}{0.7968}(\textcolor[rgb]{1,0,1}{0.9372} \textcolor[rgb]{1,0,0}{$\Uparrow$})
&\textcolor[rgb]{0,0,1}{0.9916}(\textcolor[rgb]{1,0,1}{0.9998} \textcolor[rgb]{1,0,0}{$\Uparrow$})
&\textcolor[rgb]{0,0,1}{0.0084}(\textcolor[rgb]{1,0,1}{0.0002} \textcolor[rgb]{1,0,0}{$\Uparrow$})
&\textcolor[rgb]{0,0,1}{0.2032}(\textcolor[rgb]{1,0,1}{0.0628} \textcolor[rgb]{1,0,0}{$\Uparrow$})
&\textcolor[rgb]{0,0,1}{1.5895}(\textcolor[rgb]{1,0,1}{0.2083} \textcolor[rgb]{1,0,0}{$\Uparrow$})
&\textcolor[rgb]{0,0,1}{0.7658}(\textcolor[rgb]{1,0,1}{0.9826} \textcolor[rgb]{1,0,0}{$\Uparrow$})
&\textcolor[rgb]{0,0,1}{0.7420}(\textcolor[rgb]{1,0,1}{0.9570} \textcolor[rgb]{1,0,0}{$\Uparrow29\%$})\\
\bottomrule
\end{tabular}}
\begin{tablenotes}
\item[1] Note that blue entries indicate the original SuBSENSE results.  In parentheses, the purple entries indicate the RTSS$_{\text{SuBSENSE+MFCN}}$ results, and the arrows show the variation when compared to the original SuBSENSE results.
\end{tablenotes}
\label{resultsOfCD2014mfcn}
\end{threeparttable}
\end{table*}

\end{appendices}


\begin{thebibliography}{10}
\providecommand{\url}[1]{#1}
\csname url@samestyle\endcsname
\providecommand{\newblock}{\relax}
\providecommand{\bibinfo}[2]{#2}
\providecommand{\BIBentrySTDinterwordspacing}{\spaceskip=0pt\relax}
\providecommand{\BIBentryALTinterwordstretchfactor}{4}
\providecommand{\BIBentryALTinterwordspacing}{\spaceskip=\fontdimen2\font plus
\BIBentryALTinterwordstretchfactor\fontdimen3\font minus
  \fontdimen4\font\relax}
\providecommand{\BIBforeignlanguage}[2]{{%
\expandafter\ifx\csname l@#1\endcsname\relax
\typeout{** WARNING: IEEEtran.bst: No hyphenation pattern has been}%
\typeout{** loaded for the language `#1'. Using the pattern for}%
\typeout{** the default language instead.}%
\else
\language=\csname l@#1\endcsname
\fi
#2}}
\providecommand{\BIBdecl}{\relax}
\BIBdecl

\bibitem{bouwmans2014traditional}
T.~Bouwmans, ``Traditional and recent approaches in background modeling for
  foreground detection: An overview,'' \emph{Comput. Sci. Rev.}, vol.~11, pp.
  31--66, 2014.

\bibitem{sobral2014comprehensive}
A.~Sobral and A.~Vacavant, ``A comprehensive review of background subtraction
  algorithms evaluated with synthetic and real videos,'' \emph{Comput. Vision
  Image Understanding}, vol. 122, pp. 4--21, 2014.

\bibitem{stauffer1999adaptive}
C.~Stauffer and W.~E.~L. Grimson, ``Adaptive background mixture models for
  real-time tracking,'' in \emph{Proc. IEEE Conf. Comput. Vis. Pattern
  Recognit.}, vol.~2.\hskip 1em plus 0.5em minus 0.4em\relax IEEE, 1999, pp.
  246--252.

\bibitem{zivkovic2006efficient}
Z.~Zivkovic and F.~Van Der~Heijden, ``Efficient adaptive density estimation per
  image pixel for the task of background subtraction,'' \emph{Pattern Recognit.
  Lett.}, vol.~27, no.~7, pp. 773--780, 2006.

\bibitem{allili2007robust}
M.~S. Allili, N.~Bouguila, and D.~Ziou, ``A robust video foreground
  segmentation by using generalized gaussian mixture modeling,'' in \emph{Proc.
  IEEE Int. Conf. Comput. Robot Vis.}\hskip 1em plus 0.5em minus 0.4em\relax
  IEEE, 2007, pp. 503--509.

\bibitem{elgammal2002background}
A.~Elgammal, R.~Duraiswami, D.~Harwood, and L.~S. Davis, ``Background and
  foreground modeling using nonparametric kernel density estimation for visual
  surveillance,'' \emph{Proc. {IEEE}}, vol.~90, no.~7, pp. 1151--1163, 2002.

\bibitem{kim2005real}
K.~Kim, T.~H. Chalidabhongse, D.~Harwood, and L.~Davis, ``Real-time
  foreground--background segmentation using codebook model,'' \emph{Real-time
  Imaging}, vol.~11, no.~3, pp. 172--185, 2005.

\bibitem{guo2011hierarchical}
J.-M. Guo, Y.-F. Liu, C.-H. Hsia, M.-H. Shih, and C.-S. Hsu, ``Hierarchical
  method for foreground detection using codebook model,'' \emph{{IEEE} Trans.
  Circuits Syst. Video Technol.}, vol.~21, no.~6, pp. 804--815, 2011.

\bibitem{barnich2011vibe}
O.~Barnich and M.~Van~Droogenbroeck, ``{ViBe}: A universal background
  subtraction algorithm for video sequences,'' \emph{{IEEE} Trans. Image
  Process.}, vol.~20, no.~6, pp. 1709--1724, 2011.

\bibitem{st2015subsense}
P.-L. St-Charles, G.-A. Bilodeau, and R.~Bergevin, ``Subsense: A universal
  change detection method with local adaptive sensitivity,'' \emph{{IEEE}
  Trans. Image Process.}, vol.~24, no.~1, pp. 359--373, 2015.

\bibitem{st2016universal}
P.-L. St-Charles, G.-A. Bilodeau, and R.~Bergevin, ``Universal background
  subtraction using word consensus models,'' \emph{{IEEE} Trans. Image
  Process.}, vol.~25, no.~10, pp. 4768--4781, 2016.

\bibitem{javed2013foreground}
S.~Jabri, Z.~Duric, H.~Wechsler, and A.~Rosenfeld, ``Detection and location of
  people in video images using adaptive fusion of color and edge information,''
  in \emph{Proc. IEEE Int. Conf. Pattern Recognit.}, vol.~4.\hskip 1em plus
  0.5em minus 0.4em\relax IEEE, 2000, pp. 627--630.

\bibitem{maddalena2017exploiting}
L.~Maddalena and A.~Petrosino, ``Exploiting color and depth for background
  subtraction,'' in \emph{Int. Conf. Image Anal. Process.}\hskip 1em plus 0.5em
  minus 0.4em\relax Springer, 2017, pp. 254--265.

\bibitem{zhou2005modified}
D.~Zhou and H.~Zhang, ``Modified {GMM} background modeling and optical flow for
  detection of moving objects,'' in \emph{{IEEE} Trans. Syst., Man, Cybern.},
  vol.~3.\hskip 1em plus 0.5em minus 0.4em\relax IEEE, 2005, pp. 2224--2229.

\bibitem{heikkila2006texture}
M.~Heikkila and M.~Pietikainen, ``A texture-based method for modeling the
  background and detecting moving objects,'' \emph{{IEEE} Trans. Pattern Anal.
  Mach. Intell.}, vol.~28, no.~4, pp. 657--662, 2006.

\bibitem{st2014improving}
P.-L. St-Charles and G.-A. Bilodeau, ``Improving background subtraction using
  local binary similarity patterns,'' in \emph{Proc. IEEE Winter Conf. Appl.
  Comput. Vision.}\hskip 1em plus 0.5em minus 0.4em\relax IEEE, 2014, pp.
  509--515.

\bibitem{braham2017semantic}
M.~Braham, S.~Pi{\'e}rard, and M.~Van~Droogenbroeck, ``Semantic background
  subtraction,'' in \emph{Proc. IEEE Int. Conf. Image Process.}\hskip 1em plus
  0.5em minus 0.4em\relax IEEE, 2017, pp. 4552--4556.

\bibitem{zhao2017pyramid}
H.~Zhao, J.~Shi, X.~Qi, X.~Wang, and J.~Jia, ``Pyramid scene parsing network,''
  in \emph{Proc. IEEE Conf. Comput. Vis. Pattern Recognit.}\hskip 1em plus
  0.5em minus 0.4em\relax IEEE, 2017, pp. 2881--2890.

\bibitem{fan2017parallel}
H.~Fan and H.~Ling, ``Parallel tracking and verifying: A framework for
  real-time and high accuracy visual tracking,'' \emph{arXiv preprint
  arXiv:1708.00153}, 2017.

\bibitem{wang2014cdnet}
Y.~Wang, P.-M. Jodoin, F.~Porikli, J.~Konrad, Y.~Benezeth, and P.~Ishwar,
  ``{CDnet} 2014: An expanded change detection benchmark dataset,'' in
  \emph{Proc. IEEE Conf. Comput. Vis. Pattern Recognit. Workshops}.\hskip 1em
  plus 0.5em minus 0.4em\relax IEEE, 2014, pp. 393--400.

\bibitem{garcia2017review}
A.~Garcia-Garcia, S.~Orts-Escolano, S.~Oprea, V.~Villena-Martinez, and
  J.~Garcia-Rodriguez, ``A review on deep learning techniques applied to
  semantic segmentation,'' \emph{arXiv preprint arXiv:1704.06857}, 2017.

\bibitem{lee2005effective}
D.-S. Lee, ``Effective gaussian mixture learning for video background
  subtraction,'' \emph{{IEEE} Trans. Pattern Anal. Mach. Intell.}, vol.~27,
  no.~5, pp. 827--832, 2005.

\bibitem{mittal2004motion}
A.~Mittal and N.~Paragios, ``Motion-based background subtraction using adaptive
  kernel density estimation,'' in \emph{Proc. IEEE Conf. Comput. Vis. Pattern
  Recognit.}, vol.~2.\hskip 1em plus 0.5em minus 0.4em\relax IEEE, 2004, pp.
  II--II.

\bibitem{tan2010enhanced}
X.~Tan and B.~Triggs, ``Enhanced local texture feature sets for face
  recognition under difficult lighting conditions,'' \emph{{IEEE} Trans. Image
  Process.}, vol.~19, no.~6, pp. 1635--1650, 2010.

\bibitem{liao2010modeling}
S.~Liao, G.~Zhao, V.~Kellokumpu, M.~Pietik{\"a}inen, and S.~Z. Li, ``Modeling
  pixel process with scale invariant local patterns for background subtraction
  in complex scenes,'' in \emph{Proc. IEEE Conf. Comput. Vis. Pattern
  Recognit.}\hskip 1em plus 0.5em minus 0.4em\relax IEEE, 2010, pp. 1301--1306.

\bibitem{braham2016deep}
M.~Braham and M.~Van~Droogenbroeck, ``Deep background subtraction with
  scene-specific convolutional neural networks,'' in \emph{Proc. IEEE Int.
  Conf. Syst., Signals and Image Process.}\hskip 1em plus 0.5em minus
  0.4em\relax IEEE, 2016, pp. 1--4.

\bibitem{babaee2018deep}
M.~Babaee, D.~T. Dinh, and G.~Rigoll, ``A deep convolutional neural network for
  video sequence background subtraction,'' \emph{Pattern Recognit.}, vol.~76,
  pp. 635--649, 2018.

\bibitem{wang2017interactive}
Y.~Wang, Z.~Luo, and P.-M. Jodoin, ``Interactive deep learning method for
  segmenting moving objects,'' \emph{Pattern Recognit. Lett.}, vol.~96, pp.
  66--75, 2017.

\bibitem{lim2018foreground}
L.~A. Lim and H.~Y. Keles, ``Foreground segmentation using a triplet
  convolutional neural network for multiscale feature encoding,'' \emph{arXiv
  preprint arXiv:1801.02225}, 2018.

\bibitem{zeng2018multiscale}
D.~Zeng and M.~Zhu, ``Background subtraction using multiscale fully
  convolutional network,'' \emph{{IEEE} Access}, vol.~6, pp. 16\,010--16\,021,
  2018.

\bibitem{farabet2013learning}
C.~Farabet, C.~Couprie, L.~Najman, and Y.~LeCun, ``Learning hierarchical
  features for scene labeling,'' \emph{{IEEE} Trans. Pattern Anal. Mach.
  Intell.}, vol.~35, no.~8, pp. 1915--1929, 2013.

\bibitem{long2015fully}
J.~Long, E.~Shelhamer, and T.~Darrell, ``Fully convolutional networks for
  semantic segmentation,'' in \emph{Proc. IEEE Conf. Comput. Vis. Pattern
  Recognit.}\hskip 1em plus 0.5em minus 0.4em\relax IEEE, 2015, pp. 3431--3440.

\bibitem{yu2015multi}
F.~Yu and V.~Koltun, ``Multi-scale context aggregation by dilated
  convolutions,'' \emph{arXiv preprint arXiv:1511.07122}, 2015.

\bibitem{chen2018deeplab}
L.-C. Chen, G.~Papandreou, I.~Kokkinos, K.~Murphy, and A.~L. Yuille, ``Deeplab:
  Semantic image segmentation with deep convolutional nets, atrous convolution,
  and fully connected crfs,'' \emph{{IEEE} Trans. Pattern Anal. Mach. Intell.},
  vol.~40, no.~4, pp. 834--848, 2018.

\bibitem{ronneberger2015u}
O.~Ronneberger, P.~Fischer, and T.~Brox, ``U-net: Convolutional networks for
  biomedical image segmentation,'' in \emph{Int. Conf. Medical Image Computing
  Computer-Assisted Intervention}.\hskip 1em plus 0.5em minus 0.4em\relax
  Springer, 2015, pp. 234--241.

\bibitem{liu2015parsenet}
W.~Liu, A.~Rabinovich, and A.~C. Berg, ``Parsenet: Looking wider to see
  better,'' \emph{arXiv preprint arXiv:1506.04579}, 2015.

\bibitem{chen2016attention}
L.-C. Chen, Y.~Yang, J.~Wang, W.~Xu, and A.~L. Yuille, ``Attention to scale:
  Scale-aware semantic image segmentation,'' in \emph{Proc. IEEE Conf. Comput.
  Vis. Pattern Recognit.}\hskip 1em plus 0.5em minus 0.4em\relax IEEE, 2016,
  pp. 3640--3649.

\bibitem{xia2016zoom}
F.~Xia, P.~Wang, L.-C. Chen, and A.~L. Yuille, ``Zoom better to see clearer:
  Human and object parsing with hierarchical auto-zoom net,'' in \emph{European
  Conf. on Comput. Vis.}\hskip 1em plus 0.5em minus 0.4em\relax Springer, 2016,
  pp. 648--663.

\bibitem{noh2015learning}
H.~Noh, S.~Hong, and B.~Han, ``Learning deconvolution network for semantic
  segmentation,'' in \emph{Proc. IEEE Int. Conf. Comput. Vis.}, 2015, pp.
  1520--1528.

\bibitem{liu2015semantic}
Z.~Liu, X.~Li, P.~Luo, C.-C. Loy, and X.~Tang, ``Semantic image segmentation
  via deep parsing network,'' in \emph{Proc. IEEE Int. Conf. Comput.
  Vis.}\hskip 1em plus 0.5em minus 0.4em\relax IEEE, 2015, pp. 1377--1385.

\bibitem{badrinarayanan2017segnet}
V.~Badrinarayanan, A.~Kendall, and R.~Cipolla, ``Segnet: A deep convolutional
  encoder-decoder architecture for image segmentation,'' \emph{{IEEE} Trans.
  Pattern Anal. Mach. Intell.}, vol.~39, no.~12, pp. 2481--2495, 2017.

\bibitem{paszke2016enet}
A.~Paszke, A.~Chaurasia, S.~Kim, and E.~Culurciello, ``Enet: A deep neural
  network architecture for real-time semantic segmentation,'' \emph{arXiv
  preprint arXiv:1606.02147}, 2016.

\bibitem{chaurasia2017linknet}
A.~Chaurasia and E.~Culurciello, ``Linknet: Exploiting encoder representations
  for efficient semantic segmentation,'' \emph{arXiv preprint
  arXiv:1707.03718}, 2017.

\bibitem{romera2018erfnet}
E.~Romera, J.~M. Alvarez, L.~M. Bergasa, and R.~Arroyo, ``{ERFNet}: Efficient
  residual factorized convnet for real-time semantic segmentation,''
  \emph{{IEEE} Trans. Intell. Transp. Syst.}, vol.~19, no.~1, pp. 263--272,
  2018.

\bibitem{zhao2018icnet}
H.~Zhao, X.~Qi, X.~Shen, J.~Shi, and J.~Jia, ``{ICNet} for real-time semantic
  segmentation on high-resolution images,'' in \emph{European Conf. on Comput.
  Vis.}, 2018.

\bibitem{hofmann2012background}
M.~Hofmann, P.~Tiefenbacher, and G.~Rigoll, ``Background segmentation with
  feedback: The pixel-based adaptive segmenter,'' in \emph{Proc. IEEE Conf.
  Comput. Vis. Pattern Recognit. Workshops}.\hskip 1em plus 0.5em minus
  0.4em\relax IEEE, 2012, pp. 38--43.

\bibitem{zhou2017scene}
B.~Zhou, H.~Zhao, X.~Puig, S.~Fidler, A.~Barriuso, and A.~Torralba, ``Scene
  parsing through ade20k dataset,'' in \emph{Proc. IEEE Conf. Comput. Vis.
  Pattern Recognit.}\hskip 1em plus 0.5em minus 0.4em\relax IEEE, 2017.

\bibitem{bianco2017far}
S.~Bianco, G.~Ciocca, and R.~Schettini, ``How far can you get by combining
  change detection algorithms?'' in \emph{Int. Conf. Image Anal.
  Process.}\hskip 1em plus 0.5em minus 0.4em\relax Springer, 2017, pp. 96--107.

\bibitem{chen2015learning}
Y.~Chen, J.~Wang, and H.~Lu, ``Learning sharable models for robust background
  subtraction,'' in \emph{Proc. IEEE Int. Conf. Multimultimedia and
  Expo}.\hskip 1em plus 0.5em minus 0.4em\relax IEEE, 2015, pp. 1--6.

\bibitem{jiang2017wesambe}
S.~Jiang and X.~Lu, ``{WeSamBE}: A weight-sample-based method for background
  subtraction,'' \emph{{IEEE} Trans. Circuits Syst. Video Technol.}, 2017.

\bibitem{allebosch2015c}
G.~Allebosch, D.~Van~Hamme, F.~Deboeverie, P.~Veelaert, and W.~Philips,
  ``{C-EFIC}: Color and edge based foreground background segmentation with
  interior classification,'' in \emph{Int. Conf. Comput. Vis. Imaging Comput.
  Graph.}\hskip 1em plus 0.5em minus 0.4em\relax Springer, 2015, pp. 433--454.
  
\bibitem{bouwmans2018deep}
T.~Bouwmans, S.~Javed, M.~Sultana, and S.~K. Jung, ``Deep neural network
  concepts for background subtraction: A systematic review and comparative
  evaluation,'' \emph{arXiv preprint arXiv:1811.05255}, 2018.

\end{thebibliography}
\end{document}